\documentclass{article}

\usepackage{microtype}
\usepackage{graphicx}
\usepackage{subcaption}
\usepackage{booktabs} %

\usepackage{amsmath}
\usepackage{amssymb}
\usepackage{mathtools}
\usepackage{amsthm}
\usepackage{amsfonts,bm}

\newcommand{\expect}{\mathbb{E}}
\newcommand{\nan}{\mathrm{NaN}}

\newcommand{\defeq}{:=}
\renewcommand{\Re}{\mathbb{R}}
\renewcommand{\xi}{\mX[i,:]}
\newcommand{\xj}{\mX[j,:]}

\def\eqref#1{equation~\ref{#1}}

\def\1{\bm{1}}

\def\va{{\bm{a}}}
\def\vb{{\bm{b}}}

\def\vx{{\bm{x}}}
\def\vy{{\bm{y}}}
\def\vz{{\bm{z}}}

\def\mG{{\bm{G}}}

\def\mM{{\bm{M}}}

\def\mP{{\bm{P}}}

\def\mX{{\bm{X}}}

\def\mZ{{\bm{Z}}}

\DeclareMathAlphabet{\mathsfit}{\encodingdefault}{\sfdefault}{m}{sl}
\SetMathAlphabet{\mathsfit}{bold}{\encodingdefault}{\sfdefault}{bx}{n}

\DeclareMathOperator*{\argmax}{arg\,max}
\DeclareMathOperator*{\argmin}{arg\,min}

\usepackage{hyperref}
\usepackage[accepted]{icml2023}
\usepackage[algo2e,lined,commentsnumbered,ruled]{algorithm2e}
\makeatletter
\newcommand{\removelatexerror}{\let\@latex@error\@gobble}
\makeatother

\theoremstyle{plain}
\newtheorem{theorem}{Theorem}[section]
\newtheorem{proposition}[theorem]{Proposition}
\newtheorem{lemma}[theorem]{Lemma}

\theoremstyle{definition}

\theoremstyle{remark}

\newcommand{\rowa}[1]{\renewcommand{\arraystretch}{#1}}

\usepackage[textsize=tiny]{todonotes}

\icmltitlerunning{Transformed Distribution Matching for Missing Value Imputation}

\begin{document}

\twocolumn[
\icmltitle{Transformed Distribution Matching for Missing Value Imputation}

\icmlsetsymbol{equal}{*}

\begin{icmlauthorlist}
\icmlauthor{He Zhao}{d}
\icmlauthor{Ke Sun}{d}
\icmlauthor{Amir Dezfouli}{d}
\icmlauthor{Edwin V. Bonilla}{d}
\end{icmlauthorlist}

\icmlaffiliation{d}{CSIRO's Data61, Australia}

\icmlcorrespondingauthor{He Zhao}{he.zhao@ieee.org}

\icmlkeywords{Machine Learning, ICML}

\vskip 0.3in
]

\printAffiliationsAndNotice{}  %

\begin{abstract}
We study the problem of imputing missing values in a dataset, which has important applications in many domains. The key to missing value imputation is to capture the data distribution with incomplete samples and impute the missing values accordingly.
In this paper, by leveraging the fact that any two batches of data with missing values come from the same data distribution, we propose to impute the missing values of two batches of samples by transforming them into a latent space through deep invertible functions and matching them distributionally.
To learn the transformations and impute the missing values simultaneously, a simple and well-motivated algorithm is proposed.
Our algorithm has fewer hyperparameters to fine-tune and generates high-quality imputations regardless of how missing values are generated.
Extensive experiments over a large number of datasets and competing benchmark algorithms show that our method achieves state-of-the-art performance\footnote{Code at~\url{https://github.com/hezgit/TDM}}.
\end{abstract}

\section{Introduction}
In practice, real-world data are usually incomplete and consist of many missing values. For example, in the medical domain, the health record of a patient may have considerable missing items as not all of the characteristics are properly recorded or not all of the tests have been done for the patient~\cite{barnard1999applications}. 
In this paper, we are interested in imputing missing values in an unsupervised way~\citep{van2011mice,yoon2018gain,mattei2019miwae,muzellec2020missing,jarrett2022hyperimpute}.
Here the meaning of ``unsupervised'' is twofold: We do not know the ground truth of the missing values during training and we do not assume a specific downstream task. 

The key to missing value imputation is how to model the data distribution with a considerable amount of missing values, which is a notoriously challenging problem. 
To address this challenge, existing approaches either choose to model the conditional data distribution instead (i.e., the distribution of one feature conditioned on the other features), such as in~\citet{heckerman2000dependency,raghunathan2001multivariate,gelman2004parameterization,van2006fully,van2011mice,liu2014stationary,zhu2015convergence} or use deep generative models to capture the data distribution, such as in~\citet{gondara2017multiple,ivanov2018variational,mattei2019miwae,nazabal2020handling,gong2021variational,peis2022missing,yoon2018gain,li2018misgan,yoon2020gamin,dai2021multiple,fang2022fragmgan,richardson2020mcflow,ma2021emflow,wang2022generative}.
Alternatively, a recent interesting idea proposed by~\citet{muzellec2020missing} has found success, whose key insight is that any two batches of data (with missing values) come from the same data distribution. Thus, a good method should impute the missing values to make the empirical distributions of the two batches matched, i.e., distributionally close to each other. This is a more general and applicable assumption that can be used in various data under different missing value mechanisms. In this paper, we refer to this idea as \textit{distribution matching} (DM), the appealing property of which is that it bypasses modelling the data distribution explicitly or implicitly, a difficult task even without missing values.

As the pioneering study, \citet{muzellec2020missing} does DM by minimising the optimal transport (OT) distance whose cost function is the quadratic distance in the data space between data samples. However, real-world data usually exhibit complex geometry, which might not be captured well by the quadratic distance in the data space. This can lead to wrong imputations and poor performance.
In this paper, we propose a new, straightforward, yet powerful DM method, which first transforms data samples into a latent space through a deep invertible function and then does distribution matching with OT in the latent space. In the latent space, the quadratic distance of two samples is expected to better reflect their (dis)similarity under the geometry of the data considered. In the missing-value setting, learning a good transformation is non-trivial. We propose a simple and elegant algorithm that learns the transformations and imputes the missing values simultaneously, which is well-motivated by the theory of OT and representation learning.

The contributions of this paper include:
\textbf{1)} As DM is a new and promising line of research in missing value imputation,
we propose a well-motivated transformed distribution matching method, which significantly improves over previous methods.
\textbf{2)} In practice, the ground truth of missing values is usually unknown, making it hard to fine-tune methods with complex algorithms and many hyperparameters.
We develop a simple and theoretically sound learning algorithm with a single loss and very few hyperparameters, alleviating the need for extensive fine-tuning.
\textbf{3)} We conduct extensive experiments over a large number of datasets and competing benchmark algorithms in multiple missing-value mechanisms and report comprehensive evaluation metrics. These experiments show that our method achieves state-of-the-art performance.

\section{Background}
\subsection{Data with Missing Values}
\label{sec-bg-missing}
Here we consider $N$ data samples with $D$-dimensional features stored in matrix $\mX \in \Re^{N\times D}$,
where a row vector $\mX[i,:] \in \Re^D$ ($1 \le i \le N$) represents the $i^{\text{th}}$ sample.
The missing values contained in $\mX$ are indicated by a binary mask $\mM \in \{0,1\}^{N \times D}$ such that  $\mM[i,d]=1$ indicates that the $d^{\text{th}}$ feature of sample $i$ is missing and $\mM[i,d]=0$ otherwise.
Moreover, we assign $\nan$ (Not a Number) to the missing values and use the following Python/Numpy style matrix indexing $\mX[\mM]=\nan$ to denote the missing data. Similarly, the observed data can be denoted as $\mX[\mathbf{1} - \mM]$, where $\mathbf{1}_{N \times D}$ is the matrix with the same dimension as $\mM$ filled with ones. The task of imputation is to fill $\mX[\mM]$ with the imputed values given the mask $\mM$ and the observed values $\mX[\mathbf{1} - \mM]$.

In practice, three missing value patterns/mechanisms (i.e., ways of generating the mask) have been widely explored~\citep{rubin1976inference,rubin2004multiple,van2018flexible,seaman2013meant}: missing completely at random (MCAR) where the missingness is independent of the data; missing at random (MAR) where the probability of being missing depends only on observed values; missing not at random (MNAR) where the probability of missingness then depends on the unobserved values.
In MCAR and MAR, the missing value patterns are ``ignorable'' as it is unnecessary to model the distribution of missing values explicitly while MNAR can be a harder case where missing values may lead to important biases in data~\citep{muzellec2020missing,jarrett2022hyperimpute}.

\subsection{Optimal Transport}
Optimal transport (OT) provides sound and meaningful distances to compare  distributions~\cite{peyre2019computational}, which has been used in many problems, such as computer vision~\cite{ge2021ota,zhang2022deepemd}, text analysis~\cite{huynh2020otlda,zhaoneural,danrepresenting}, adversarial robustness~\cite{buiunified}, probabilistic (generative) models~\cite{vuong2023vector,vo2023learning}, and other machine learning applications~\cite{nguyen2021most,danlearning,nguyen2022cycle,guo2022learning,danadaptive}.
Here we briefly introduce OT between two discrete distributions of dimensionality $B$ and $B^\prime$, respectively. 
Let  $\alpha(\mX^1) \defeq \sum_{i=1}^B a_i \delta_{\mX^1[i,:]}$ and $\beta(\mX^2) \defeq \sum_{j=1}^{B'} b_j \delta_{\mX^2[j,:]}$, where $\mX^1 \in \mathbb{R}^{B \times D}$ and $\mX^2 \in \mathbb{R}^{B' \times D}$ denote the supports of the two distributions, respectively; 
$\va \in \Delta^{B}$ and $\vb \in \Delta^{B'}$ are two probability vectors; $\Delta^B$ is the $(B-1)$-dimensional probability simplex.
The OT distance between $\alpha(\mX^1)$ and $\beta(\mX^2)$ can be defined as:
\vspace{0.01cm}
\begin{equation}
\label{eq-def-ot}
d_{\mG}\left(\alpha(\mX^1), \beta(\mX^2)\right) 
\defeq
\inf_{\mP \in U(\va, \vb)} \langle \mP , \mG \rangle,
\end{equation}
where $\langle\cdot,\cdot\rangle$ denotes the Frobenius dot-product;
$\mG \in \mathbb{R}_{\ge 0}^{B \times B'}$ is the cost matrix/function of the transport;
$\mP \in \mathbb{R}_{>0}^{B \times B'}$ is the transport matrix/plan;
$U(\va, \vb)$ denotes the transport polytope of $\va$ and $\vb$,
which is the polyhedral set of $B \times B'$ matrices:
$U(\va, \vb) \defeq \left\{\mP \in \mathbb{R}^{B \times B'}_{>0} | \mP \boldsymbol{1}_{B'} = \va, \mP^T \boldsymbol{1}_{B} = \vb\right\}$;
and $\boldsymbol{1}_{B}$ is the $N$-dimensional column vector of ones.
The cost matrix $\mG$ contains the pairwise distance/cost between $B$ and $B'$ supports, for which different distance metrics can be used.
Specifically, if the cost is defined as the pairwise quadratic distance: $\mG[i,j] = \lVert \mX^1[i,:] - \mX^2[j,:]\rVert^2$, 
where $\lVert\cdot\rVert$ is the Euclidean norm, the OT distance reduces to the $p$-Wasserstein distance ($p=2$ here), i.e., $d_{\mG}\left(\alpha(\mX^1), \beta(\mX^2)\right) = W_2^2\left(\alpha(\mX^1), \beta(\mX^2)\right)$.

\section{Method}
\label{sec-method}

\subsection{Previous Work: Optimal Transport for Missing Value Imputation}

Our method is motivated by \citet{muzellec2020missing}, the pioneering method of distribution matching,
recently proposed for missing value imputation.
Consider two batches of data of $\mX$: $\mX^1$ and $\mX^2$ with batch size of $B$, both of which can contain missing values. The key insight is that a good method should impute the missing values in $\mX^1$ and $\mX^2$ so that the empirical distributions of them are matched.
To do so, 
\citet{muzellec2020missing} propose to minimise the OT distance\footnote{The paper actually uses the Sinkhorn divergence~\cite{genevay2018learning,feydy2019interpolating}, a surrogate divergence to OT.} between them in terms of the missing values.
\begin{align}
\label{eq-ot-loss}
    \min_{\mX^{1\cup 2}[\mM^{1\cup 2}]} W_2^2\left(\mu(\mX^1), \mu(\mX^2)\right),
\end{align}
where $\mu(\mX^1)=\frac{1}{B} \sum_i \delta_{\mX^1[i,:]}$ denotes the empirical measure associated to the $B$ samples of $\mX^{1}$ (similarly for $\mu(\mX^2)$); $\mX^{1\cup 2}$ denotes the union of $\mX^1$ and $\mX^2$, i.e., the unique data samples of the two batches; $\mM^{1\cup 2}$ denotes the union of $\mM^1$ and $\mM^2$, i.e., the mask of missing values in $\mX^{1\cup 2}$. To impute the missing values, one can take an iterative update of $\mX^{1\cup 2}[\mM^{1\cup 2}]$ by gradient descent, e.g., RMSprop~\cite{tieleman2012lecture}: $\mX^{1\cup 2}[\mM^{1\cup 2}] \leftarrow \mX^{1\cup 2}[\mM^{1\cup 2}] - \alpha \text{RMSprop}\left(\nabla_{\mX^{1\cup 2}[\mM^{1\cup 2}]} W_2^2\left(\mu(\mX^1), \mu(\mX^2)\right)\right)$. We refer to this method as ``OTImputer''.

\begin{figure}[t]
\captionsetup[subfigure]{justification=centering}
        \centering
         \begin{subfigure}[b]{0.45\linewidth}
                 \centering
                 \caption{Ground truth}
                 \includegraphics[width=0.99\textwidth]{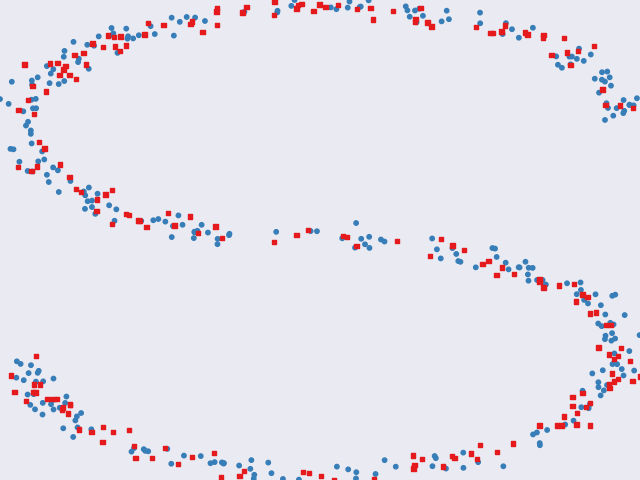}
         \end{subfigure}
         \begin{subfigure}[b]{0.45\linewidth}
                 \centering
                 \caption{OTImputer}
                 \includegraphics[width=0.99\textwidth]{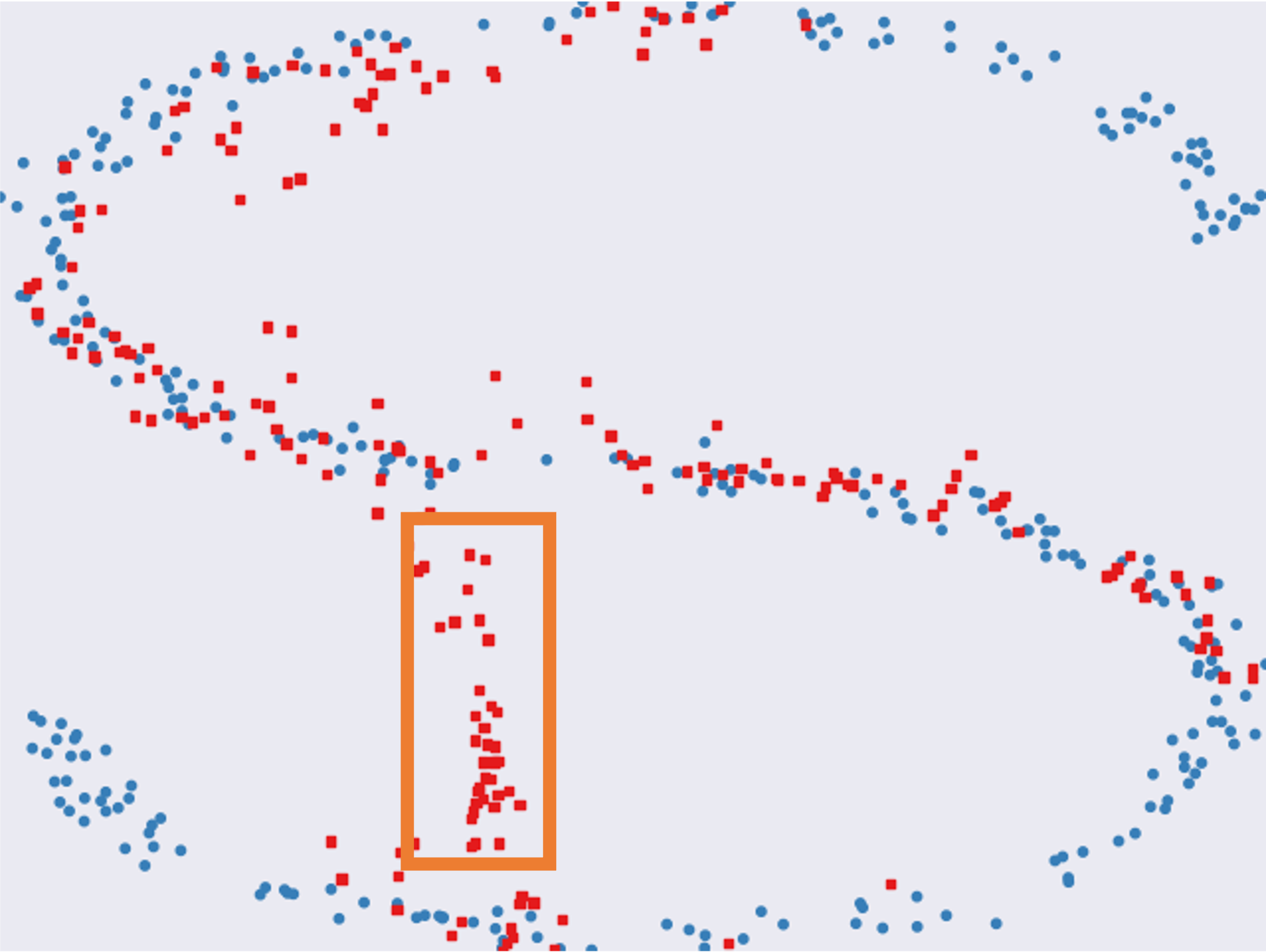}
         \end{subfigure} \\
         \begin{subfigure}[b]{0.45\linewidth}
                 \centering
                 \includegraphics[width=0.99\textwidth]{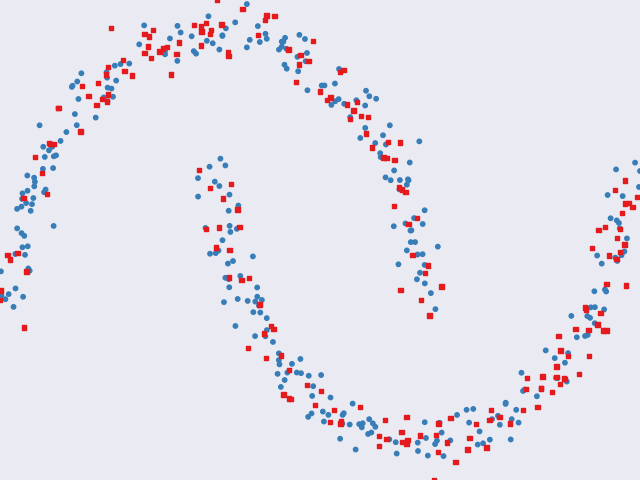}
         \end{subfigure}
         \begin{subfigure}[b]{0.45\linewidth}
                 \centering
                 \includegraphics[width=0.99\textwidth]{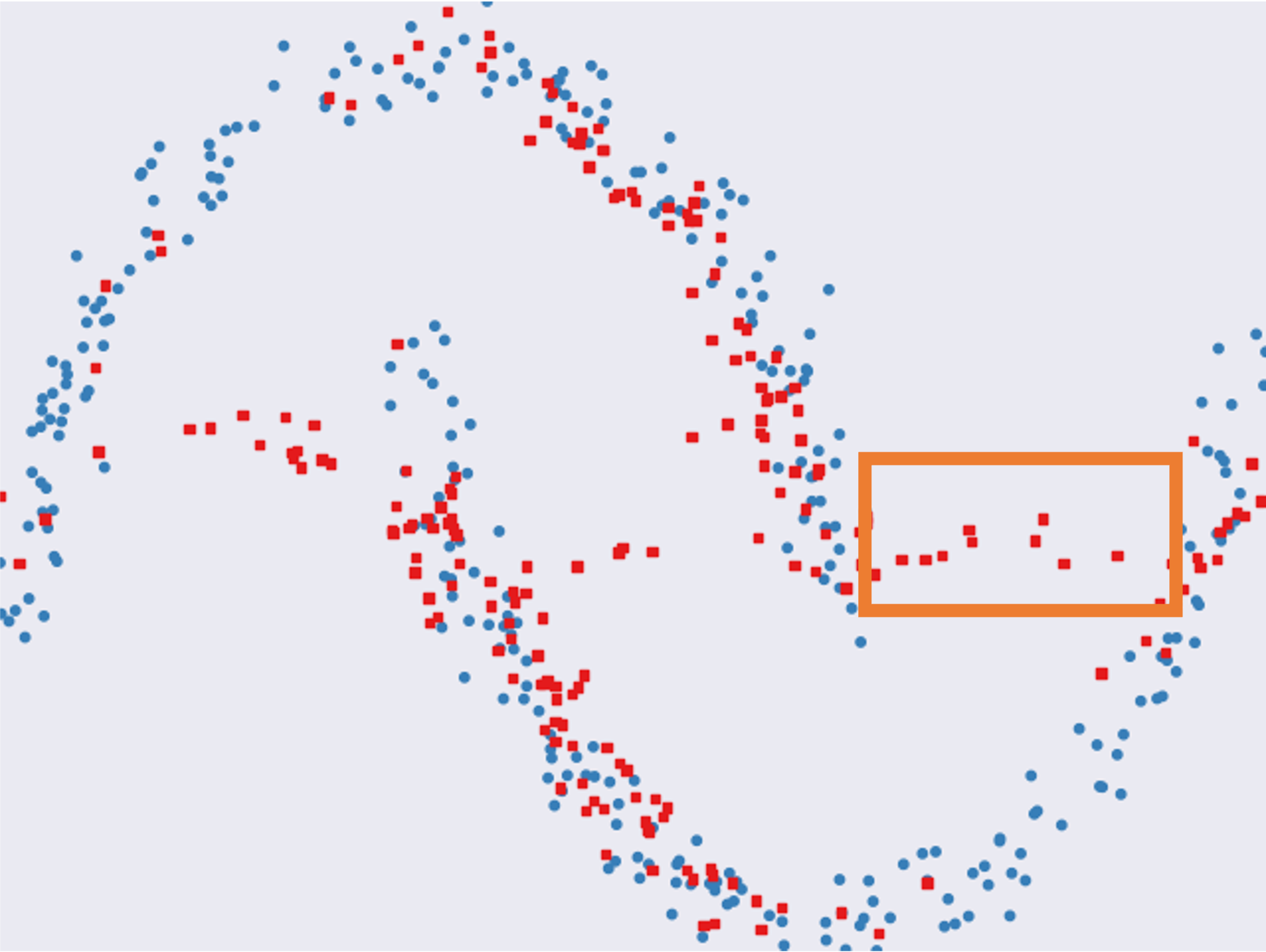}
         \end{subfigure} 
        
\caption{Two synthetic datasets (two rows) each of which is with 500 samples. (a) Ground truth: Blue points (60\%) have no missing values and red points (40\%) have one missing value on either coordinate (following MCAR). (b) The imputed values for the red points by OTImputer\footnotemark, where the orange rectangle highlights the region of interest.}
\label{fig-motivation-1}
\end{figure}

\subsection{Motivations}
\label{sec-motivation}

We now motivate the proposed method by giving a closer look at OTImputer.
\begin{lemma}
\label{lemma}
For any given $\mX^1$ and $\mX^2$, we have:
\begin{align*}
    &W_2^2\left(\mu(\mX^1), \mu(\mX^2)\right)\\
    &= \min_{\pi} \frac{1}{B} \sum_{i=1}^B \lVert \mX^1[i,:] - \mX^2[\pi(i),:]\rVert^2, \nonumber
\end{align*}
where the minimum is taken over all possible permutations $\pi$ of the sequence $(1,\dots, B)$, and $\pi(i)$ is the permuted index of $i$ with $1 \le \pi(i) \le B$.
\vspace{-2mm}
\end{lemma}
\begin{proof}
It is a special case of Proposition 2 of~\citet{nguyen2011wasserstein}.
\vspace{-5mm}
\end{proof}

\footnotetext{Note that Figure 1 is not directly comparable with Figure 2 in~\citet{muzellec2020missing} because the synthetic dataset is generated differently, the missing value proportion is different, and the computation of OT is done differently.}

The above lemma shows that 2-Wasserstein distance between two empirical distributions used in OTImputer is equivalent to the minimisation of a matching distance by finding the optimal permutation.
In the missing value imputation context, OTImputer finds the closest pairs of data samples in the two batches in terms of the quadratic distance in the data space (by the computation of 2-Wasserstein) and then tries the hardest to minimise their quadratic distance to impute missing values (by the minimisation of 2-Wasserstein). Real-world data usually exhibit complex geometry, which can hardly be captured by the quadratic distance in the data space.  Figure~\ref{fig-motivation-1} shows the imputation results on two synthetic datasets.
It can be seen that OTImputer incorrectly imputes a considerable amount of missing values within the orange rectangles.
This is because these imputed samples have a small quadratic distance to others in the data space but they are not good imputations. Therefore, the quadratic distance in the data space is unable to reflect the data geometry.

\subsection{Proposed Method}
In this paper, we introduce \textbf{T}ransformed \textbf{D}istribution \textbf{M}atching (TDM), which carries out OT-based missing value imputation on a \emph{transformed} space, 
where the distances between the transformed  samples can reveal the  similarity/dissimilarity between them better, respecting the underlying geometry of the data.
Specifically, we aim to learn a deep transformation parameterised by $\theta$, $f_\theta: \mathbb{R}^{D'} \rightarrow \mathbb{R}^D$ that projects a data sample $\vx \in \mathbb{R}^D$ to a transformed one $\vz \in \mathbb{R}^{D'}$: $\vz \defeq f_\theta(\vx)$.
With a slight abuse of notation, we denote the batch-level transformation as $\mZ = f_\theta(\mX)$ where $f_\theta$ is applied to each sample (row vector) in $\mX$. 
Generalising Eq.~(\ref{eq-ot-loss}), we learn $f_\theta$ and the imputations by:
\begin{align}\label{eq-proj-loss-1}
     &\min_{\mX^{1\cup 2}[\mM^{1\cup 2}], \theta} \mathcal{L}^{W}(\mX^1, \mX^2), \\
     &\mathcal{L}^{W}(\mX^1, \mX^2) = W_2^2\left(f_\#\mu(\mX^1), f_\#\mu(\mX^2)\right),
\end{align}
where 
$f_\#{\mu}(\mX) \defeq \mu(f_\theta(\mX))$.
If $f_\theta$ is an isometry, then our TDM reduces to OTImputer~\citep{muzellec2020missing}. We provide more theoretical analysis of this loss in Section~\ref{sec:analysis} of the appendix.

\begin{figure*}[t]
\captionsetup[subfigure]{justification=centering}
        \centering
         \begin{subfigure}[b]{0.19\linewidth}
                 \centering
                 \caption{TDM (ours)}
                 \includegraphics[width=0.99\textwidth]{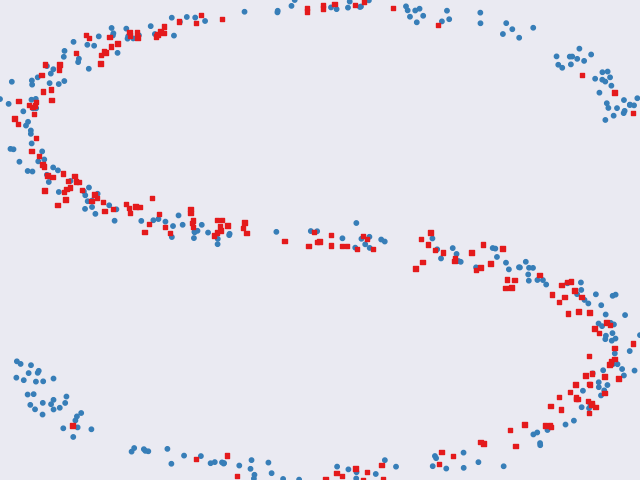}
         \end{subfigure}
         \begin{subfigure}[b]{0.19\linewidth}
                 \centering
                 \caption{$\mX$ with OOD noises}
                 \includegraphics[width=0.99\textwidth]{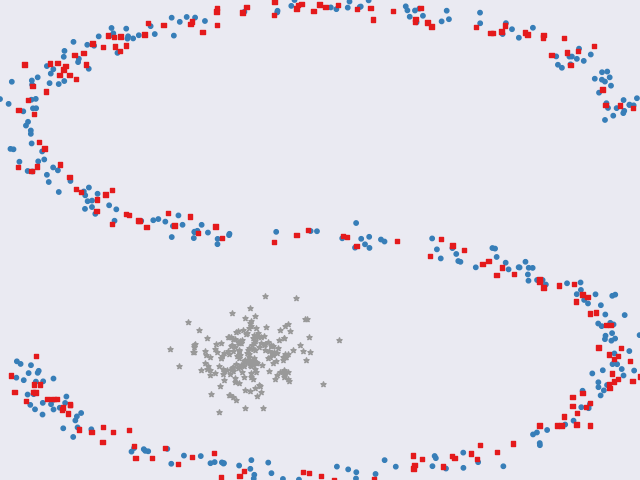}
         \end{subfigure} 
         \begin{subfigure}[b]{0.19\linewidth}
                 \centering
                 \caption{$f_{1}(\mX)$}
                 \includegraphics[width=0.99\textwidth]{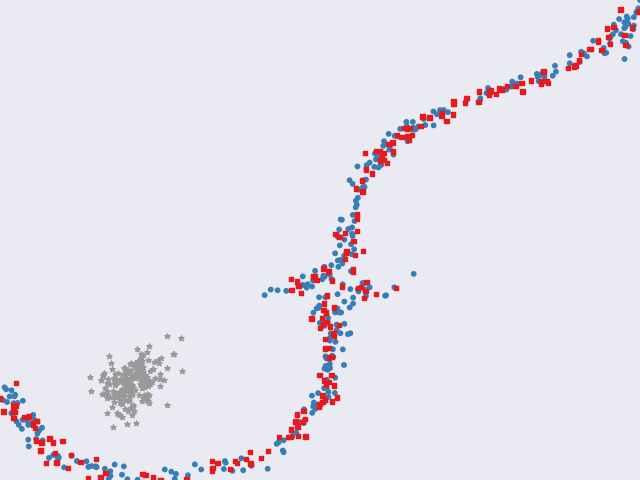}
         \end{subfigure}
          \begin{subfigure}[b]{0.19\linewidth}
                 \centering
                 \caption{$f_{1:2}(\mX)$}
                 \includegraphics[width=0.99\textwidth]{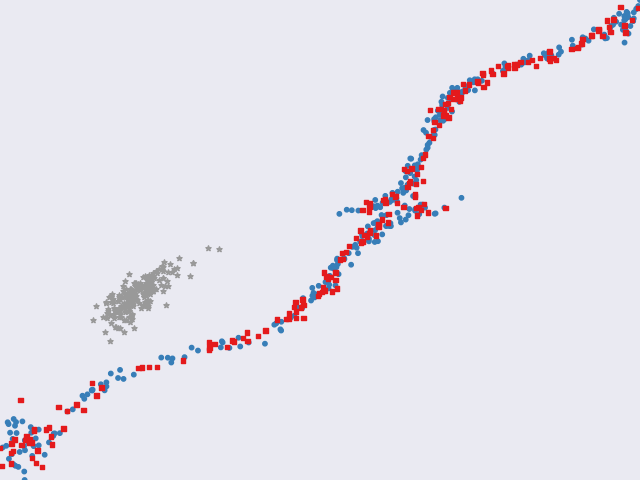}
         \end{subfigure} 
        \begin{subfigure}[b]{0.19\linewidth}
                 \centering
                 \caption{$f_{1:3}(\mX)$}
                 \includegraphics[width=0.99\textwidth]{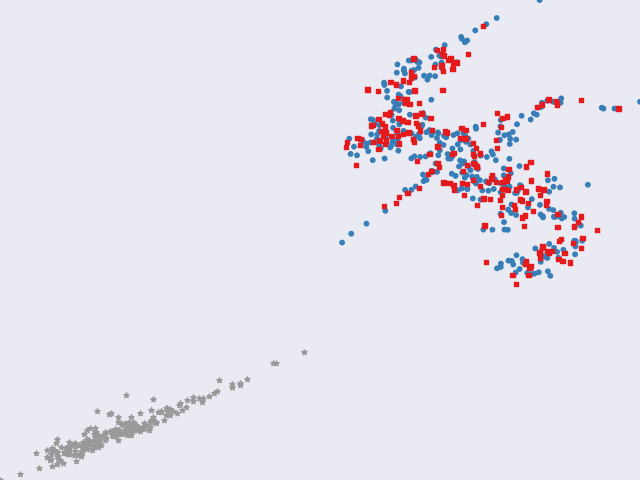}
         \end{subfigure}\\
         \begin{subfigure}[b]{0.19\linewidth}
                 \centering
                 \includegraphics[width=0.99\textwidth]{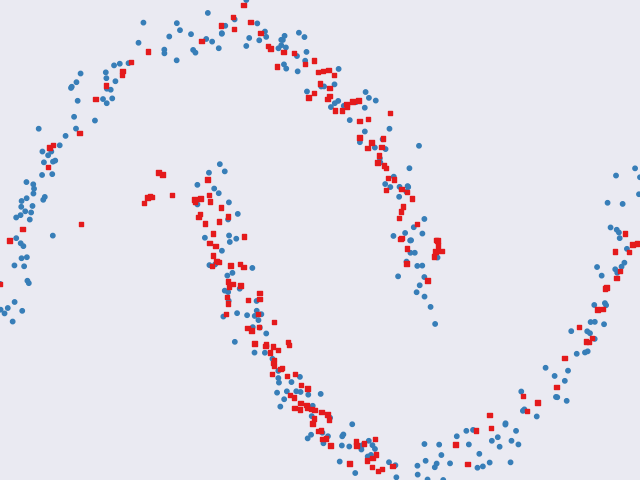}
         \end{subfigure}
         \begin{subfigure}[b]{0.19\linewidth}
                 \centering
                 \includegraphics[width=0.99\textwidth]{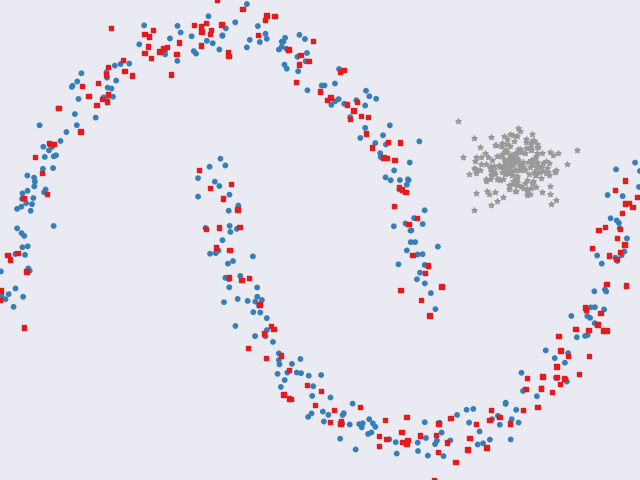}
         \end{subfigure} 
         \begin{subfigure}[b]{0.19\linewidth}
                 \centering
                 \includegraphics[width=0.99\textwidth]{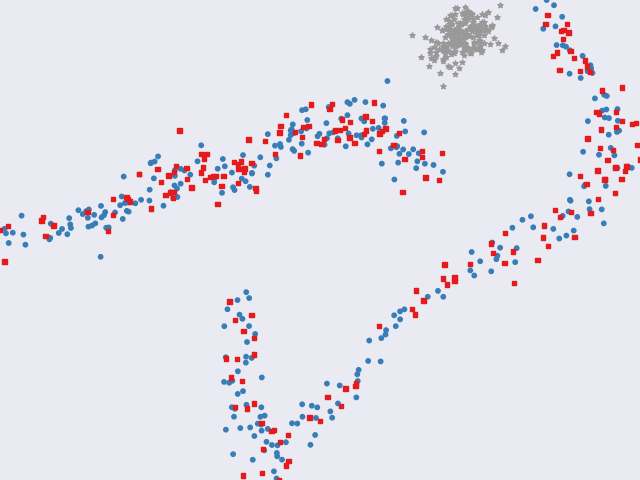}
         \end{subfigure}
          \begin{subfigure}[b]{0.19\linewidth}
                 \centering
                 \includegraphics[width=0.99\textwidth]{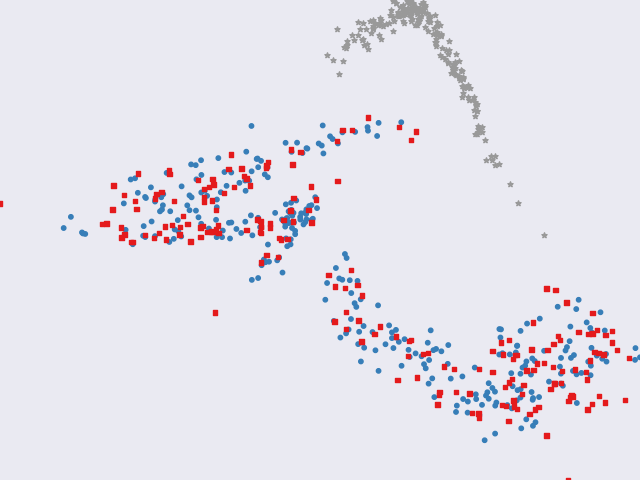}
         \end{subfigure} 
        \begin{subfigure}[b]{0.19\linewidth}
                 \centering
                 \includegraphics[width=0.99\textwidth]{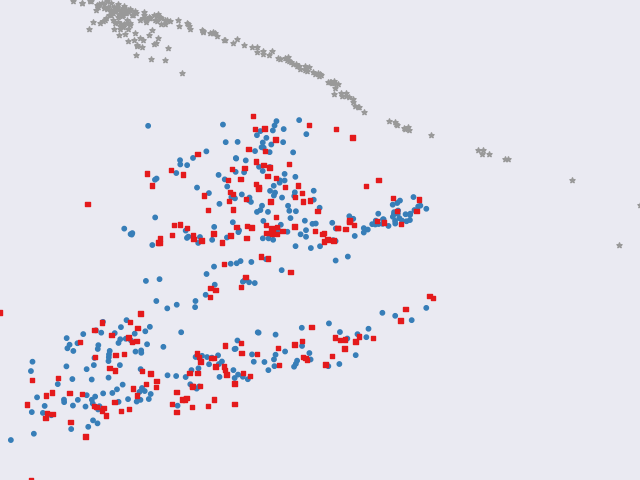}
         \end{subfigure} 
\caption{(a) The imputed values for the red points by TDM, corresponding to the ground truth of Figure~\ref{fig-motivation-1}(a). (b) $\mX$ with OOD noises (grey points). (c-e) The transformed points by different blocks of $f_\theta$.}
\label{fig-motivation-2}
 \vspace{-0.5cm}
\end{figure*}

The above optimisation is straightforward, however, simply minimising the Wasserstein loss can lead to \textit{model collapsing}, meaning that no matter what the input sample is, $f_\theta$ always transforms it into the same point in the latent space. It is easy to see that model collapsing is the trivial solution that minimises the Wasserstein distance between any two samples (or batches) (i.e., to be zero), regardless of how the missing values are imputed.
To prevent model collapsing, we are inspired by the viewpoint of representation learning, where our method can be viewed to learn the latent representation $\vz$ from the data sample $\vx$ with missing values in an unsupervised way.
Representation learning based on mutual information (MI) has been shown promising in several domains~\cite{oord2018representation,bachman2019learning,hjelm2018learning,tschannen2019mutual}, where it has been motivated by the InfoMax principle~\cite{linsker1988self}.
To avoid model collapsing, we propose to add a constraint to $f_\theta$ 
such that it also  %
maximises the MI $I\left(\mX, f_\theta(\mX)\right)$.  %
Thus, we define
\begin{equation}\label{eq-proj-loss-mi}
     \mathcal{L}^{\text{MI}}(\mX) = - I\left(\mX, f_\theta(\mX)\right),
\end{equation}
and aim to learn $\theta$ by minimising both $\mathcal{L}^{W}(\mX^1, \mX^2)$ and 
$\mathcal{L}^{\text{MI}}(\mX^1)+\mathcal{L}^{\text{MI}}(\mX^2)$.

Estimating the above MI in high-dimensional spaces has been known as a difficult task~\cite{tschannen2019mutual}. Although several methods have been proposed to approximate the estimation, e.g., in~\citet{oord2018representation},  they may inevitably add significant complexity and parameters to our method.
Instead of maximising a tractable lower bound as in \citet{oord2018representation,poole2019variational}, we propose a simpler approach that constrains $f_\theta$ to be a smooth invertible map.

\begin{proposition}\label{thm:infomax}
If $f_\theta$ is a smooth invertible map, then 
$f_\theta\in\argmax_{f'} I\left(\mX, f'(\mX)\right)$.
\end{proposition}
\begin{proof}
By definition, we have $I(\mX, f'(\mX)) = H(\mX) - H(\mX | f'(\mX))$ where $H(\mX)$ and $H(\mX | f'(\mX))$ are the entropy and conditional entropy, respectively. As we consider $\mX$ and $f'(\mX)$ as empirical random variables with finite supports of their samples, $H(\mX | f'(\mX)) \ge 0$. Therefore, $I(\mX, f'(\mX)) \le  H(\mX) = I(\mX, \mX)$. If  $f_\theta$ is a smooth invertible map, it is known that: 
$I(\mX, f_\theta(\mX)) = I(\mX, \mX)$ according to Eq.~(45) of~\citet{kraskov2004estimating}. Therefore, $I(\mX, f_\theta(\mX)) \ge I(\mX, f'(\mX))$.
\end{proof}

Proposition~\ref{thm:infomax} shows that $f_\theta$ being invertible\footnote{One may also say that $f_\theta$ is bijective or $f_\theta$ is a diffeomorphism~\cite{papamakarios2021normalizing}.} (i.e., $f_\theta$ projects $\vx$ to $\vz$ and its inverse function $f_\theta^{-1}$ projects $\vz$ back to $\vx$) prevents model collapsing, without explicitly maximising the mutual information.
Accordingly, we implement $f_\theta$ with invertible neural networks (INNs)~\citep{dinh2014nice,dinh2016density,kingma2018glow}, which are approximators to invertible functions~\citep{jacobsen2018excessive,gomez2017reversible,ardizzone2018analyzing,kobyzev2020normalizing,papamakarios2021normalizing}.
Specifically, the INNs for $f_\theta$ consists of a succession of $T$ blocks, $f_\theta=f_1 \circ f_2 \circ \cdots f_T$, each of which is an invertible function\footnote{As $f_1,\dots,f_T$ are invertible, so is $f_\theta$~\citep{kobyzev2020normalizing}}.
Note that now we have $D'=D$, meaning that the output and input dimensions are the same for $f_\theta$ and every $f_t~(1\le t \le T)$. 

For one block $f_t$, whose input and output vectors are denoted as $\vy^{\text{in}} \in \mathbb{R}^D$ and $\vy^{\text{out}}\in \mathbb{R}^D$ respectively, we implement it as an affine coupling block by following~\citet{ardizzone2018analyzing}, which consists of two complementary affine coupling layers~\citep{dinh2016density}:
\begin{align}
    &\vy^{\text{out}}_{1:d} = \vy_{1:d}^{\text{in}} \odot \exp\left(g_1(\vy_{d+1:D}^{\text{in}})\right) + h_1(\vy_{d+1:D}^{\text{in}}),\label{eq:dinh1}\\
    &\vy_{d+1:D}^{\text{out}} = \vy_{d+1:D}^{\text{in}} \odot \exp\left(g_2(\vy_{1:d}^{\text{out}})\right) + h_2(\vy_{1:d}^{\text{out}}),\label{eq:dinh2}
\end{align}
where $\vy^{\text{in}}$ and $\vy^{\text{out}}$ are 
decomposed into two disjoint subsets, respectively: $\vy^{\text{in}} = [\vy^{\text{in}}_{1:d}, \vy^{\text{in}}_{d+1:D}]$ and $\vy^{\text{out}} = [\vy^{\text{out}}_{1:d}, \vy^{\text{out}}_{d+1:D}]$; $d$ is set to $\lfloor D/2 \rfloor$; $\odot$ denotes the element-wise product.
Moreover, $g_1$, $g_2$, $h_1$ and $h_2$ are neural networks, each of which is implemented by a succession of fully connected layers with the SELU activation~\citep{selu}. In addition, the output of $g_1$ and $g_2$ is clamped by the $\arctan$ function.
The implementation ensures that $f_t$ is invertible~\citep{dinh2016density} 
as proved in Section~\ref{sec-invertibility} of the appendix.
It is noticeable that our method is agnostic to the implementation of INNs and other coupling layers such as NICE~\citep{dinh2014nice} and GLOW~\citep{kingma2018glow} can also be used as drop-in replacements.

\subsection{When TDM Is Better?}
We believe that TDM outperforms OTImputer when the data exhibit complex geometry. 
To demonstrate this, 
Figure~\ref{fig-motivation-2}(a) shows the imputation of TDM on the same data shown in Figure~\ref{fig-motivation-1}, where it can be observed that the imputed values of TDM (with three blocks, i.e., $T=3$) align with the data distribution significantly better than OTImputer.
To demonstrate the learned transformed spaces,
we feed the data with out-of-distribution (OOD) samples generated from a two-dimensional normal distribution shown in Figure~\ref{fig-motivation-2}(b) into the learned $f_\theta$ of TDM with a succession of three blocks. Note that TDM is trained without these OOD noises.
Figures~\ref{fig-motivation-2}(c-e) show the latent space after the first ($f_{1}(\mX)$), second ($f_{1:2}(\mX)$), and third (final) ($f_{1:3}(\mX)$) block, respectively. In the latent spaces, the in-domain samples are close to each other, and are well separated from the OOD samples. 
That explains why TDM does not have the false imputations as OTImputer, because the OOD samples are far away from the in-domain ones in terms of the quadratic distance in the latent spaces.
This also demonstrates that TDM does not simply push all the points in the data space close to each other and model collapsing is avoided. 
In Figure~\ref{fig-motivation-44} of the appendix, we show two additional synthetic datasets, whose geometry is simpler than the previous ones. In these simpler cases, one can see that OTImputer is able to correctly impute the missing values as TDM. This is because the quadratic distance in the data space used in OTImputer can capture the data geometry well. In these cases, TDM does not have to learn a complex series of transformations. Therefore, the transformed spaces of TDM look similar to the data space, i.e., 
 is close to an isometry, which leads TDM to reduce to OTImputer.

\subsection{Implementation Details}
As discussed before, we only need to minimise the Wasserstein distance without explicitly maximising the mutual information, which requires $f_\theta$ to be invertible, denoted as $f_\theta \in \mathcal{F}^{\text{INN}}$. Our final loss then becomes:
\begin{align}
\label{eq-proj-loss-2}
    &\min_{\mX^{1\cup 2}[\mM^{1\cup 2}], \theta} \mathcal{L}^{W} \text{~s.t.~} f_\theta \in \mathcal{F}^{\text{INN}},
\end{align}
where $\theta$ consists of parameters of the neural networks $g_1$, $g_2$, $h_1$, $h_2$ used in every block of $f_\theta$.

\textbf{Computation of Wasserstein Distances~}
The exact computation of Wasserstein distances can be done by network simplex methods that take $\mathcal{O}(D^3)$ ($D$ is the feature dimension of $\mX^1$ and $\mX^2$)~\citep{ahuja1995applications}.
\citet{muzellec2020missing} uses Sinkhorn iterations~\citep{cuturi2013sinkhorn} with the entropic regularisation to compute the 2-Wasserstein distances in $\mathcal{O}(D^2 \log{D})$~\citep{altschuler2017near,dvurechensky2018computational}:
$\hat{d}_{\mG}\left(\mu(\mX^1), \mu(\mX^2)\right) 
\defeq d_{\mG}\left(\mu(\mX^1), \mu(\mX^2)\right) + \epsilon r(\mP)$, where $r(\mP)$ is the negative entropy of the transport plan and
$\epsilon$ is set in an ad-hoc manner: 5\% of the median distance between initialised values in each dataset.
We find that $\epsilon$ is critical to the imputation results and the ad-hoc setting may achieve sub-optimal performance. 
To avoid selecting $\epsilon$, instead of Sinkhorn iterations, we use the network simplex method~\citep{bonneel2011displacement} efficiently implemented in the POT package~\cite{flamary2021pot}.
Theoretically, the network simplex method has $\mathcal{O}(D^3)$ complexity, however, \citet{bonneel2011displacement} reports it behaves in $\mathcal{O}(D^2)$ in practice. 
Unlike $W_2$ used in TDM, 
neither $\hat{d}_{\mG}$ nor the Sinkhorn divergence~\cite{genevay2018learning} used in~\citet{muzellec2020missing} is guaranteed to be
a metric distance.

\textbf{Computation of Gradients~}
Recall that in Eq.~(\ref{eq-def-ot}) OT/Wasserstein distances are computed by finding the optimal transport plan $\mP \in \mathbb{R}_{>0}^{B \times B}$ ($B$ is the batch size in Eq.~(\ref{eq-proj-loss-2})). 
Given $\mP$, if a quadratic cost function is used, the gradient of $\mathcal{L}^W$ in terms of $f_\theta(\mX^1[i,:])$ ($\mX^1[i,:]$ is the $i^{\text{th}}$ sample of $\mX^1$) is~\cite{cuturi2014fast,muzellec2020missing} :
\begin{align}
   \frac{\partial \mathcal{L}^W}{\partial f_\theta( \mX^1[i,:])} = \sum_{j=1}^B \mP[i,j] \left(f_\theta( \mX^1[i,:]) - f_\theta(\mX^2[j,:])\right)\nonumber,
\end{align} with which, one can use backpropagation to update $\theta$ and the missing values in $\mX^1[i,:]$.
The algorithm of our proposed method is shown in Algorithm~\ref{alg}.

\begin{figure}[t]
  \centering
  \begin{minipage}{0.48\textwidth}
    \centering
\removelatexerror
\begin{algorithm2e}[H]
\SetKwInOut{Input}{input}\SetKwInOut{Output}{output}
\Input{Data $\mX$ with missing values indicated by $\mM$}
\Output{$\mX$ with $\mX[\mM]$ imputed, $f_\theta$}
Initialise $\theta$ of $f$\;

\#~\textit{Initialise missing values with noisy mean}~\#

$\mX[\mM] \leftarrow \text{nanmean}(\mX, \text{dim=0}) + \mathcal{N}(0, 0.1)$ \;

\While{Not converged}
{
	Sample two batches of $B$ data samples $\mX^1$ and $\mX^2$;
 
	Feed $\mX^1$ and $\mX^2$ to $f_\theta$;
 
	\For{$i=1\dots B$, $j=1\dots K$}
	{
	    Compute $\mG'[i,j]$; \#~\textit{Quadratic cost function}~\#
	}
	Compute $\mathcal{L}^W$;
 
	Update the missing values $\mX^{1\cup 2}[\mM^{1\cup 2}]$ and $\theta$ with gradient update;
}
\caption{TDM. Learnable parameters include missing values $\mX[\mM]$ and parameters $\theta$ of $f$.}
\label{alg}
\end{algorithm2e}
\end{minipage}
\vspace{-3mm}
\end{figure}

\begin{figure*}[t]
\captionsetup[subfigure]{justification=centering}
        \centering
         \begin{subfigure}[b]{0.99\linewidth}
                 \centering
                 \includegraphics[width=0.99\textwidth]{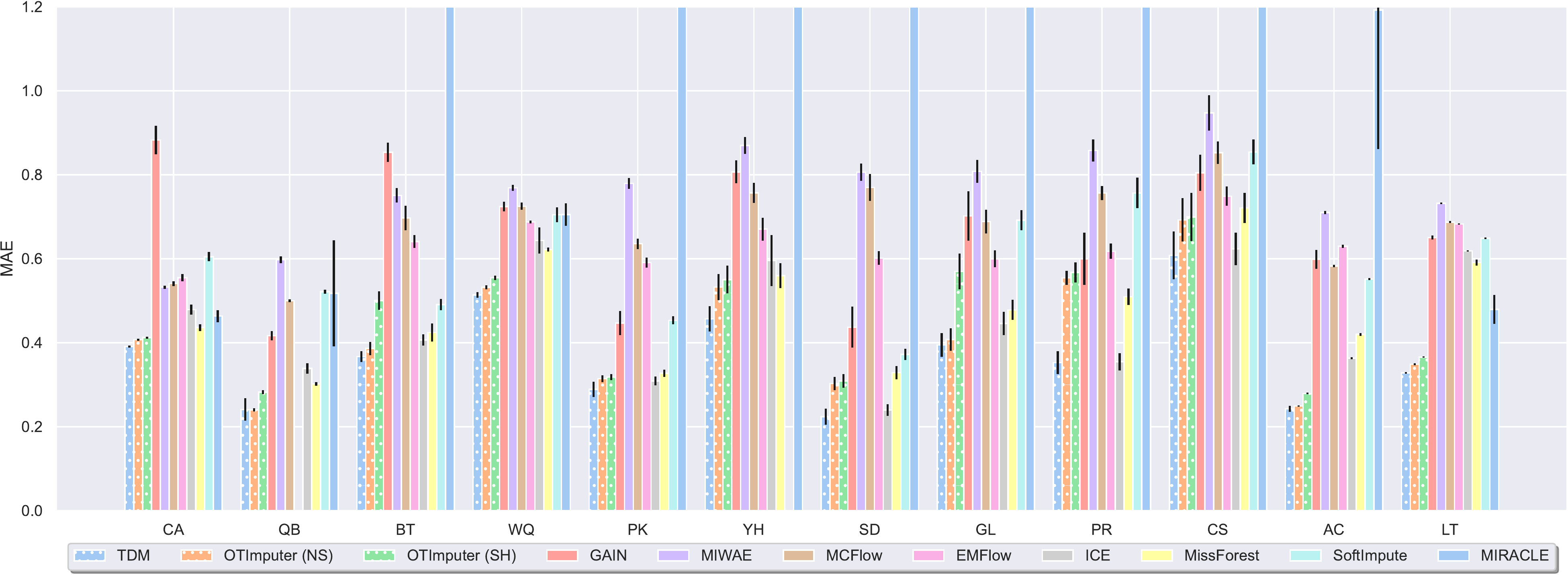}
         \end{subfigure}
         \\
         \begin{subfigure}[b]{0.99\linewidth}
                 \centering
                 \includegraphics[width=0.99\textwidth]{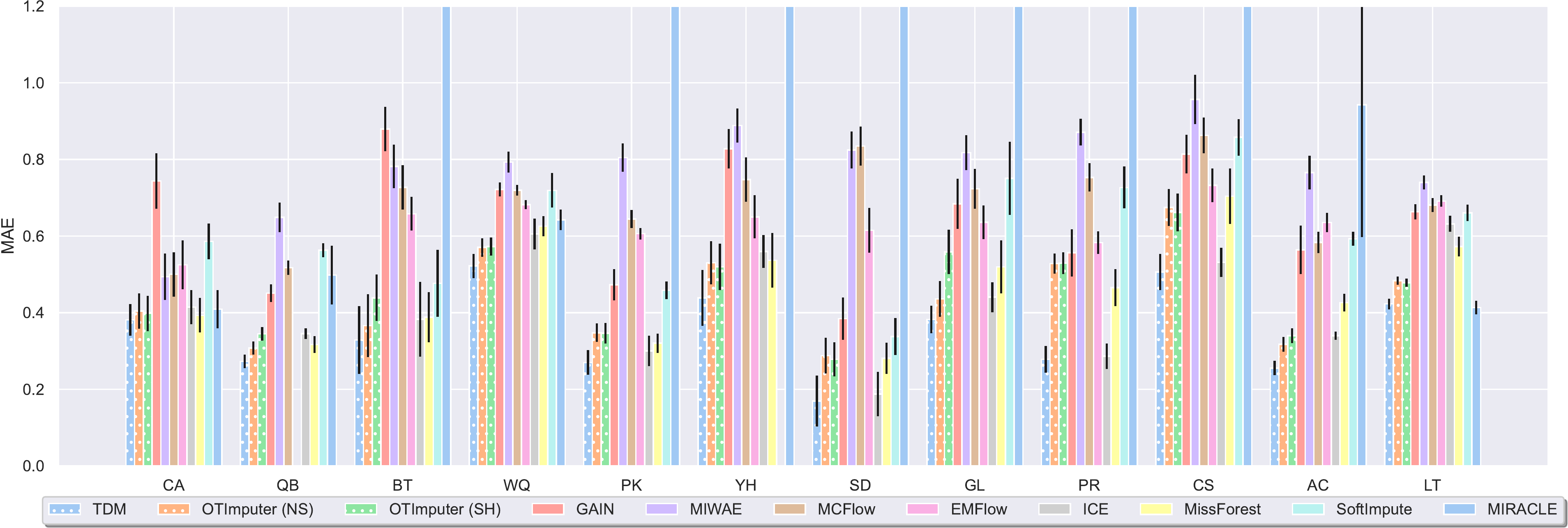}
         \end{subfigure}
\centering
\caption{MAE in the MCAR (top) and MAR (bottom) settings.}
\label{fig-mae-mcar-mar}
\end{figure*}

\section{Related Work}
As missing values are ubiquitous in many domains, missing value imputation has been an active research area~\citep{little2019statistical,mayer2019r}.
Existing methods can be categorised differently from different aspects~\citep{muzellec2020missing} such as the type of the missing variables (e.g., real, categorical, or mixed) the mechanisms of missing data (e.g., MCAR, MAR, or MNAR discussed in Section~\ref{sec-bg-missing}).
In this paper, we focus on imputing real-valued missing data.
Besides simple baselines such as imputation with mean/median/most-frequent values,
we introduce the related work from the perspective of whether a method treats data features (i.e., the columns in data $\mX \in \mathbb{R}^{N \times D}$) separately or jointly, following~\citet{jarrett2022hyperimpute}.

Methods in the former category estimate the distributions of one feature conditioned on the other features and perform iterative imputations for one feature at a time, such as in~\citet{heckerman2000dependency,raghunathan2001multivariate,gelman2004parameterization,van2006fully,van2011mice,liu2014stationary,zhu2015convergence}.
As each feature's conditional distribution may be different, these methods need to specify different models for them, which can be cumbersome in practice, especially when the missing values are unknown.

In the latter category, methods learn a joint distribution of all the features explicitly or implicitly. To do so, various methods have been proposed, 
such as the ones based on matrix completion~\citep{mazumder2010spectral}, (variational) autoencoders~\citep{gondara2017multiple,ivanov2018variational,mattei2019miwae,nazabal2020handling,gong2021variational,peis2022missing}, generative adversarial nets~\citep{yoon2018gain,li2018misgan,yoon2020gamin,dai2021multiple,fang2022fragmgan}, graph neural networks~\citep{you2020handling,vinas2021graph,chen2022gedi,huang2022graph,morales2022simultaneous,gao2022handling}, normalising flows~\citep{richardson2020mcflow,ma2021emflow,wang2022generative}, and Gaussian process~\citep{dai2022multiple}. 

In addition to these two categories, several recent work proposes general refinements to existing imputation methods. For example, \citet{wang2021missingness} introduces data augmentation methods to improve generative methods such as those in ~\citet{yoon2018gain,nazabal2020handling,richardson2020mcflow}.
\citet{kyono2021miracle} proposes causally-aware~\citep{mohan2013graphical} refinements to existing methods. Recently, \citet{jarrett2022hyperimpute} proposes to automatically select an imputation method among multiple ones for each feature, which shows improved results over individual methods. Our method is a stand-alone approach and many of the above refinements can be applied to ours as well such as the data augmentation and causally-aware methods.

Among the above works, the closest one to ours is OTImputer~\cite{muzellec2020missing}, whose differences from ours have been comprehensively discussed. \citet{muzellec2020missing} also introduces a parametric version of OTImputer trained in a round-robin fashion. The parametric algorithm does not work as well as the standard OTImputer and our method can be easily extended with the parametric algorithm if needed.
Methods based on normalising flows, e.g., MCFlow~\cite{richardson2020mcflow} and EMFlow~\cite{ma2021emflow} also use INNs for imputation. However, there are fundamental differences of TDM to them, the most significant one of which is that MCFlow and EMFlow can still be viewed as deep generative models using INNs as the encoder and decoder and their losses are still reconstruction losses or maximum likelihood in the data space, while ours uses a matching distance in the latent space as the loss. 
Going beyond missing value imputations, a recent work by \citet{coeurdoux2022swotf} proposes to learn sliced-Wasserstein distances with normalising flows, which we do not consider as a close related work to ours as the primary goal, motivation, and methodology are different.

\section{Experiments}

\begin{figure*}[t]
\captionsetup[subfigure]{justification=centering}
        \centering
         \begin{subfigure}[b]{0.99\linewidth}
                 \centering                 \includegraphics[width=0.99\textwidth]{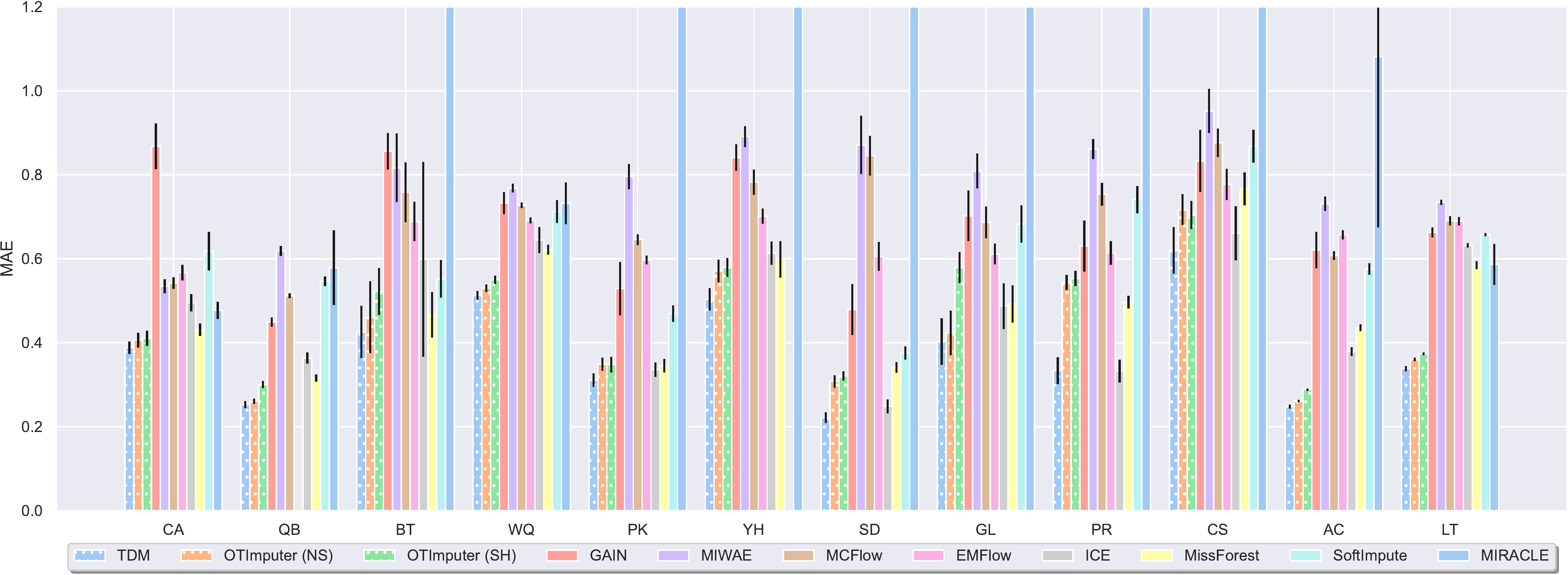}
         \end{subfigure}
                          \\
         \begin{subfigure}[b]{0.99\linewidth}
                 \centering            \includegraphics[width=0.99\textwidth]{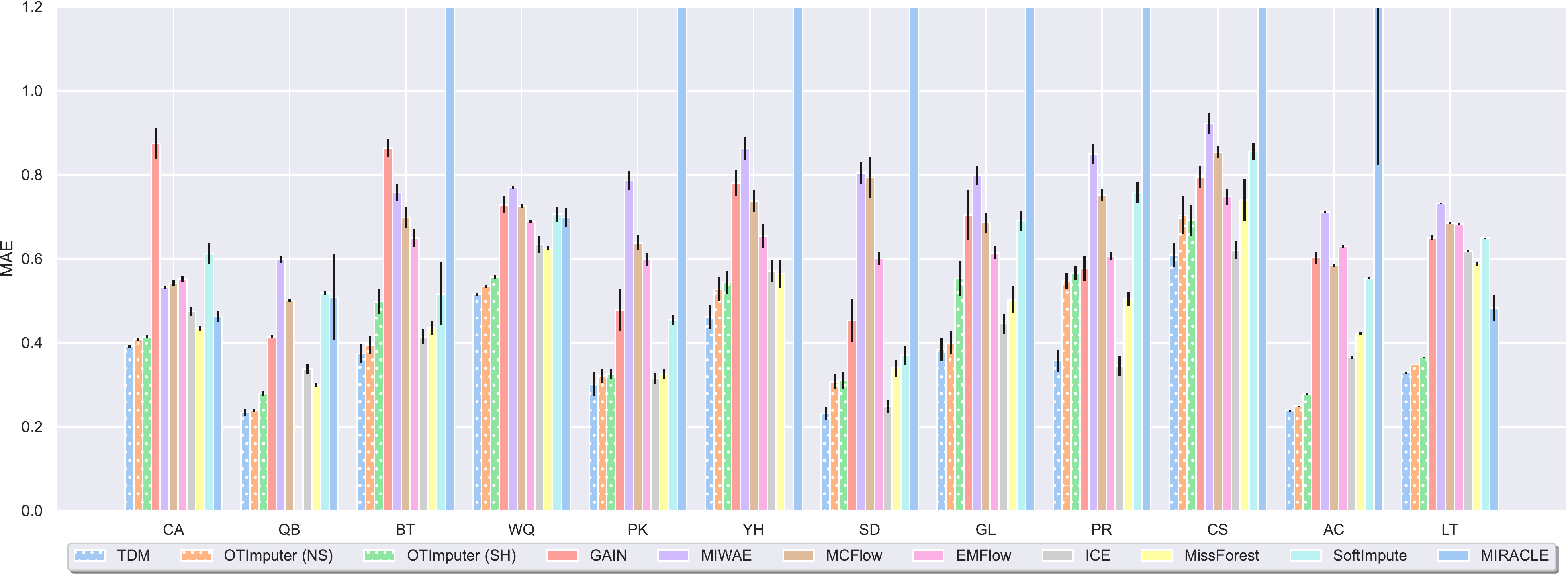}
         \end{subfigure}
 \caption{MAE in the MNARL (top) and MNARQ (bottom) settings.}
  \label{fig-mae-mnar}
\end{figure*}

\subsection{Experimental Settings}
\textbf{Datasets~}
Similar to many recent works~\citep{yoon2018gain,mattei2019miwae,muzellec2020missing,jarrett2022hyperimpute}, UCI datasets\footnote{\url{https://archive-beta.ics.uci.edu}} with different sizes are used in the experiments, the statistics of which are shown in Table~\ref{tb-ds} of the appendix. Each dataset is standardised by the scale function of sklearn\footnote{\url{https://scikit-learn.org/stable/modules/generated/sklearn.preprocessing.scale.html}}.
Following~\citet{muzellec2020missing,jarrett2022hyperimpute}, we generate the missing value mask for each dataset with three mechanisms in four settings, 
which, to our knowledge, include all the cases used in the literature.
Specifically, for MCAR, we generate the mask for each data sample by drawing from a Bernoulli random variable with a fixed parameter.
For MAR, we first sample a subset of features (columns in $\mX$) that will not contain missing values and then we use a logistic model with these non-missing columns as input to determine the missing values of the remaining columns and we employ line search of the bias term to get the desired proportion of missing values.
Finally, we also generate the masks with MNAR in two ways: 1) MNARL: Using a logistic model with the input masked by MCAR; 2) MNARQ: Randomly sampling missing values from the range of the lower and upper $p^{\text{th}}$ percentiles.
For each of the four settings, we use 30\% missing rate, sample 10 masks for one dataset with different random seeds~\cite{jarrett2022hyperimpute}, and report the mean and standard deviation (std) of the corresponding performance metric.

\textbf{Evaluation Metrics~}
We evaluate the performance of a method by examining how close its imputation is to the ground-truth values, which is measured by the mean absolute error (MAE) and the root-mean-square error (RMSE).
Following~\citet{muzellec2020missing}, we also use the 2-Wasserstein distance, $W^2_2$, between the imputed and the ground-truth distributions in the data space. Here $W^2_2$ is similar to the loss of OTImputer in Eq.~(\ref{eq-proj-loss-2}) and the difference is $W^2_2$ as a metric is computed over all the imputed and ground-truth samples\footnote{Therefore, $W^2_2$ cannot be reported if the number of data points is too large.} while OTImputer minimises $W^2_2$ between two sampled batches. 
For all the three metrics, lower values indicate better performance.
Although we do not assume a specific downstream task, we conduct the evaluations of classification on the imputed data of different methods, whose settings are as follows:
\textbf{1)} We remove the datasets that do not have labels (e.g., parkinsons) or have binary labels (e.g., letter. In this case, the classification performance is almost the same regardless of how the missing values are imputed).
\textbf{2)} After the missing values are imputed for a dataset, we train a support vector machine with the RBF kernel and auto kernel coefficient. We report the average accuracy of 5-fold cross-validations. \textbf{3)} We report the mean and std of average accuracies in 10 runs of an imputation method with different random seeds.

\begin{figure*}[t]
\captionsetup[subfigure]{justification=centering}
        \centering
         \begin{subfigure}[b]{0.99\linewidth}
                 \centering                 \includegraphics[width=0.99\textwidth]{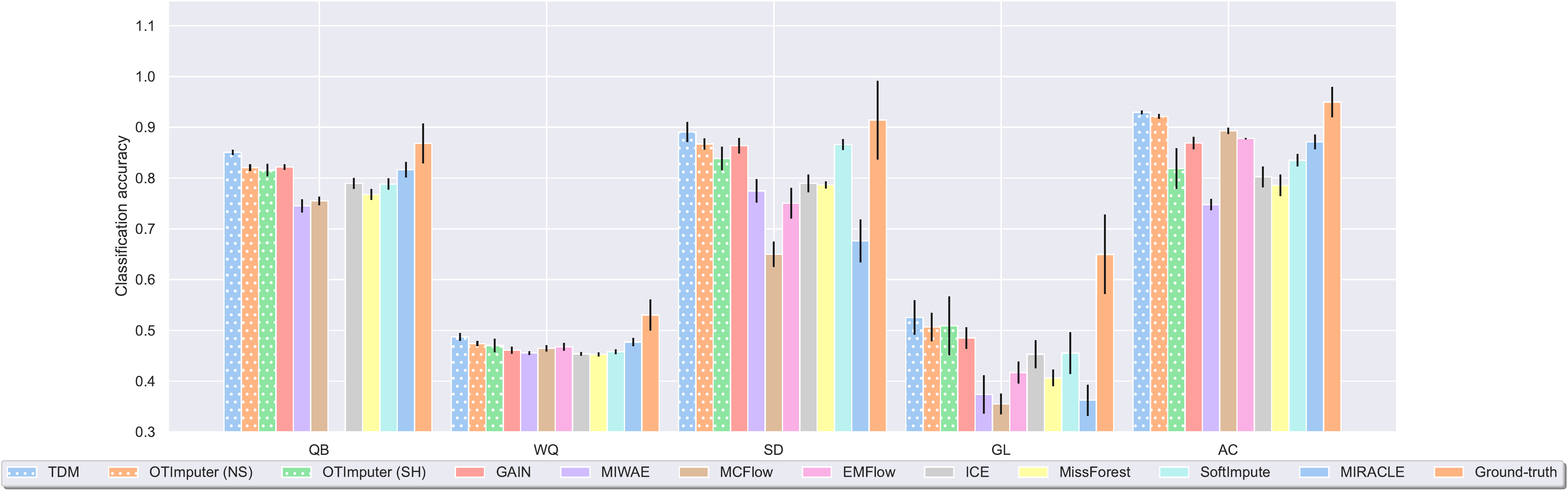}
         \end{subfigure}
 \caption{Classification accuracy in the MCAR setting.}
  \label{fig-cla-mcar}
 \vspace{-0.2cm}
\end{figure*}

\textbf{Settings of Our Method~}
To minimise the loss in Eq.~(\ref{eq-proj-loss-2}), we use RMSprop~\cite{tieleman2012lecture} as the optimiser with learning rate of $10^{-2}$
and batch size of 512\footnote{If the number of data samples $N$ is less than 512, we use $2^{\lfloor N/2 \rfloor}$, following~\citet{muzellec2020missing}.}.
Due to the simplicity of our method, there are only two main hyperparameters to set in terms of the architecture of the transformation.
The first one is the number of INN blocks $T$.
For the affine coupling layers~\cite{dinh2016density} in each transformation, we use three fully connected layers with SELU activation for each of $h_1$, $h_2$, $g_1$, and $g_2$. The size of each fully connected layer is set to $K\times{}D$ where $D$ is the feature dimension of the dataset and $K$ is the second hyperparameter. We empirically find that $T=3$ and $K=2$ work well in practice and show the hyperparameter sensitivity in Appendix~\ref{sec-ps}.
We train our method for 10,000 iterations and report the performance based on the last iteration, which is the same for all the OT-based methods.

\textbf{Baselines~}
We compare our method against four lines of ten baselines.
\textbf{1)} Iterative imputation methods: ICE~\citep{van2011mice} with linear/logistic models used in~\citet{muzellec2020missing,jarrett2022hyperimpute}; MissForest with random forests~\citep{stekhoven2012missforest}.
\textbf{2)} Deep generative models:
GAIN~\cite{yoon2018gain}\footnote{\url{https://github.com/jsyoon0823/GAIN}} using generative adversarial networks~\cite{goodfellow2020generative} where
the generator outputs the imputations and 
the discriminator classifies the imputations in an element-wise fashion; 
MIWAE~\cite{mattei2019miwae}\footnote{\url{https://github.com/pamattei/miwae}} extending the importance weighted autoencoders~\cite{burda2016importance} for missing value imputation;
MCFlow~\cite{richardson2020mcflow}\footnote{\url{https://github.com/trevor-richardson/MCFlow}} and 
EMFlow~\cite{ma2021emflow}\footnote{\url{https://openreview.net/attachment?id=bmGLlsX_iJl&name=supplementary_material}} extending normalising flows for imputation.
\textbf{3)} Methods based on OT: OTImputer (SH), the original implementation of OTImputer~\citep{muzellec2020missing}\footnote{\url{https://github.com/BorisMuzellec/MissingDataOT}} where OT distances are computed by Sinkhorn iterations; OTImputer (NS), the same to OTImputer (SH) except that the OT distances are computed by the network simplex methods with POT~\cite{flamary2021pot}.
\textbf{4)} Other methods: SoftImpute~\cite{hastie2015matrix} using matrix completion and low-rank SVD for imputation; 
MIRACLE~\cite{kyono2021miracle}\footnote{\url{https://github.com/vanderschaarlab/MIRACLE}} introducing causal learning as a regulariser to refine imputations.
For ICE and MissForest, we use the implementations in sklearn~\cite{pedregosa2011scikit}\footnote{\url{https://scikit-learn.org/stable/modules/impute.html}}
For SoftImpute, we use the implementation of~\citet{jarrett2022hyperimpute}, which follows the original implementation.
For the other methods, we use their original implementations (links of code listed above) with the best reported settings.

\subsection{Results}
\label{sec-results}
Now we show the MAE results\footnote{The empty results are due to the failure of running the code.} in the four missing value settings in Figures~\ref{fig-mae-mcar-mar} and~\ref{fig-mae-mnar}. The results of RMSE and $W^2_2$ are shown in Figures~\ref{fig-rmse} and~\ref{fig-ot} of the appendix.
From these results, it can be observed that our proposed method, TDM, consistently achieves the best results in  comparison with others in almost all the settings, metrics, and datasets.
Specifically, for OT-based methods, we can see that one may gain marginal yet consistent improvement in most cases by using the network simplex methods to compute the OT distance in the comparison between OTImputer (NS) and OTImputer (SH).
With the help of the learned transformations, TDM significantly outperforms both OT methods. 
Note that shown in Eq.~(\ref{eq-ot-loss}), OTImputer directly minimises the metric of $W^2_2$ between two batches in the data space. Alternatively, TDM minimises the Wasserstein distance in the transformed space.
Interestingly, although TDM does not directly minimises $W^2_2$ in the data space, it outperforms OTImputer on $W^2_2$ (shown in the appendix).
In the comparison with MCFlow and EMFlow that also use INNs as ours, their performance is not as good as TDM's.

\begin{figure}[t]
\captionsetup[subfigure]{justification=centering}
        \centering
         \begin{subfigure}[b]{0.49\linewidth}
                 \centering
                 \caption{glass}
                 \includegraphics[width=0.99\textwidth]{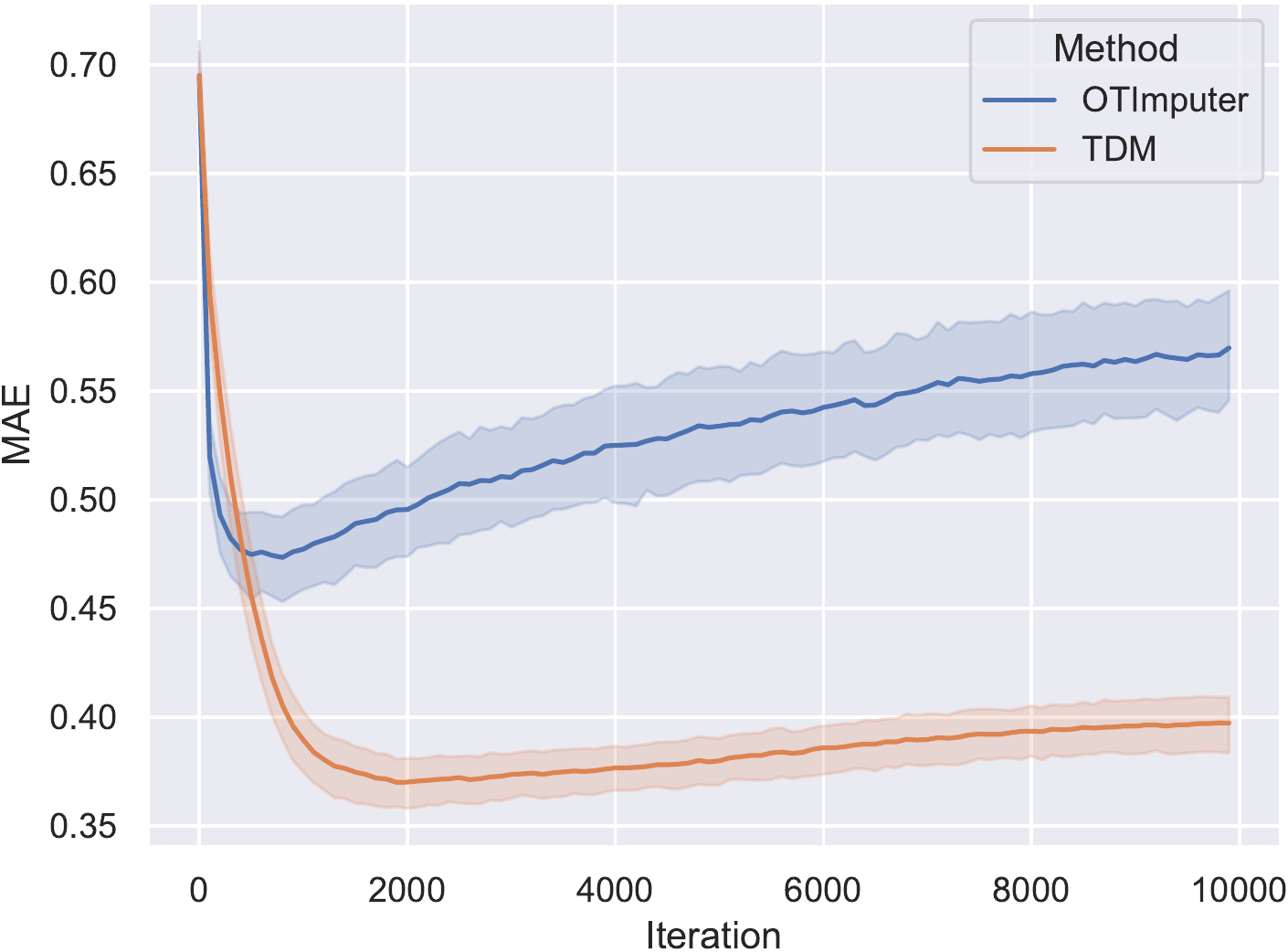}
         \end{subfigure}
                 \begin{subfigure}[b]{0.49\linewidth}
                 \centering
                 \caption{seeds}
                 \includegraphics[width=0.99\textwidth]{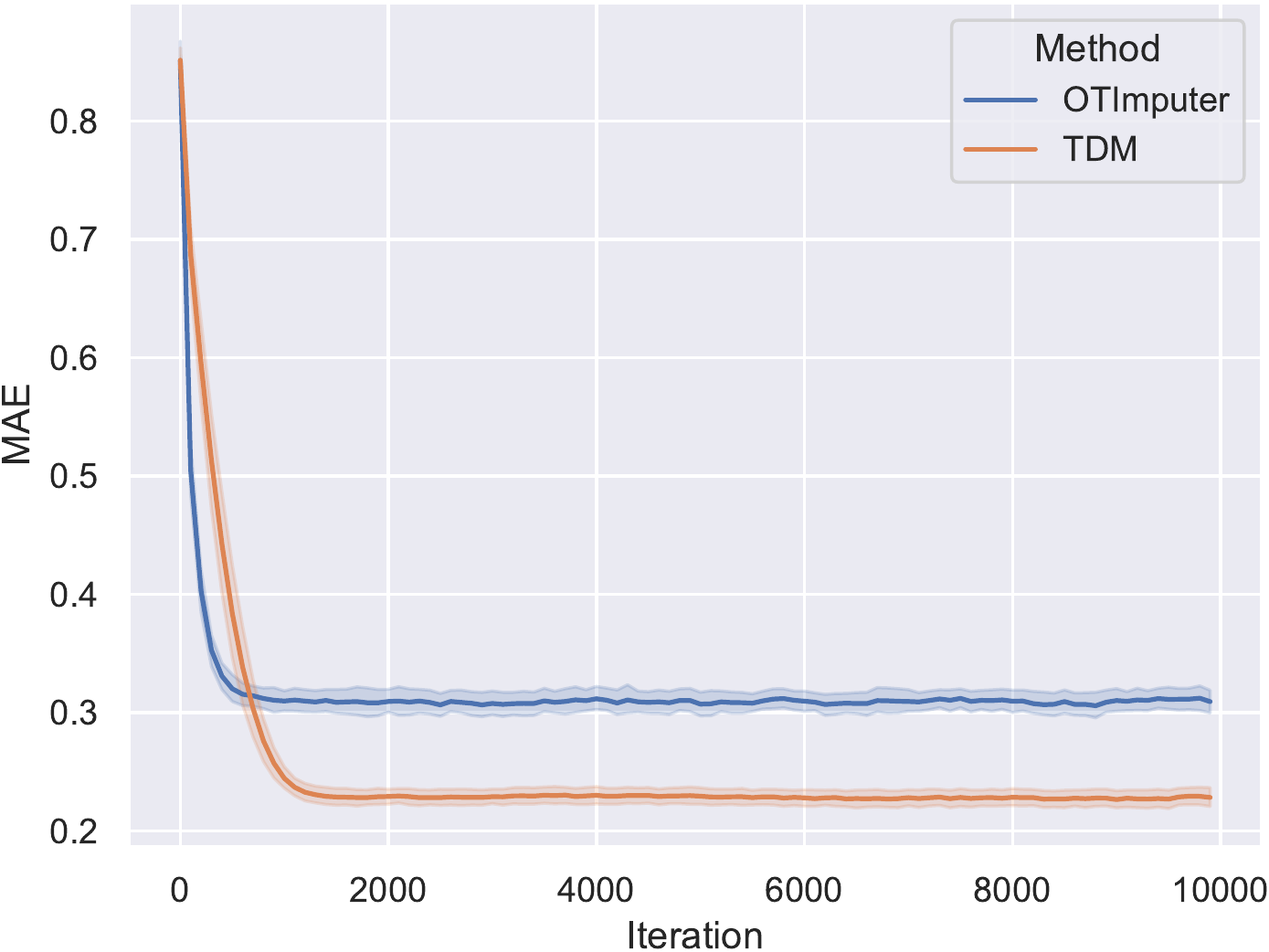}
         \end{subfigure}\\
          \begin{subfigure}[b]{0.49\linewidth}
                 \centering  
                 \caption{blood\_transfusion}
                 \includegraphics[width=0.99\textwidth]{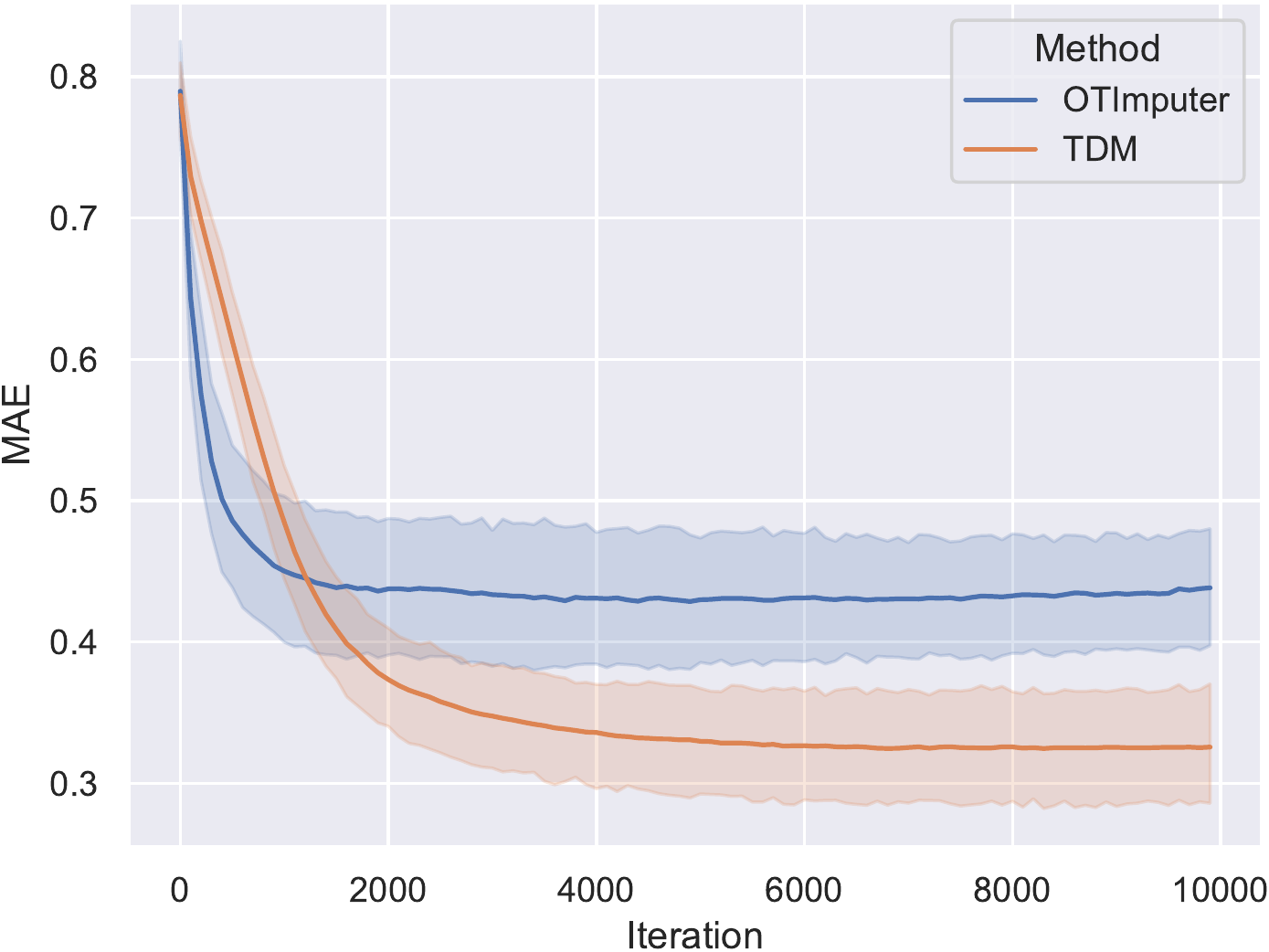}
         \end{subfigure}
                  \begin{subfigure}[b]{0.49\linewidth}
                 \centering
                 \caption{anuran\_calls}
                 \includegraphics[width=0.99\textwidth]{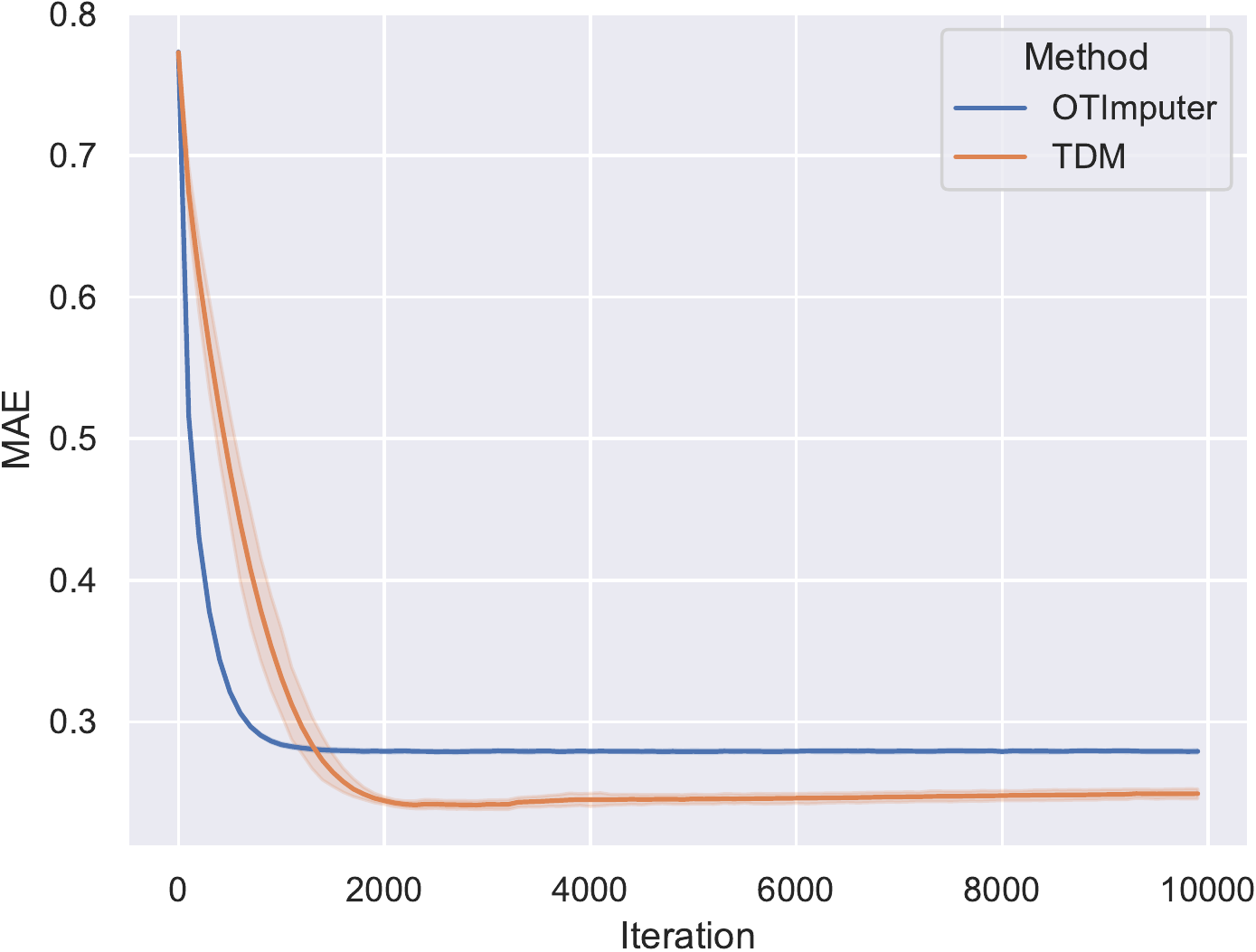}
         \end{subfigure}
 \caption{MAE over training iterations of TDM and OTImputer on four datasets in MCAR. The results are averaged over 10 runs.}
  \label{fig-iter-mae-mcar}
 \vspace{-0.5cm}
\end{figure}

Figure~\ref{fig-cla-mcar} shows the classification accuracy with the imputed data by different approaches in the MCAR setting (the other settings are shown in Figure~\ref{fig-cla} of the appendix). We also report the accuracy on the ground-truth data as a reference. From the results, it can be seen that TDM in general performs the best in the classification task, showing that good imputations do help with downstream tasks. 

Figure~\ref{fig-iter-mae-mcar} shows the MAE over the training iterations of TDM and OTImputer on four datasets in the MCAR settings (the other metrics and settings are shown in Figures~\ref{fig-iter-mcar}, ~\ref{fig-iter-mar},~\ref{fig-iter-mnar-l}, and~\ref{fig-iter-mnar-q} of the appendix). It can be observed that TDM converges slightly slower than OTImputer as a function of the number of iterations, as TDM additionally learns the deep transformations. The average running time (seconds) per iteration for the two methods in the same computing environment is as follows: glass:
OTImputer (1.14), TDM (3.20); seeds: OTImputer (1.03), TDM (3.19); blood\_transfusion: OTImputer (2.39), TDM (3.35); anuran\_calls: OTImputer (2.33), TDM (4.06). The running time per iteration of TDM is about 2 to 3 times that of OTImputer. For larger datasets, the running time gap between the two appears to be smaller. In several datasets, e.g., glass, OTImputer exhibits overfitting, while TDM is more stable during training.

\section{Conclusion}
We propose transformed distribution matching (TDM)  for missing value imputation. TDM matches data samples in a transformed space, where the distance of two samples is expected to better reflect their (dis)similarity under the geometry of the data considered. 
The transformations are implemented with invertible neural networks to avoid model collapsing. By minimising the Wasserstein distance between the transformed samples, TDM learns the transformations and imputes the missing values simultaneously. Extensive experiments show that our method significantly improves over previous approaches, achieving state-of-the-art performance. 
The limitations of TDM include: \textbf{1)} Due to the learning of neural networks, TDM is relatively slower than OTImputer. \textbf{2)} TDM (OTImputer as well) does not work with categorical data. 
We leave the development of more efficient and applicable algorithms of TDM to future work.

\bibliography{icml}
\bibliographystyle{icml2023}

\newpage
\appendix
\onecolumn

\section{Theoretical Analysis}\label{sec:analysis}
\subsection{Properties}
In this section, we discuss the basic properties of the proposed imputation method.
Essentially, the loss of TDM is an empirical estimation of
\begin{equation}
    \expect W^2_2(f_{\#}\mu(\bm{X}^1), f_{\#}\mu(\bm{X}^2)),
\end{equation}
where $\expect$ denotes the expectation wrt 
the random mini-batches $\bm{X}^1$ and $\bm{X}^2$
that are sampled independently.
We assume each sample in 
each random mini-batch is sampled independently and identically based on the uniform distribution on $\{1,2,\cdots,N\}$.

\begin{proposition}\label{thm:lb}
\begin{equation*}
    \expect W^2_2(f_{\#}\mu(\bm{X}^1), f_{\#}\mu(\bm{X}^2))
    \ge
    \expect W^2_2(f_{\#}\mu(\bm{X}^1), f_{\#}\mu(\bm{X})),
\end{equation*}
where $f_{\#}\mu(\bm{X})\defeq\frac{1}{N}\sum_{i=1}^N\delta_{f_{\theta}(\xi)}$ is the pushforward empirical measure with respect to all
observed samples in $\bm{X}$.
\end{proposition}
\begin{proof}
We first show that $\mu\to{}W_2^2(\mu,\nu)$ is convex.
Consider $\mu=\lambda\mu_1+(1-\lambda)\mu_2$,
where $\lambda\in(0,1)$.
The optimal transport plan of $\mu_1$ (resp. $\mu_2$)
is $\bm{P}_1$ (resp. $\bm{P}_2$). Then,
$\lambda\bm{P}_1+(1-\lambda)\bm{P}_2$ is a
valid transport plan from $\mu$ to $\nu$.
We have
\begin{align*}
W_2^2(\mu,\nu) 
&\defeq
\inf_{\bm{P}\in\bm{U}(\mu,\nu)} \langle \bm{P}, \bm{G}\rangle\nonumber\\
&\le
\langle \lambda\bm{P}_1 + (1-\lambda)\bm{P}_2, \bm{G} \rangle\nonumber\\
&=
\lambda \langle \bm{P}_1, \bm{G} \rangle
+
(1-\lambda) \langle \bm{P}_2, \bm{G} \rangle\nonumber\\
&=
\lambda W_2^2(\mu_1,\nu)
+(1-\lambda) W_2^2(\mu_2,\nu).
\end{align*}
By Jensen's inequality,
\begin{align*}
    \expect W^2_2(f_{\#}\mu(\bm{X}^1), f_{\#}\mu(\bm{X}^2))
    \ge
    W^2_2(f_{\#}\mu(\bm{X}^1), \expect f_{\#}\mu(\bm{X}^2)).
\end{align*}
Based on our assumption, all samples in $\bm{X}^2$ are 
sampled uniformly, and therefore
\begin{equation*}
\expect f_{\#}\mu(\bm{X}^2)
= \frac{1}{N} \sum_{i=1}^N \delta_{f_{\theta}(\xi)}
\end{equation*}
regardless of the minibatch size $B$. In summary,
\begin{equation*}
    \expect W^2_2(f_{\#}\mu(\bm{X}^1), f_{\#}\mu(\bm{X}^2))
    \ge
    \expect W^2_2(f_{\#}\mu(\bm{X}^1), f_{\#}\mu(\bm{X})).
\end{equation*}
On the LHS the operator $\expect(\cdot)$ is taken with respect to the two random batches $\mX^1$ and $\mX^2$; on the RHS 
$\expect(\cdot)$ is with respect to $\mX^1$.
\end{proof}

The inequality in Proposition~\ref{thm:lb}
holds for some given $\mX$ (with the missing entries imputed) and $f_{\theta}$. As learning goes on, both sides of the inequality change with the missing entries as well as $\theta$,
while the inequality is always valid regardless of how the missing values are set.

Our loss is a surrogate of $\expect W^2_2(f_{\#}\mu(\bm{X}^1),
f_{\#}\mu(\bm{X}))$.
During learning, our imputation method tries to make the local distribution $f_{\#}\mu(\bm{X}^1)$
to be close to the global distribution $f_{\#}\mu(\bm{X})$.
Similar lower- and upper-bounds for $W_1$ (the 1-Wasserstein distance) between empirical measures appeared in~\citet{barthe2013}.

We also have an upper bound of 
$\expect W^2_2(f_{\#}\mu(\bm{X}^1), f_{\#}\mu(\bm{X}^2))$
through the triangle inequality.
\begin{proposition}\label{thm:ub}
$\forall\mu$,
    \begin{equation}
    \mathbb{E} W^2_2(f_{\#}\mu(\bm{X}^1), f_{\#}\mu(\bm{X}^2))
    \le
    4 \expect W_2^2(f_{\#}\mu(\bm{X}^1), \mu).
    \end{equation}
\end{proposition}
\begin{proof}
The 2-Wasserstein distance is a metric distance
and therefore satisfies the triangle inequality. We have
\begin{equation*}
\forall{\mu}, \mX^1, \mX^2,
\quad
W_2^2(f_\#\mu(\mX^1), f_\#\mu(\mX^2))
\le
\left( W_2(f_\#\mu(\mX^1), \mu) + W_2(f_\#\mu(\mX^2), \mu) \right)^2.
\end{equation*}
Take the expectation wrt our random sampling protocol on both sides,
and by noting that $\mX^1$ and $\mX^2$ are sampled independently,
we get
\begin{align*}
&\expect
W_2^2\left(f_\#\mu(\mX^1), f_\#\mu(\mX^2)\right)\nonumber\\
&
\le \expect
\left( W_2(f_\#\mu(\mX^1), \mu) + W_2(f_\#\mu(\mX^2), \mu) \right)^2.
\nonumber\\
&=
\expect W_2^2(f_\#\mu(\mX^1), \mu)
+
\expect W_2^2(f_\#\mu(\mX^2), \mu)
+
2 \expect W_2(f_\#\mu(\mX^1), \mu)
\cdot
\expect W_2(f_\#\mu(\mX^2), \mu)
\nonumber\\
&=
2 \expect W_2^2(f_\#\mu(\mX^1), \mu)
+ 2 \left(\expect W_2(f_\#\mu(\mX^1), \mu)\right)^2.
\end{align*}
As $\left(\expect W_2(f_\#\mu(\mX^1), \mu)\right)^2
\le \expect W_2^2(f_\#\mu(\mX^1), \mu)$, we have
\begin{align*}
\expect
W_2^2\left(f_\#\mu(\mX^1), f_\#\mu(\mX^2)\right)
&\le
4 \expect W_2^2(f_\#\mu(\mX^1), \mu).
\end{align*}
\end{proof}
Proposition~\ref{thm:ub} is valid for an arbitrary measure $\mu$. Let
\begin{equation}
\mu
=\bar{\mu}
\defeq\argmin_{\mu}
W_2^2\left(f_\#\mu(\mX^1), \mu\right)
\end{equation}
be the Wasserstein barycentre of the random measure
$f_\#\mu(\mX^1)$. Then
$\expect W_2^2(f_\#\mu(\mX^1), \bar{\mu})$ is the Wasserstein
variance which bounds 
the loss from above.

Given $\mX$, the random distance $W_2\left(f_\#\mu(\mX^1), \bar{\mu}\right)$ is bounded. We have
\begin{equation*}
0\le W_2\left(f_\#\mu(\mX^1), \bar{\mu}\right) \le R,
\end{equation*}
where $R\defeq\max{}W_2(f_{\#}\mu(\bm{X}^1), \bar{\mu})$ is a constant depending on $f_\theta$ and the dataset $\mX$. Therefore
\begin{equation*}
\mathrm{var}
\left(
W_2\left(f_\#\mu(\mX^1), \bar{\mu}\right)
\right)
\le
\frac{1}{4}R^2,
\end{equation*}
where $\mathrm{var}(\cdot)$ denotes the variance. 
By Proposition~\ref{thm:ub} and the Popoviciu's inequality, we have
\begin{align*}
\expect
W_2^2\left(f_\#\mu(\mX^1), f_\#\mu(\mX^2)\right)
&\le
4 \left(\expect W_2(f_\#\mu(\mX^1), \bar{\mu})\right)^2
+ 2 \mathrm{var}\left(
W_2\left(f_\#\mu(\mX^1), \bar{\mu}\right)
\right)\nonumber\\
&\le
4 \left(\expect W_2(f_\#\mu(\mX^1), \bar{\mu})\right)^2
+
\frac{1}{2}R^2.
\end{align*}
Hence, the (expected) loss is bounded above by the 2-Wasserstein distance between the random measure $f_{\#}\mu(\bm{X}^1)$
and its center. It is therefore \emph{a variance-like measure}.

\begin{proposition}\label{thm:batchsize}
Let $\mX^1$, $\mX^2$ be independent random batches of size $B$,
$\mX^3$, $\mX^4$ be independent random batches of size $2B$,
then
\begin{equation}
    \expect W^2_2(f_{\#}\mu(\bm{X}^3), f_{\#}\mu(\bm{X}^4))
    \le
    \expect W^2_2(f_{\#}\mu(\bm{X}^1), f_{\#}\mu(\bm{X}^2)).
\end{equation}
If $B=1$, then
\begin{equation}
    \expect W^2_2(f_{\#}\mu(\bm{X}^1), f_{\#}\mu(\bm{X}^2))
    =
    \expect \lVert f_\theta(\xi) - f_\theta(\xj) \rVert^2.
\end{equation}
\end{proposition}
Therefore, as the batch size $B$ increases, the loss will decrease. Eventually, when $B$ is sufficiently large, 
the distance between the two measures 
$f_{\#}\mu(\bm{X}^1)$ and $f_{\#}\mu(\bm{X}^2)$ will be
close to zero as they both become 
close to $f_{\#}\mu(\bm{X})$.

To prove the above proposition, we introduce the lemma below first.
\begin{lemma}
\label{lemma-to-proof}
Let $\mX^1$ and $\mX^2$ (resp. $\mX^3$ and $\mX^4$) be multisets 
of the same size $B$ (resp. $B'$).
\begin{equation}
    (B+B')W^2_2(\mu(\mX^1\cup\mX^3), \mu(\mX^2\cup\mX^4))
    \le
    B W^2_2(\mu(\mX^1), \mu(\mX^2)) +
    B' W^2_2(\mu(\mX^3), \mu(\mX^4)).
\end{equation}
\end{lemma}

\begin{proof}
By Lemma~\ref{lemma},
\begin{align*}
   B W_2^2\left(\mu(\mX^1), \mu(\mX^2)\right) 
   &= \min_{\pi} \sum_{i=1}^B
   \lVert \mX^1[i,:] - \mX^2[\pi(i),:]\rVert^2, \nonumber\\
   B' W_2^2\left(\mu(\mX^3), \mu(\mX^4)\right) 
   &= \min_{\pi'} \sum_{j=1}^{B'}
   \lVert \mX^3[j,:] - \mX^4[\pi'(j),:]\rVert^2,
\end{align*}
where the minimum is taken over all possible permutations
of $(1,\cdots,B)$ (resp. $(1,\cdots,B')$ ).
Denote the optimal permutation as $\pi^\star$
(resp. $\pi'^{\star}$).
Then we can construct a permutation $\sigma^\star$ of the index set
$(1,\cdots,B,B+1,\cdots,B+B')$ by permuting $(1,\cdots,B)$ wrt 
$\pi^\star$ and permuting $(B+1,\cdots,B+B')$ wrt $\pi'^\star$.
Thus we have
\begin{align*}
   (B+B') W_2^2\left(\mu(\mX^1\cup\mX^3), \mu(\mX^2\cup\mX^4)\right) 
   &= \min_{\sigma} \sum_{i=1}^{B+B'}
   \lVert (\mX^1\cup\mX^3)[i,:] - (\mX^2\cup\mX^4)[\sigma(i),:]\rVert^2\nonumber\\
   &\le
   \sum_{i=1}^{B+B'}
   \lVert (\mX^1\cup\mX^3)[i,:] - (\mX^2\cup\mX^4)[\sigma^\star(i),:]\rVert^2\nonumber\\
   &=
   \sum_{i=1}^B
   \lVert \mX^1[i,:] - \mX^2[\pi^\star(i),:]\rVert^2
   +
   \sum_{i=1}^{B'}
   \lVert \mX^3[i,:] - \mX^4[\pi'^{\star}(i),:]\rVert^2
   \nonumber\\
   &=
   B W_2^2\left(\mu(\mX^1), \mu(\mX^2)\right) 
   +
   B' W_2^2\left(\mu(\mX^3), \mu(\mX^4)\right).
\end{align*}
\end{proof}

The proof of Proposition~\ref{thm:batchsize} is as follows.
\begin{proof}
In Lemma~\ref{lemma-to-proof}, let $B'=B$, and take expectation on both sides of the inequality, then Proposition~\ref{thm:batchsize} is immediate.
If $B=1$, the 2-Wasserstein distance between empirical measures becomes the Euclidean distance.
\end{proof}

\subsection{Invertibility Analysis}
\label{sec-invertibility}
We rewrite Eq.~\ref{eq:dinh1} and Eq.~\ref{eq:dinh2} as
\begin{align*}
    &\vz_{1:d} = \vy_{1:d}^{\text{in}} \odot \exp\left(g_1(\vy_{d+1:D}^{\text{in}})\right) + h_1(\vy_{d+1:D}^{\text{in}}),\\
    &\vz_{d+1:D} = \vy^{\text{in}}_{d+1:D},\\
    &\vy_{1:d}^{\text{out}} = \vz_{1:d},\\
    &\vy_{d+1:D}^{\text{out}} = \vz_{d+1:D} \odot \exp\left(g_2(\vz_{1:d})\right) + h_2(\vz_{1:d}),
\end{align*}
where $\vz$ is a $D$-dimensional intermediate vector.
Both the mappings
$\vy^{\text{in}}\to\vz$ and $\vz\to\vy^{\text{out}}$
are invertible (the inverse has a simple closed form~\cite{dinh2016density} and is omitted here), and therefore 
$\vy^{\text{in}}\to\vy^{\text{out}}$ is invertible.

Note that $\vy_i^{\text{out}}$
(the $i$'th dimension of $\vy^{\text{out}}$)
only depends on $\vy^{\text{in}}_{1:i}$. The Jacobian of
the mapping $\vy^{\text{in}}\to\vy^{\text{out}}$ has a
\emph{lower triangular} structure.

\subsection{Proof of Proposition~\ref{thm:infomax}}
\label{sec-infomax-proof}
\begin{proof}
By definition, we have $I(\mX, f'(\mX)) = H(\mX) - H(\mX | f'(\mX))$ where $H(\mX)$ and $H(\mX | f'(\mX))$ are the entropy and conditional entropy, respectively. As we consider $\mX$ and $f'(\mX)$ as empirical random variables with finite supports of their samples, $H(\mX | f'(\mX)) \ge 0$. Therefore, $I(\mX, f'(\mX)) \le  H(\mX) = I(\mX, \mX)$. If  $f_\theta$ is a smooth invertible map, it is known that: 
$I(\mX, f_\theta(\mX)) = I(\mX, \mX)$ (proof shown in Eq.~(45) of~\citet{kraskov2004estimating}). Therefore, $I(\mX, f_\theta(\mX)) \ge I(\mX, f'(\mX))$.
\end{proof}

\section{Hyperparameter Sensitivity}
\label{sec-ps}
To test the sensitivity of TDM to the two main hyperparameters $T$ and $K$, we vary them from 1 to 4 and show the imputation metrics in Figure~\ref{fig-ab-mae},~\ref{fig-ab-rmse},~\ref{fig-ab-ot}. It can be observed that increasing $T$ improves the performance in general but comparing $T=4$ with $T=3$, the improvement becomes marginal and 
overfitting can also be observed in a few datasets, e.g., QB and AC. In addition, one can see that varying $K$ does not have significant impact on the performance.

We also report the sensitivity of our method to different batch sizes (128, 256, 512) in the comparison with OTImputer, shown in Figure~\ref{fig-bs-mae},~\ref{fig-bs-rmse},~\ref{fig-bs-ot}. It can be seen that larger batch sizes in general give better performance of both TDM and OTImputer.

\begin{figure*}[t]
\captionsetup[subfigure]{justification=centering}
        \centering
         \begin{subfigure}[b]{0.16\linewidth}
                 \centering
                 \caption{Ground truth}
                 \includegraphics[width=0.99\textwidth]{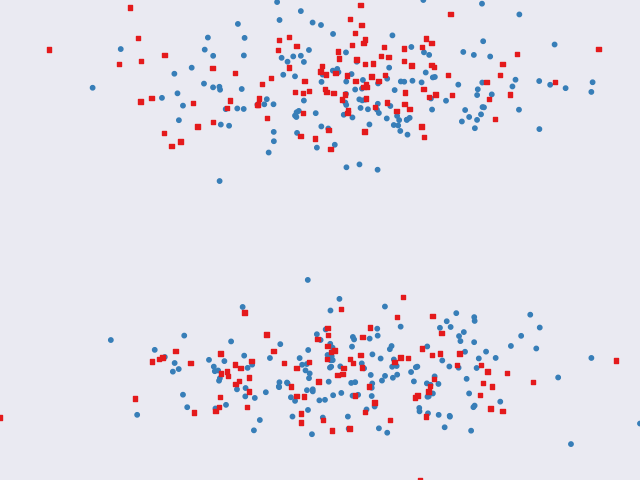}
         \end{subfigure}
         \begin{subfigure}[b]{0.16\linewidth}
                 \centering
                 \caption{OTImputer}
                 \includegraphics[width=0.99\textwidth]{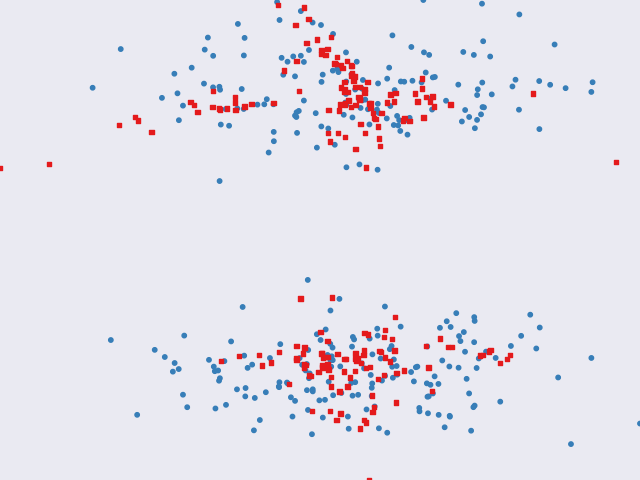}
         \end{subfigure} 
         \begin{subfigure}[b]{0.16\linewidth}
                 \centering
                 \caption{TDM (ours)}
                 \includegraphics[width=0.99\textwidth]{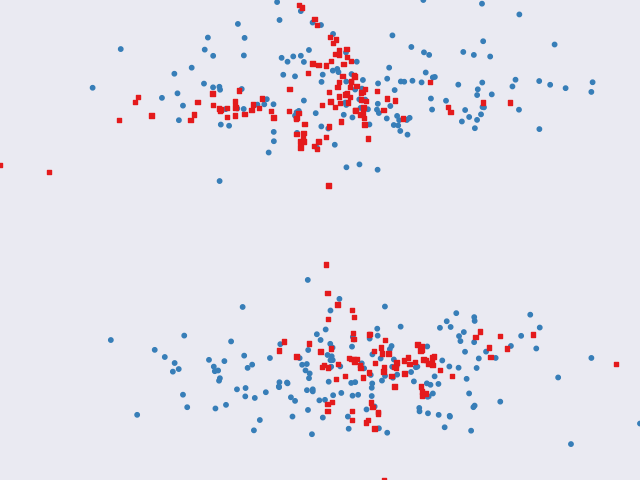}
         \end{subfigure}
         \begin{subfigure}[b]{0.16\linewidth}
                 \centering
                 \caption{$f_{1}(\mX)$}
                 \includegraphics[width=0.99\textwidth]{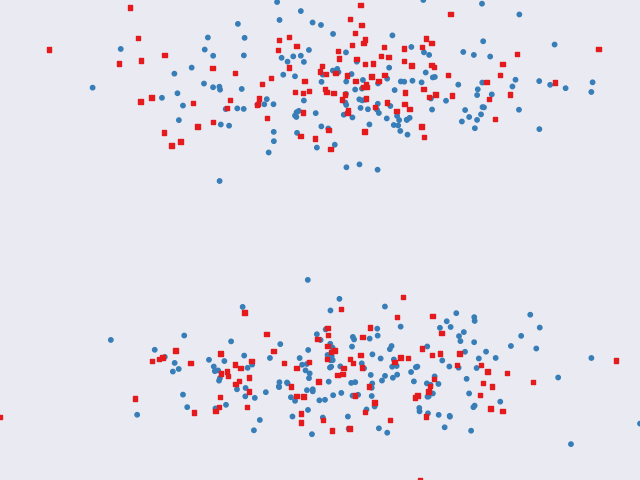}
         \end{subfigure}
          \begin{subfigure}[b]{0.16\linewidth}
                 \centering
                 \caption{$f_{1:2}(\mX)$}
                 \includegraphics[width=0.99\textwidth]{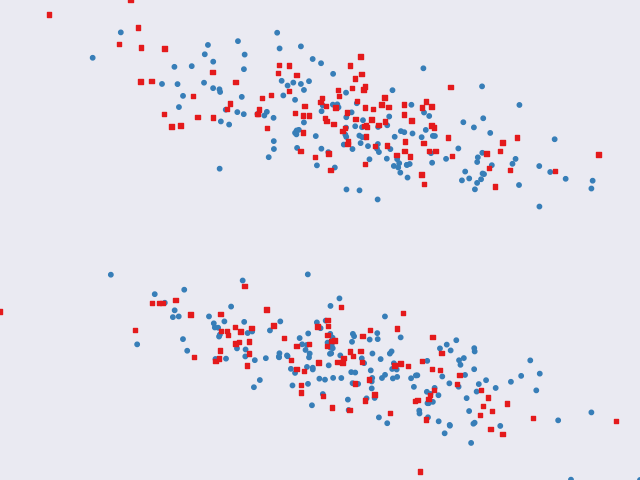}
         \end{subfigure} 
        \begin{subfigure}[b]{0.16\linewidth}
                 \centering
                 \caption{$f_{1:3}(\mX)$}
                 \includegraphics[width=0.99\textwidth]{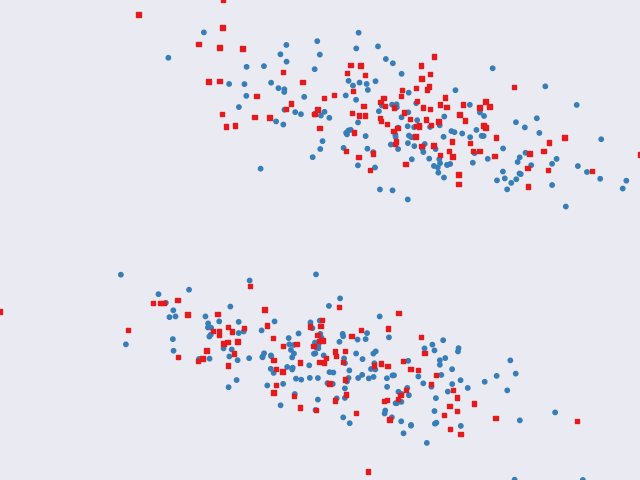}
         \end{subfigure} \\
        \begin{subfigure}[b]{0.16\linewidth}
                 \centering
                 \includegraphics[width=0.99\textwidth]{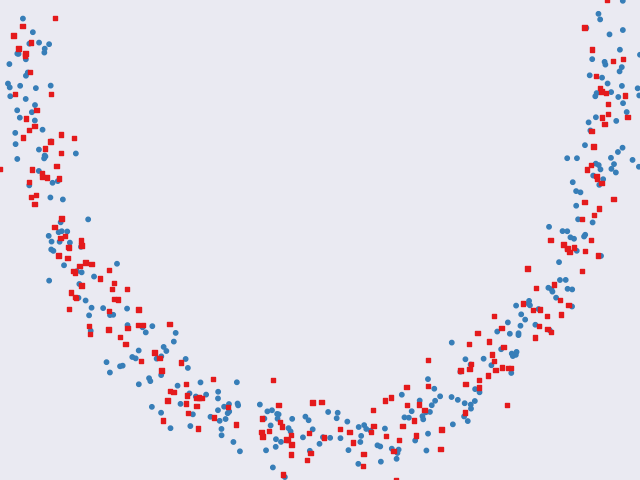}
         \end{subfigure}
         \begin{subfigure}[b]{0.16\linewidth}
                 \centering                 \includegraphics[width=0.99\textwidth]{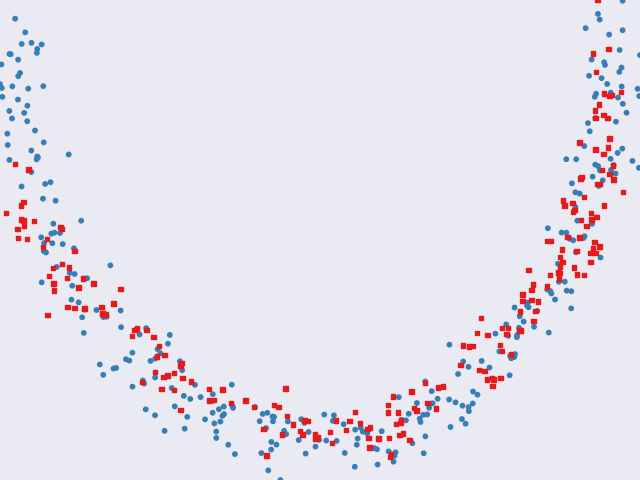}
         \end{subfigure} 
         \begin{subfigure}[b]{0.16\linewidth}
                 \centering
                 \includegraphics[width=0.99\textwidth]{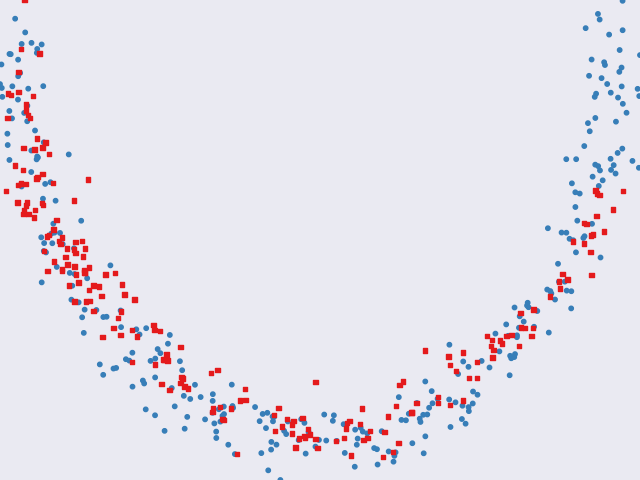}
         \end{subfigure}
         \begin{subfigure}[b]{0.16\linewidth}
                 \centering
                 \includegraphics[width=0.99\textwidth]{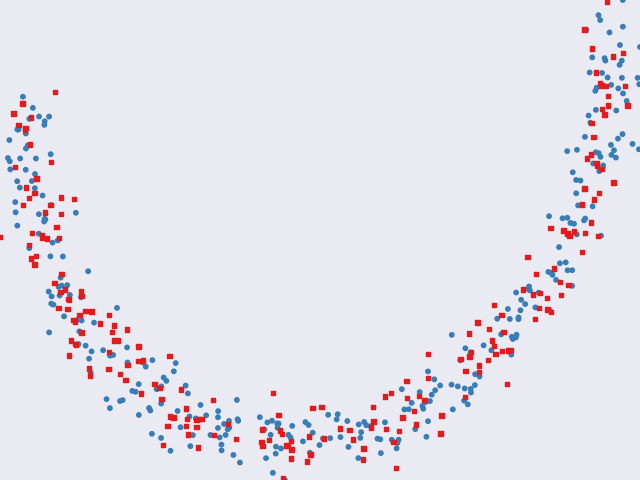}
         \end{subfigure}
          \begin{subfigure}[b]{0.16\linewidth}
                 \centering
                 \includegraphics[width=0.99\textwidth]{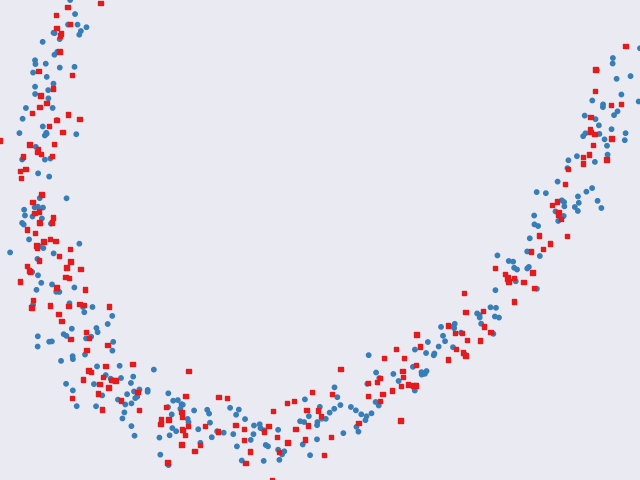}
         \end{subfigure} 
        \begin{subfigure}[b]{0.16\linewidth}
                 \centering
                 \includegraphics[width=0.99\textwidth]{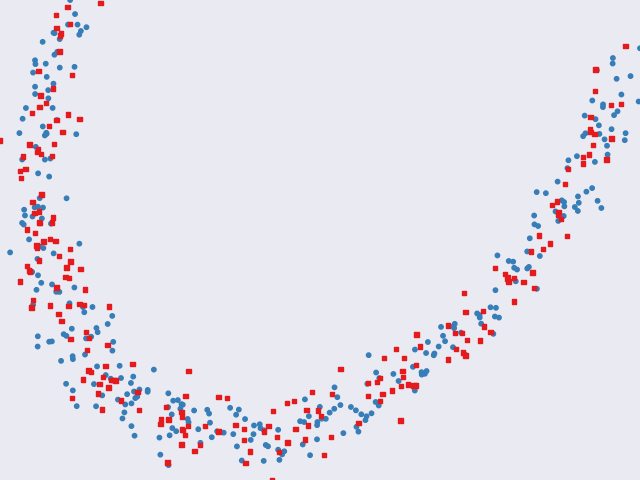}
         \end{subfigure} 
\caption{Two synthetic datasets (two rows) each of which is with 500 samples. (a) Ground truth: Blue points (60\%) have no missing values and red points (40\%) have one missing value on either coordinate (following MCAR). (b) The imputed values for the red points by OTImputer. (c) The imputed values for the red points by TDM. (d-f) The transformed points by different blocks of $f_\theta$.}
\label{fig-motivation-44}
 \vspace{-0.5cm}
\end{figure*}

\begin{table}[]
\rowa{1.3}
\centering
\caption{Dataset statistics}
\label{tb-ds}
\resizebox{0.45\linewidth}{!}{
\begin{tabular}{@{}cccc@{}}
\toprule
Dataset              & $N$      & $D$  & Abbreviation \\ 
\midrule
california           & 20,640 & 8  & CA           \\ 
qsar\_biodegradation & 1,055  & 41 & QB           \\ 
blood\_transfusion   & 748    & 4  & BT           \\ 
wine\_quality        & 4,898  & 11 & WQ           \\ 
parkinsons           & 195    & 23 & PK           \\ 
yacht\_hydrodynamics & 308    & 6  & YH           \\ 
seeds                & 210    & 7  & SD           \\ 
glass                & 214    & 9  & GL           \\ 
planning\_relax      & 182    & 12 & PR           \\ 
concrete\_slump      & 103    & 7  & CS           \\ 
anuran\_calls        & 7,195  & 22 & AC           \\ 
letter               & 20,000 & 16 & LT           \\ \bottomrule
\end{tabular}
}
\end{table}

\clearpage

\begin{figure*}[t]
\captionsetup[subfigure]{justification=centering}
        \centering
         \begin{subfigure}[b]{0.88\linewidth}
                 \centering                 \includegraphics[width=0.99\textwidth]{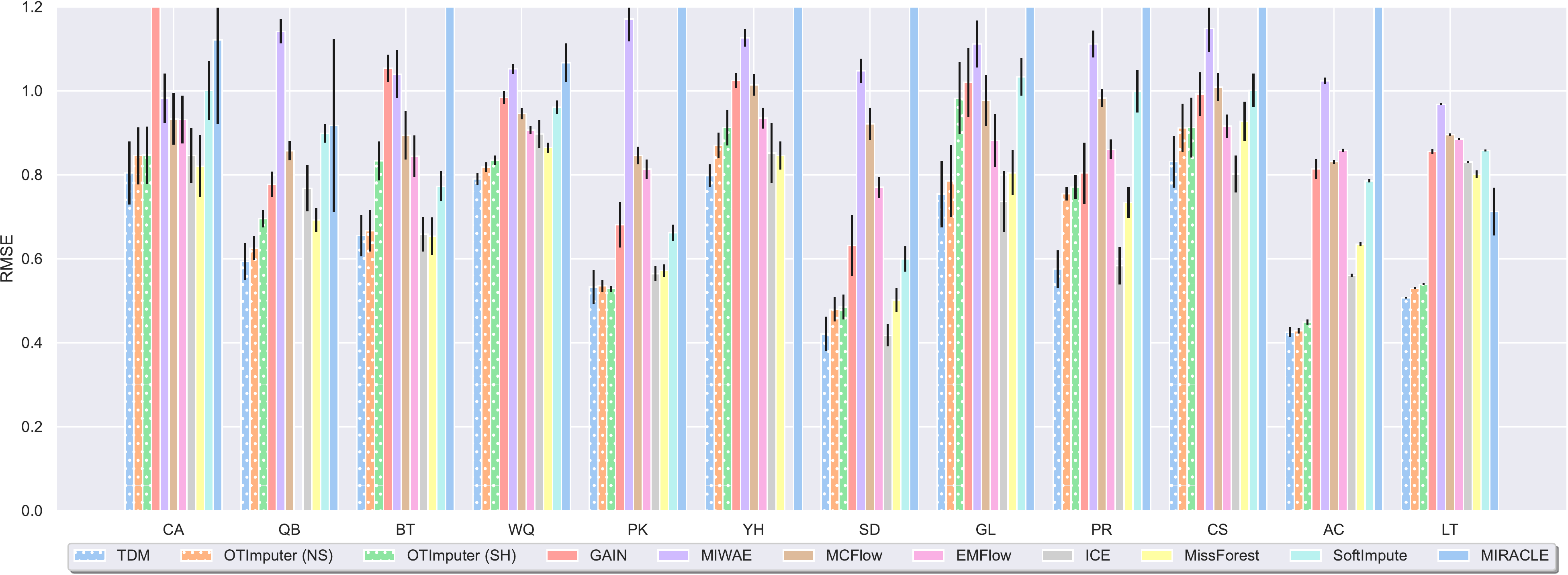}
         \end{subfigure}
         \\
         \begin{subfigure}[b]{0.88\linewidth}
                 \centering                 \includegraphics[width=0.99\textwidth]{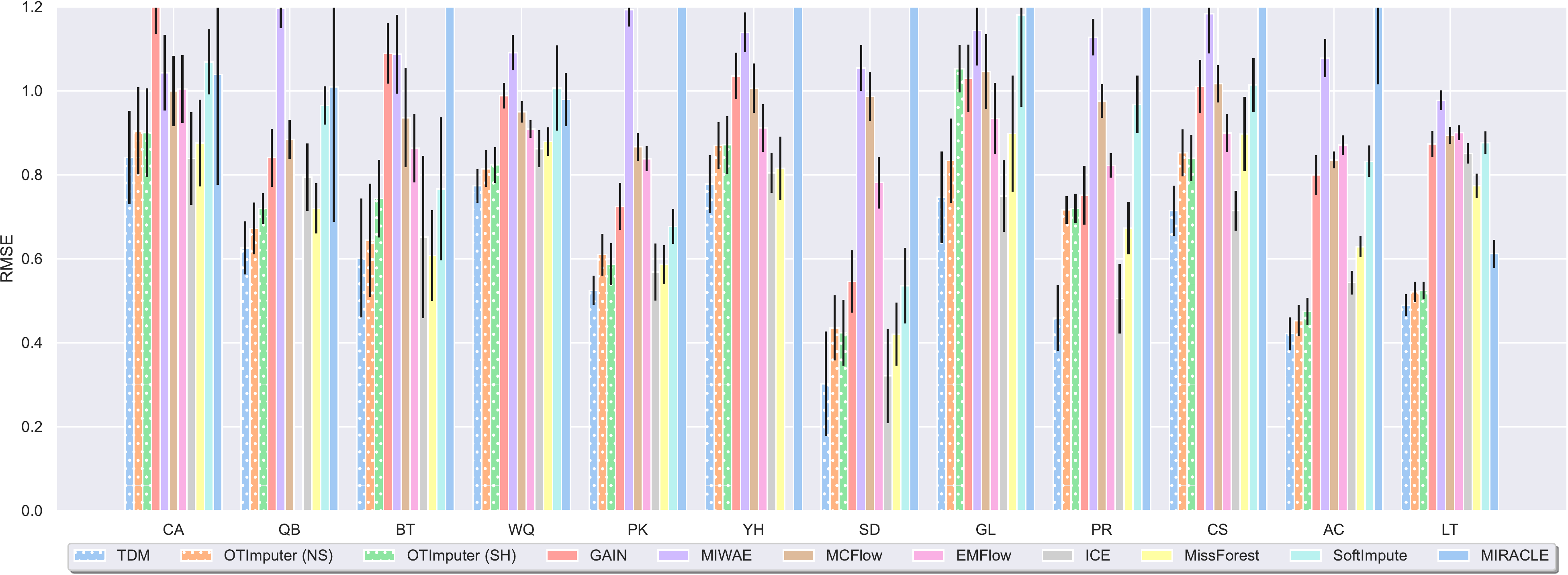}
         \end{subfigure}\\
         \begin{subfigure}[b]{0.88\linewidth}
                 \centering                 \includegraphics[width=0.99\textwidth]{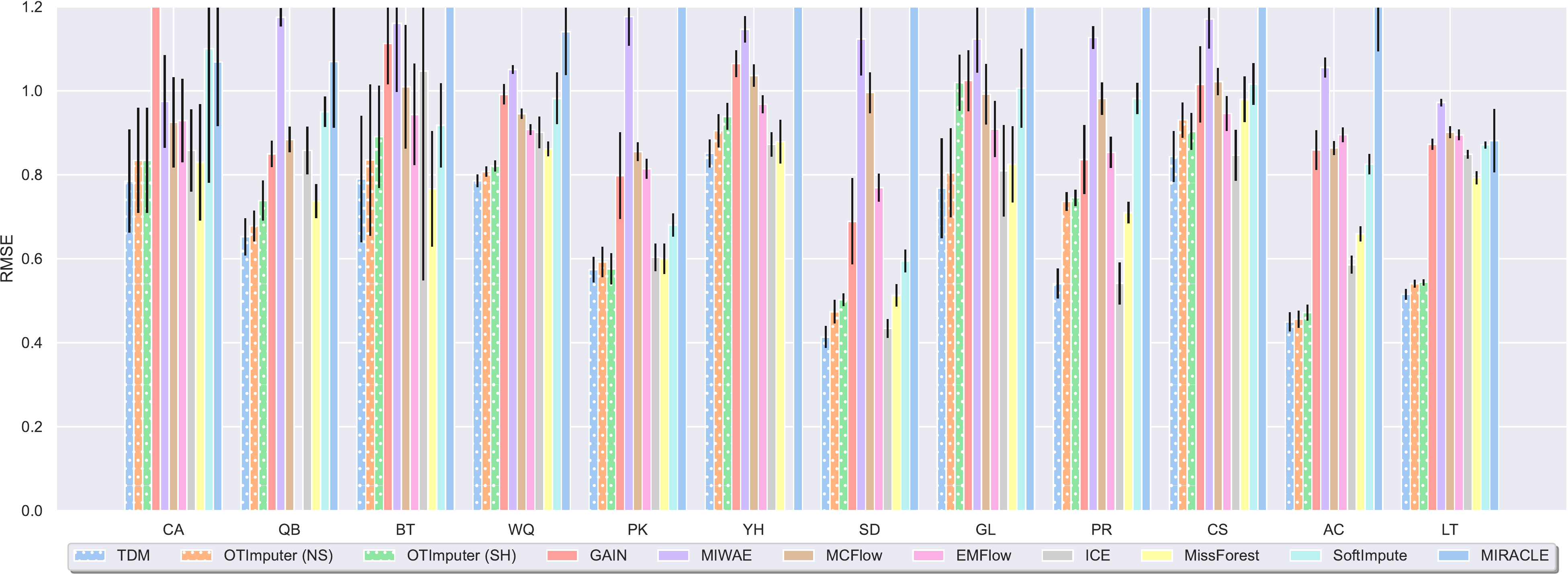}
         \end{subfigure}
         \\
         \begin{subfigure}[b]{0.88\linewidth}
                 \centering                 \includegraphics[width=0.99\textwidth]{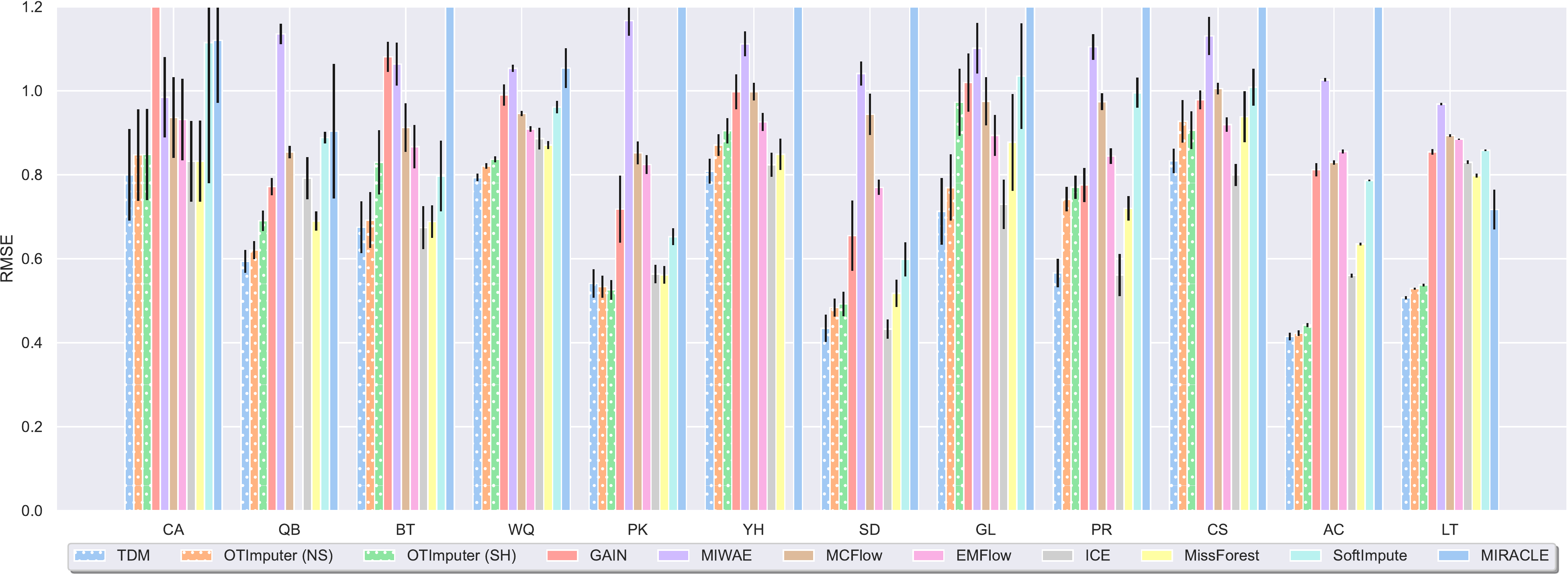}
         \end{subfigure}
 \caption{From top to bottom: RMSE in the MCAR, MAR, MNARL, and MNARQ settings.}
  \label{fig-rmse}
 \vspace{-0.5cm}
\end{figure*}

\begin{figure*}[t]
\captionsetup[subfigure]{justification=centering}
        \centering
         \begin{subfigure}[b]{0.88\linewidth}
                 \centering                 \includegraphics[width=0.99\textwidth]{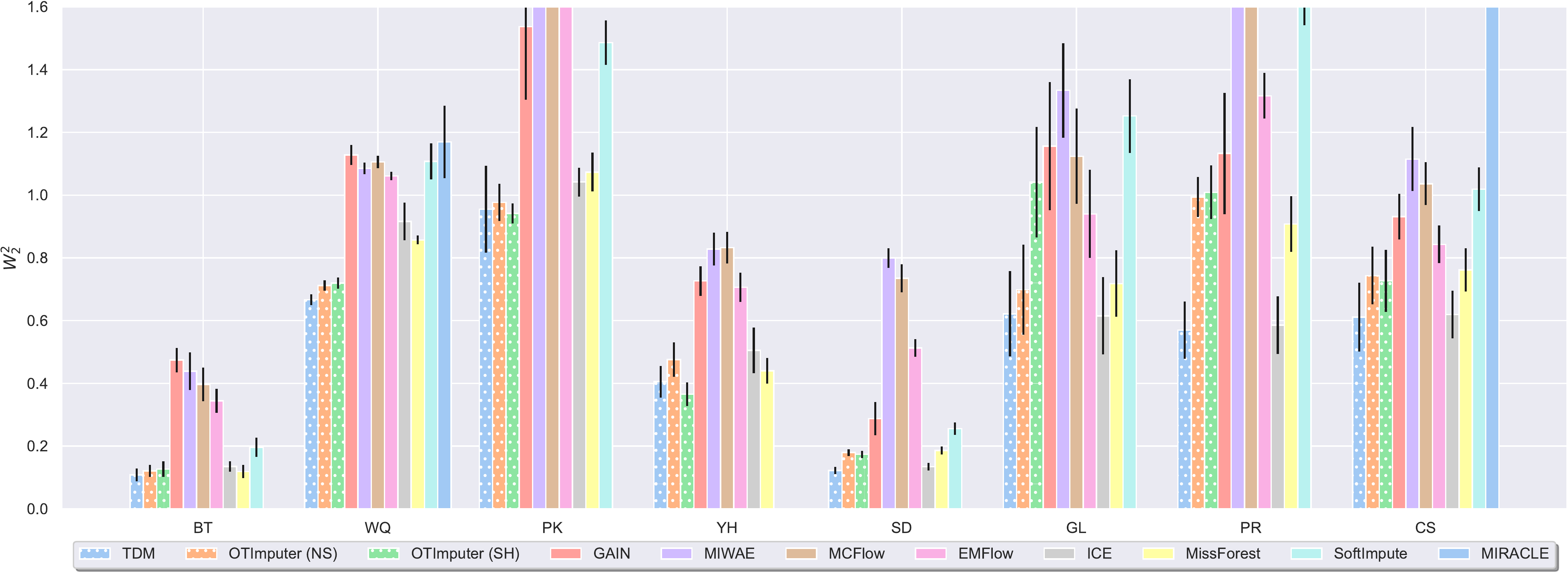}
         \end{subfigure}
         \\
         \begin{subfigure}[b]{0.88\linewidth}
                 \centering                 \includegraphics[width=0.99\textwidth]{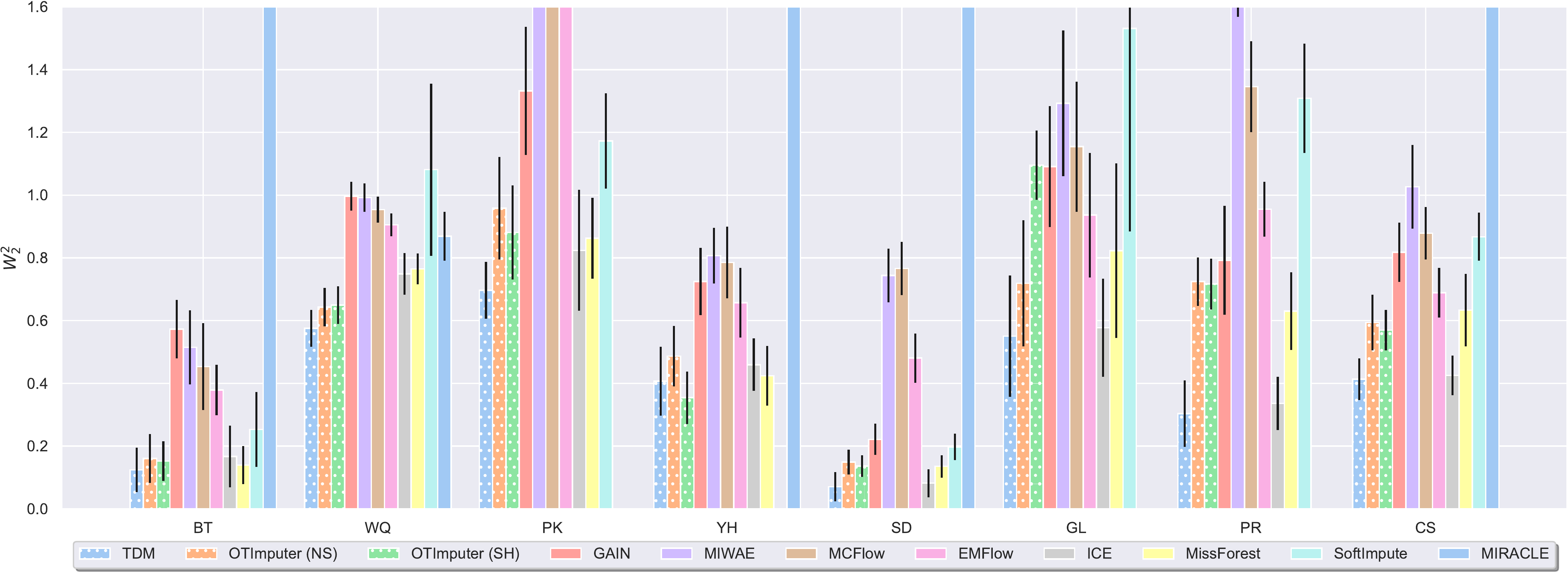}
         \end{subfigure}
         \\
         \begin{subfigure}[b]{0.88\linewidth}
                 \centering                 \includegraphics[width=0.99\textwidth]{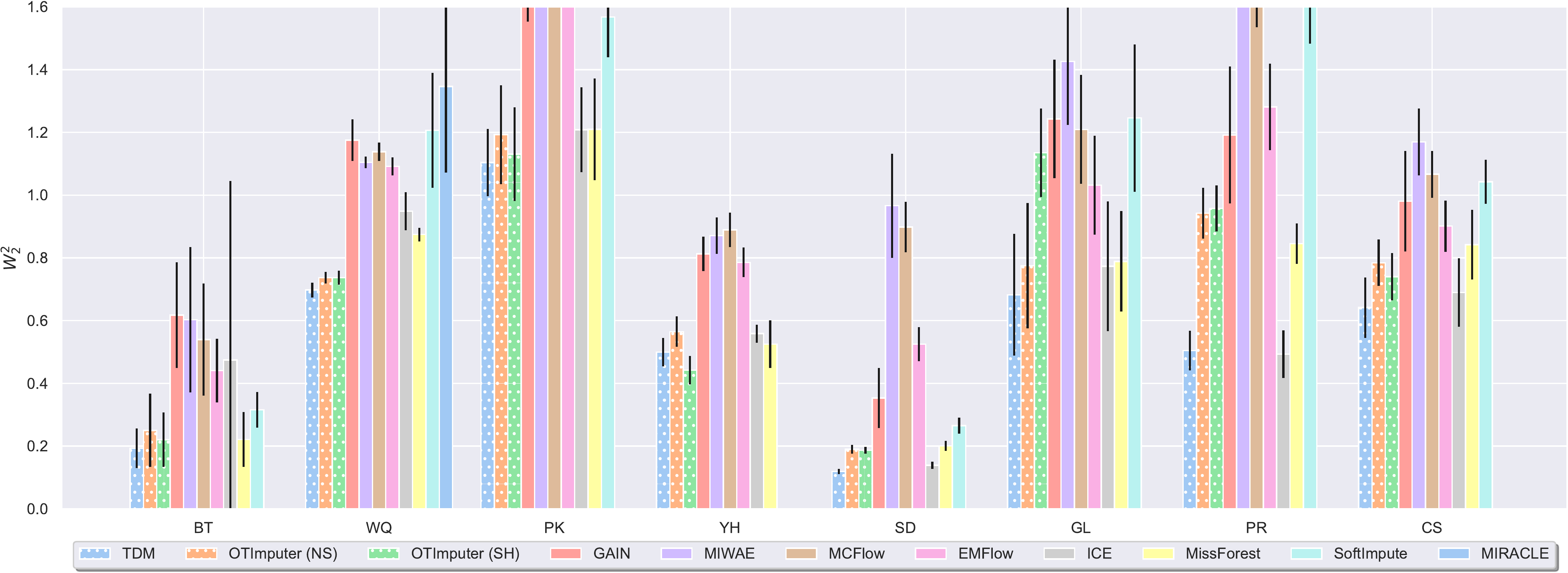}
         \end{subfigure}
         \\
         \begin{subfigure}[b]{0.88\linewidth}
                 \centering                 \includegraphics[width=0.99\textwidth]{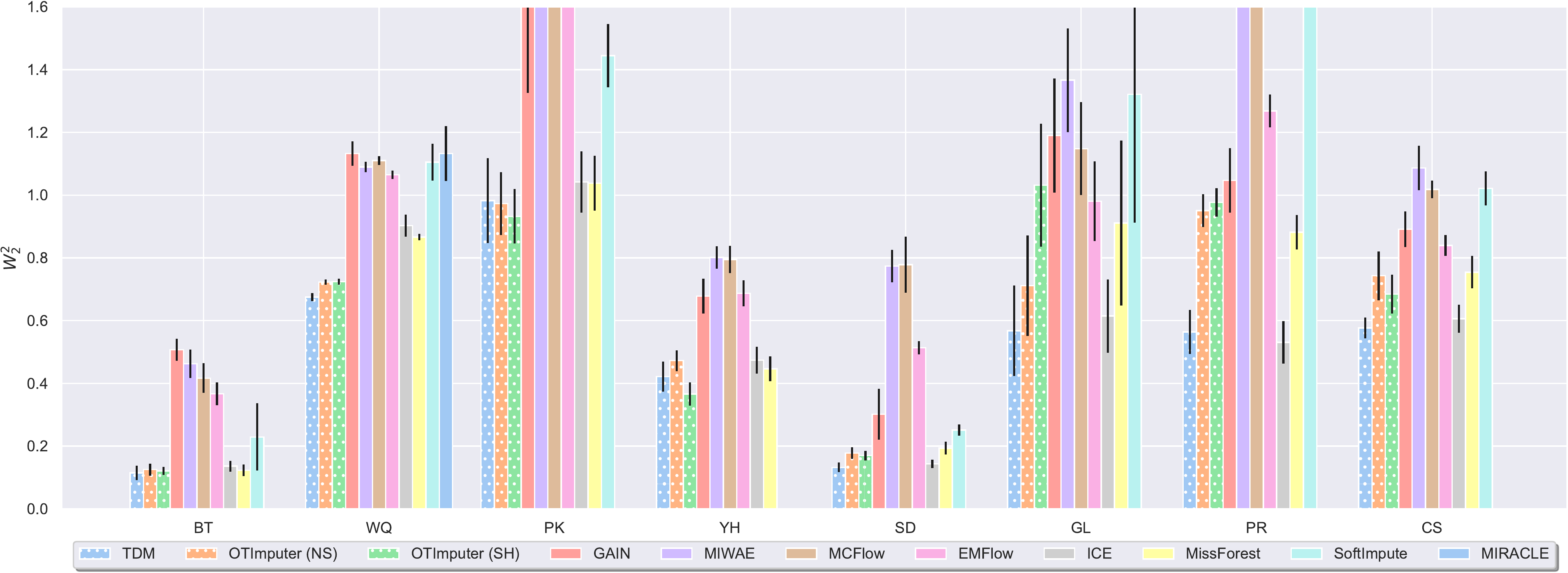}
         \end{subfigure}
 \caption{From top to bottom: $W^2_2$ in the MCAR, MAR, MNARL, and MNARQ settings.}
  \label{fig-ot}
 \vspace{-0.5cm}
\end{figure*}

\begin{figure*}[t]
\captionsetup[subfigure]{justification=centering}
        \centering
         \begin{subfigure}[b]{0.99\linewidth}
                 \centering                 \includegraphics[width=0.99\textwidth]{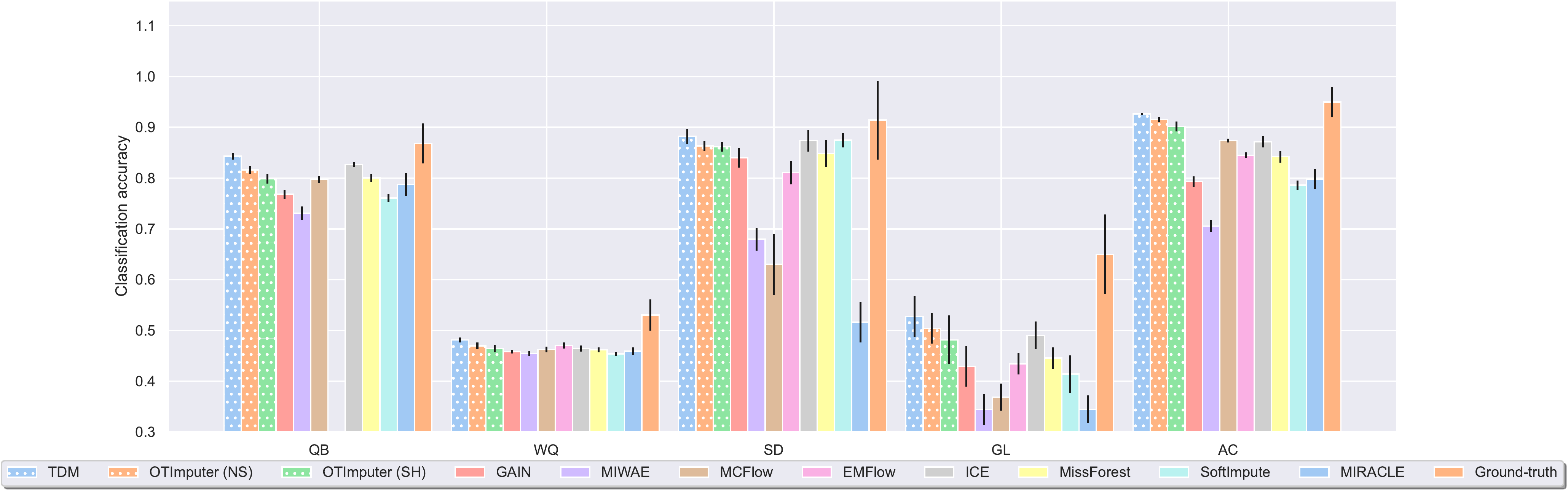}
         \end{subfigure}\\
          \begin{subfigure}[b]{0.99\linewidth}
                 \centering                 \includegraphics[width=0.99\textwidth]{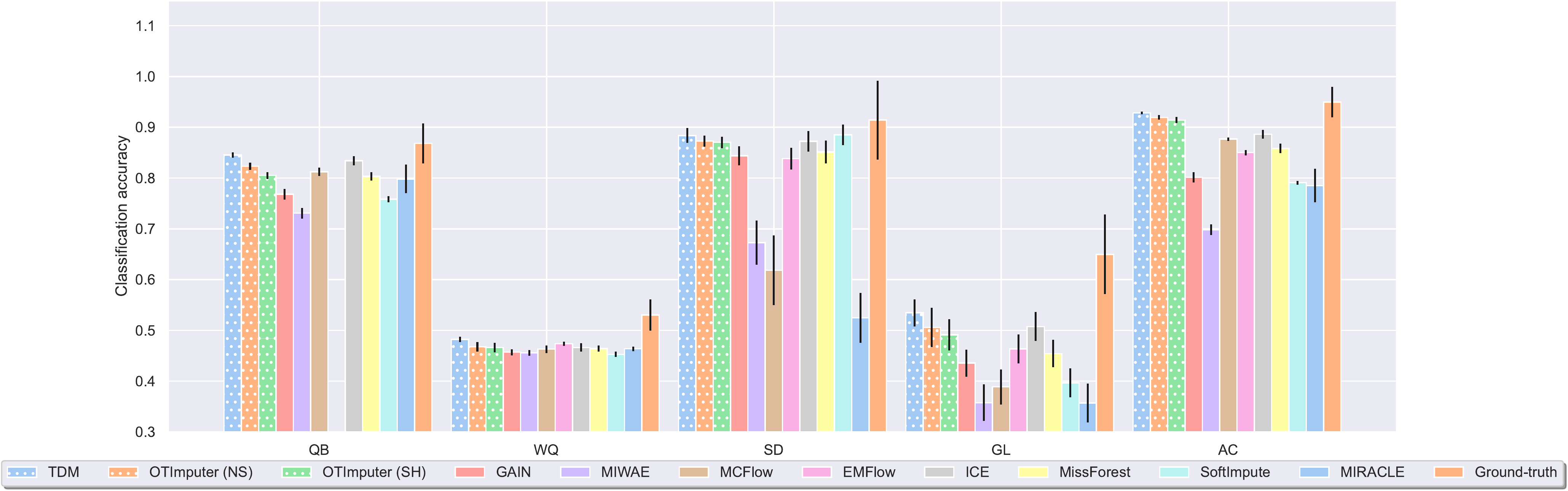}
         \end{subfigure}\\
          \begin{subfigure}[b]{0.99\linewidth}
                 \centering                 \includegraphics[width=0.99\textwidth]{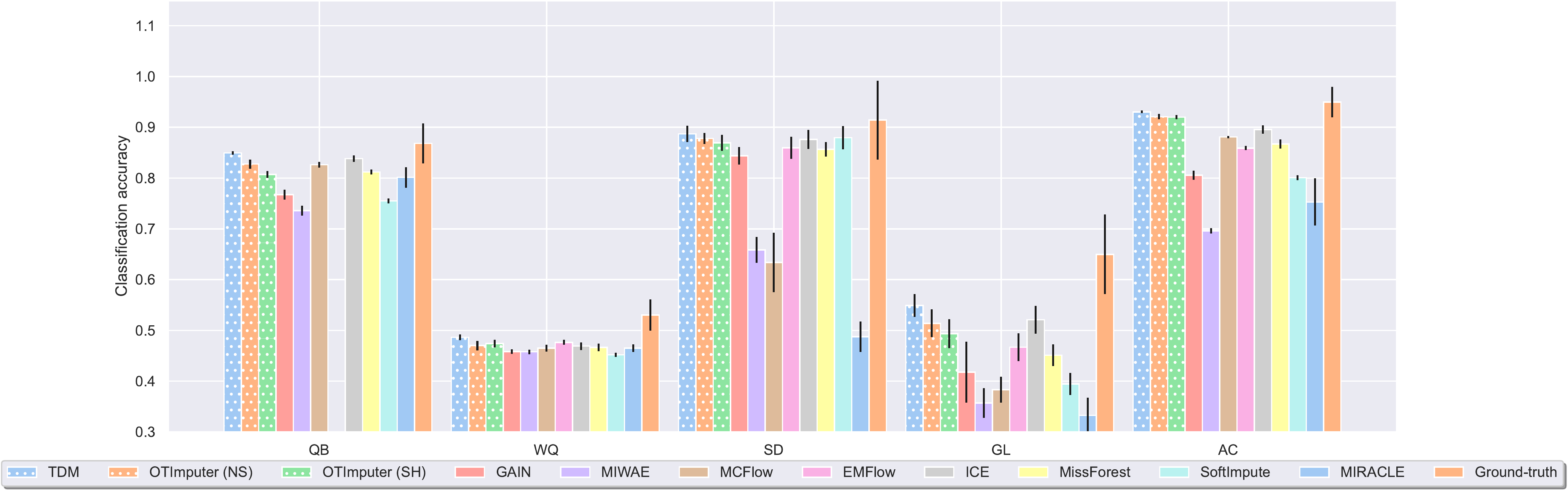}
         \end{subfigure}
 \caption{From top to bottom: Classification accuracy in the MAR, MNARL, and MNARQ settings.}
  \label{fig-cla}
 \vspace{-0.5cm}
\end{figure*}

\begin{figure*}[t]
\captionsetup[subfigure]{justification=centering}
        \centering
         \begin{subfigure}[b]{0.88\linewidth}
                 \centering                 \includegraphics[width=0.99\textwidth]{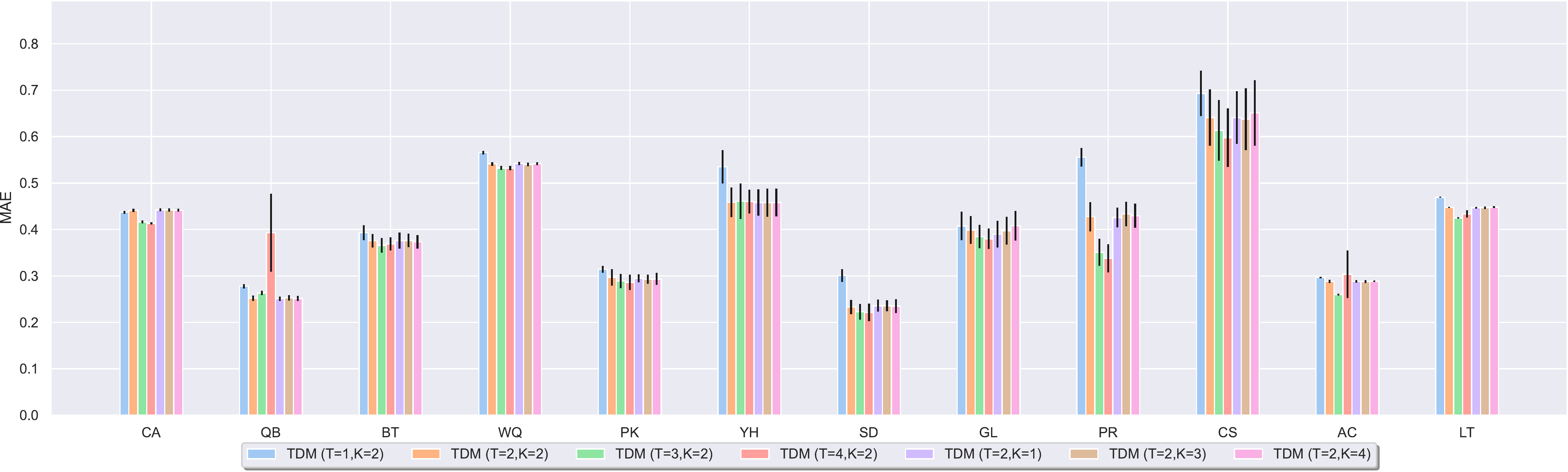}
         \end{subfigure}\\
         \begin{subfigure}[b]{0.88\linewidth}
                 \centering                 \includegraphics[width=0.99\textwidth]{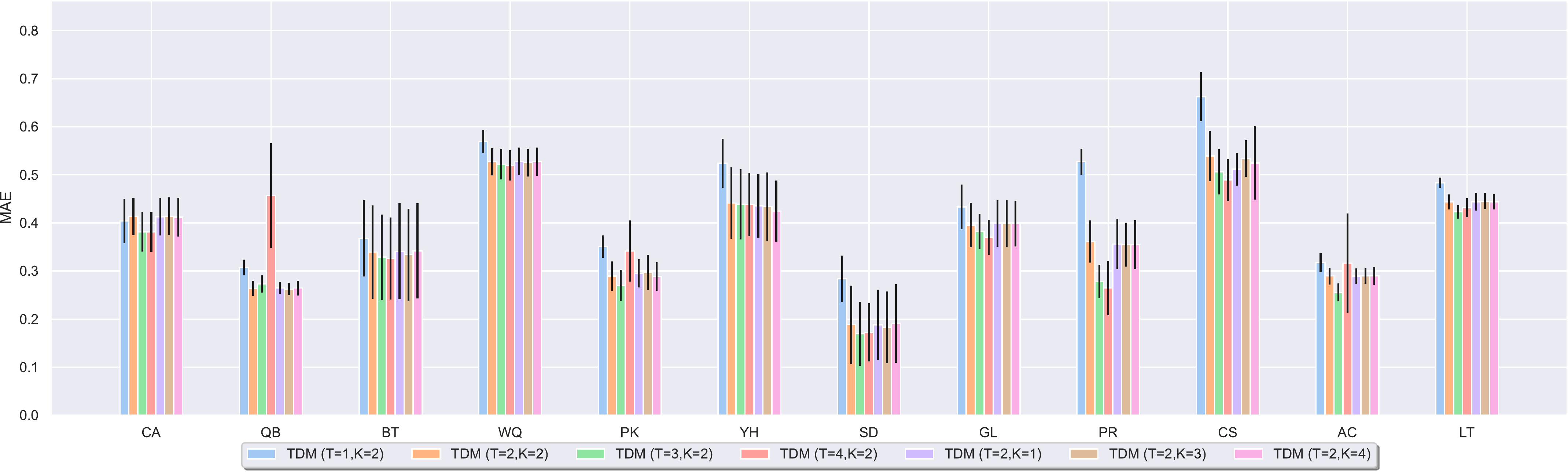}
         \end{subfigure}\\
         \begin{subfigure}[b]{0.88\linewidth}
                 \centering                 \includegraphics[width=0.99\textwidth]{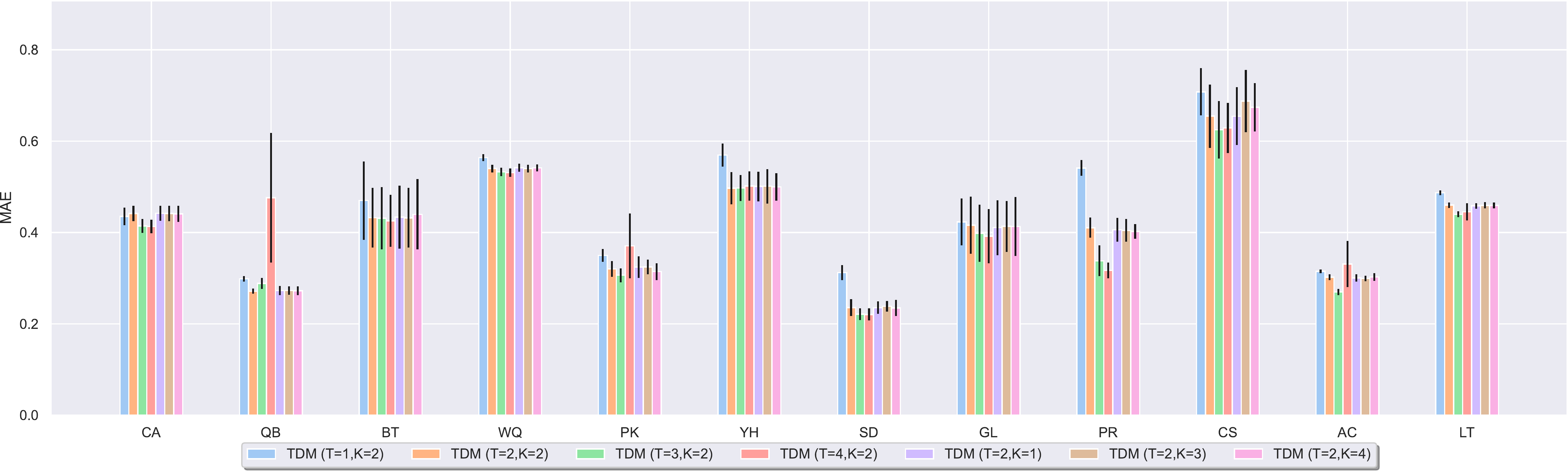}
         \end{subfigure}
         \\
         \begin{subfigure}[b]{0.88\linewidth}
                 \centering                 \includegraphics[width=0.99\textwidth]{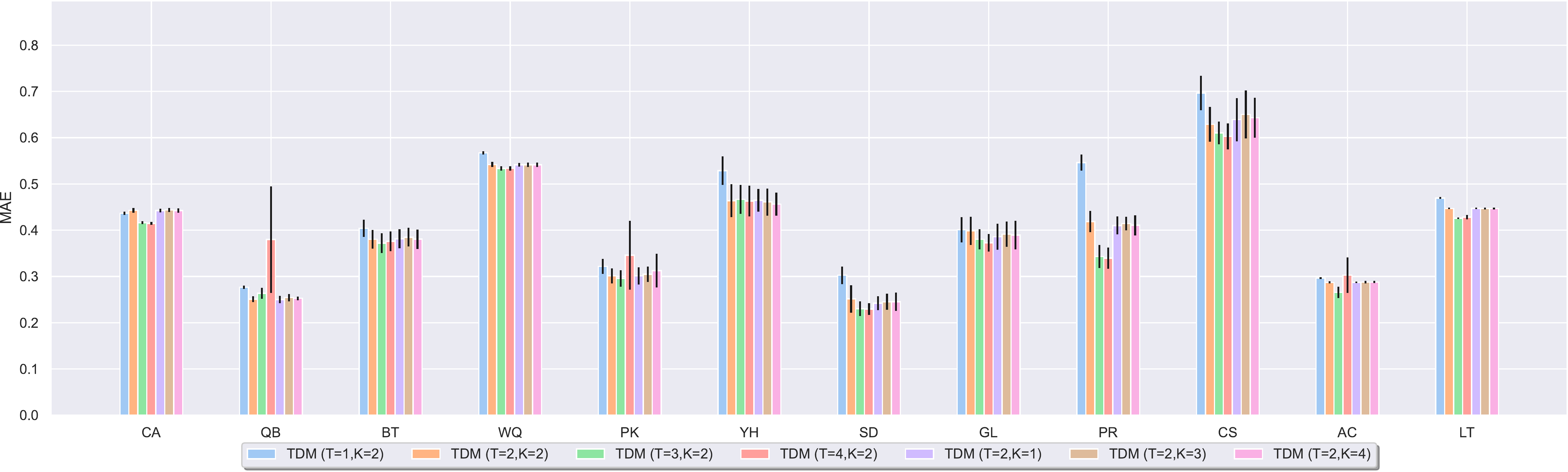}
         \end{subfigure}
\centering
 \caption{From top to bottom: MAE in the MCAR, MAR, MNARL, and MNARQ settings for TDM with different $T$ and $K$.}
  \label{fig-ab-mae}
 \vspace{-0.5cm}
\end{figure*}

\begin{figure*}[t]
\captionsetup[subfigure]{justification=centering}
        \centering
         \begin{subfigure}[b]{0.88\linewidth}
                 \centering                 \includegraphics[width=0.99\textwidth]{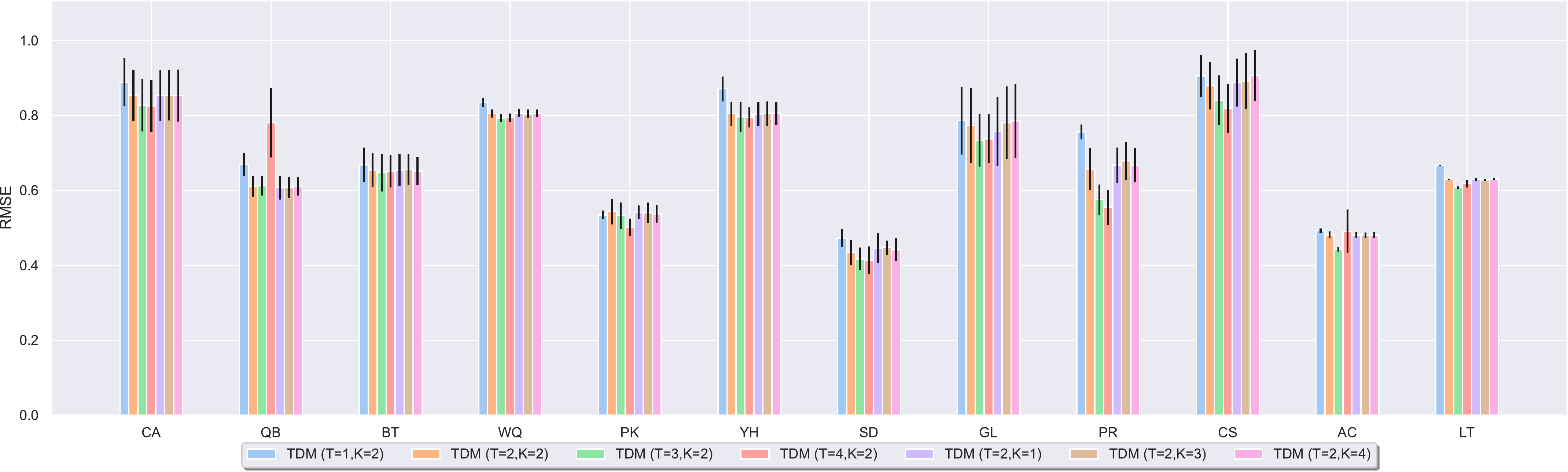}
         \end{subfigure}
         \\
         \begin{subfigure}[b]{0.88\linewidth}
                 \centering                 \includegraphics[width=0.99\textwidth]{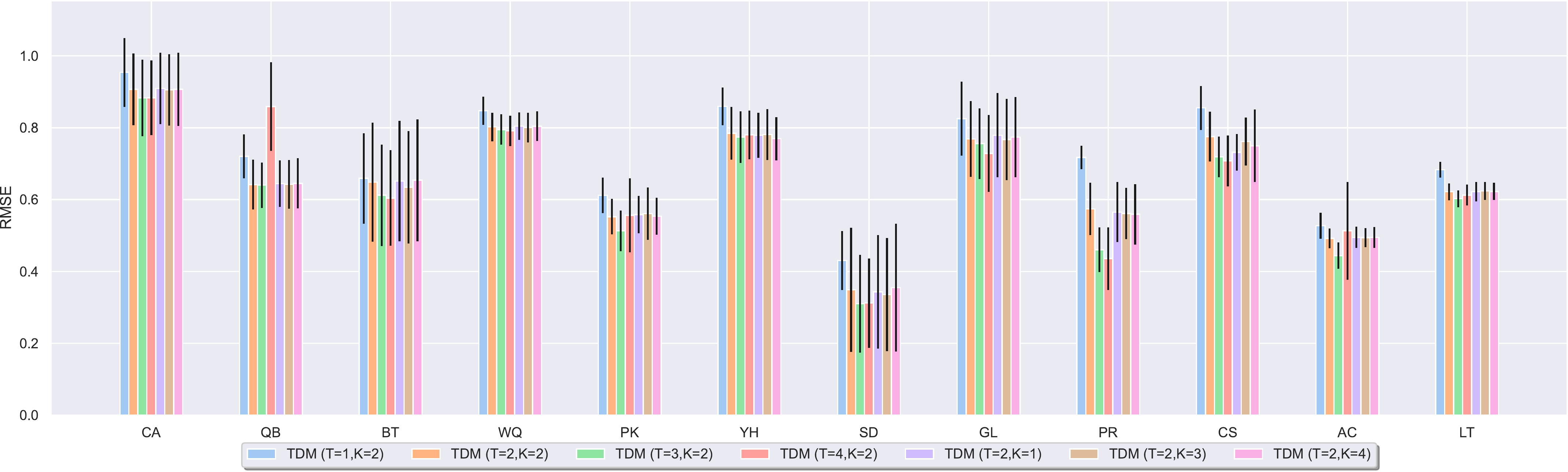}
         \end{subfigure}
        \\
         \begin{subfigure}[b]{0.88\linewidth}
                 \centering                 \includegraphics[width=0.99\textwidth]{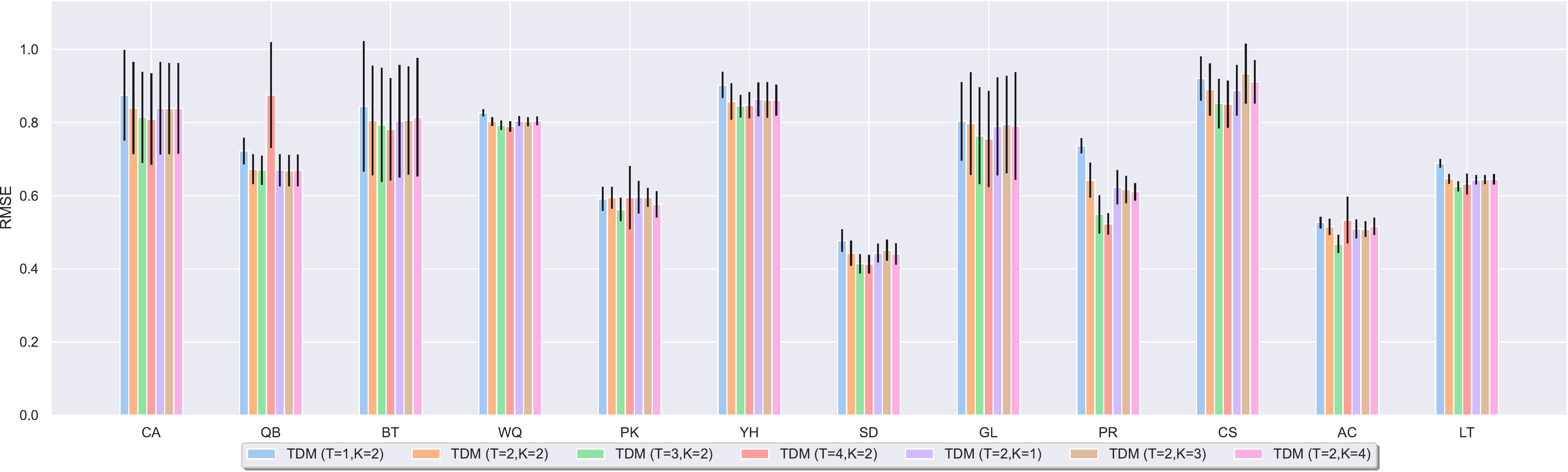}
         \end{subfigure}
         \\
         \begin{subfigure}[b]{0.88\linewidth}
                 \centering                 \includegraphics[width=0.99\textwidth]{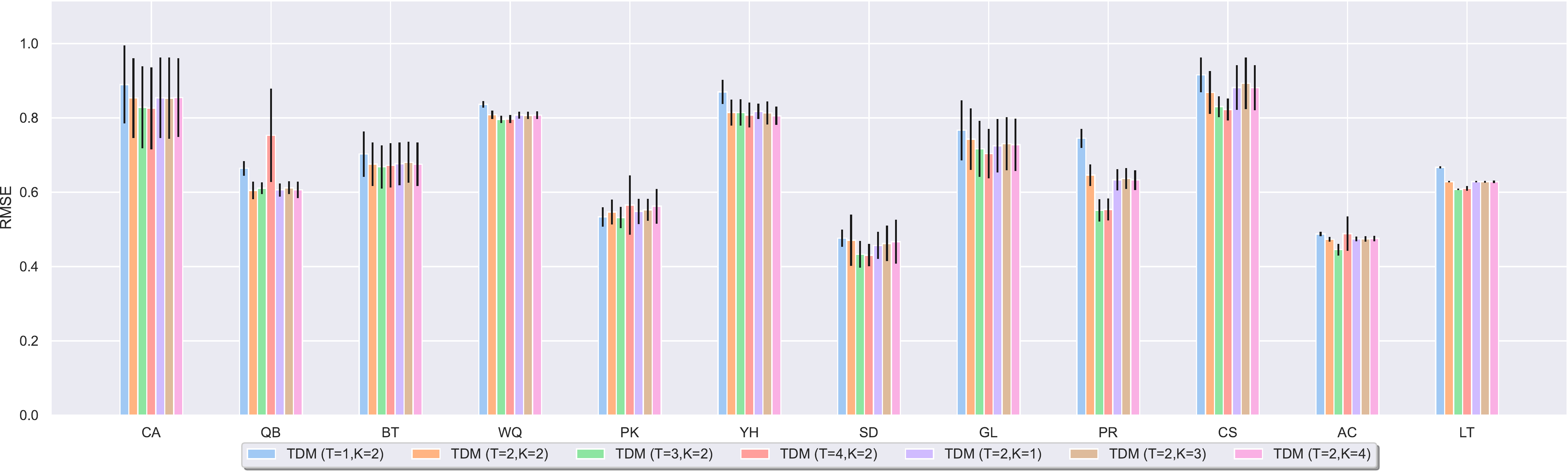}
         \end{subfigure}
 \caption{From top to bottom: RMSE in the MCAR, MAR, MNARL, and MNARQ settings for TDM with different  $T$ and $K$.}
  \label{fig-ab-rmse}
 \vspace{-0.5cm}
\end{figure*}

\begin{figure*}[t]
\captionsetup[subfigure]{justification=centering}
        \centering
         \begin{subfigure}[b]{0.88\linewidth}
                 \centering                 \includegraphics[width=0.99\textwidth]{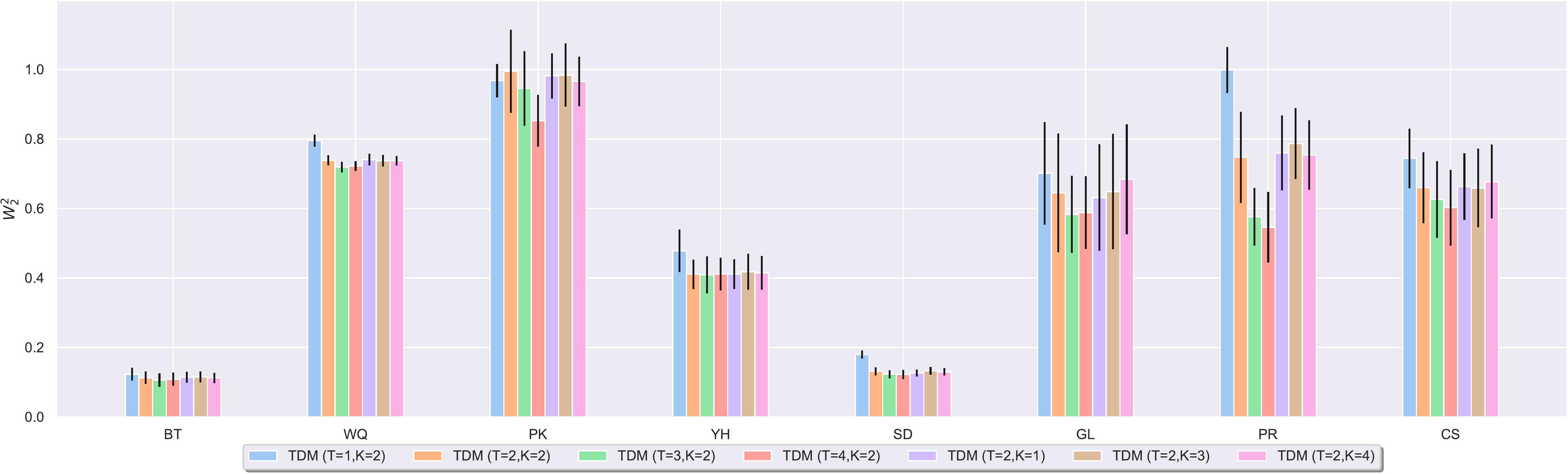}
         \end{subfigure}
         \\
         \begin{subfigure}[b]{0.88\linewidth}
                 \centering                 \includegraphics[width=0.99\textwidth]{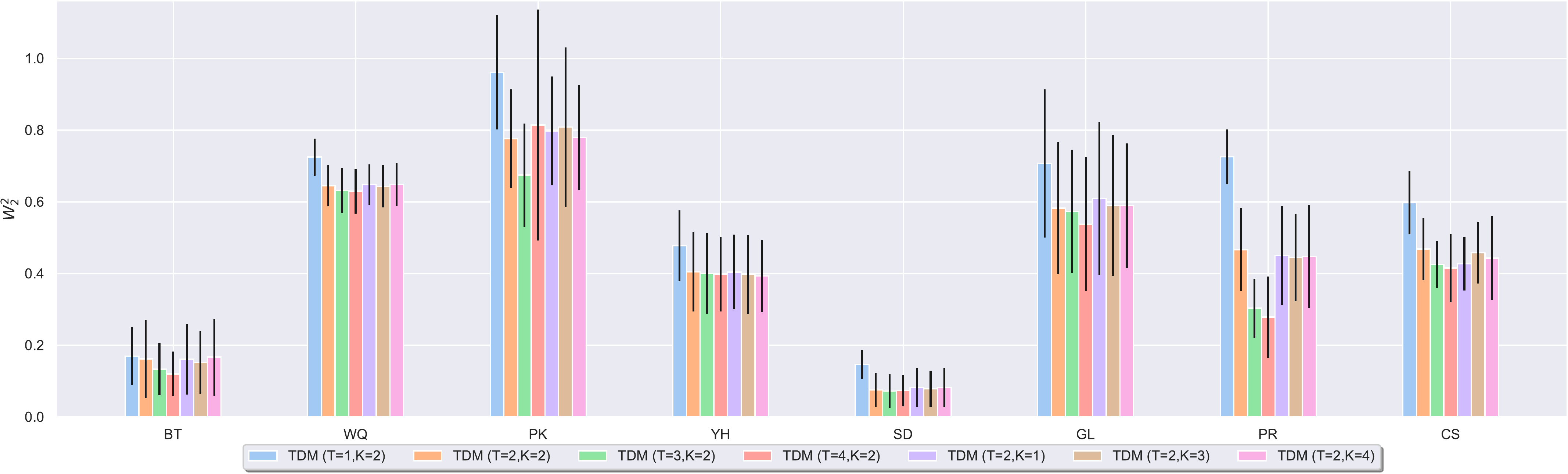}
         \end{subfigure}
        \\
         \begin{subfigure}[b]{0.88\linewidth}
                 \centering                 \includegraphics[width=0.99\textwidth]{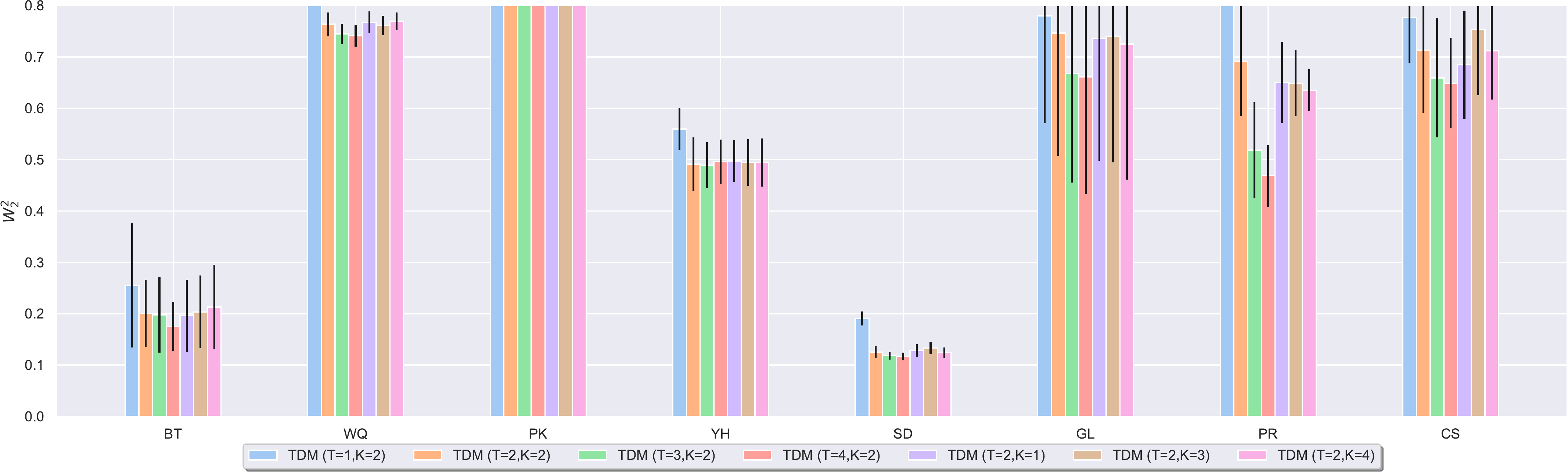}
         \end{subfigure}
         \\
         \begin{subfigure}[b]{0.88\linewidth}
                 \centering                 \includegraphics[width=0.99\textwidth]{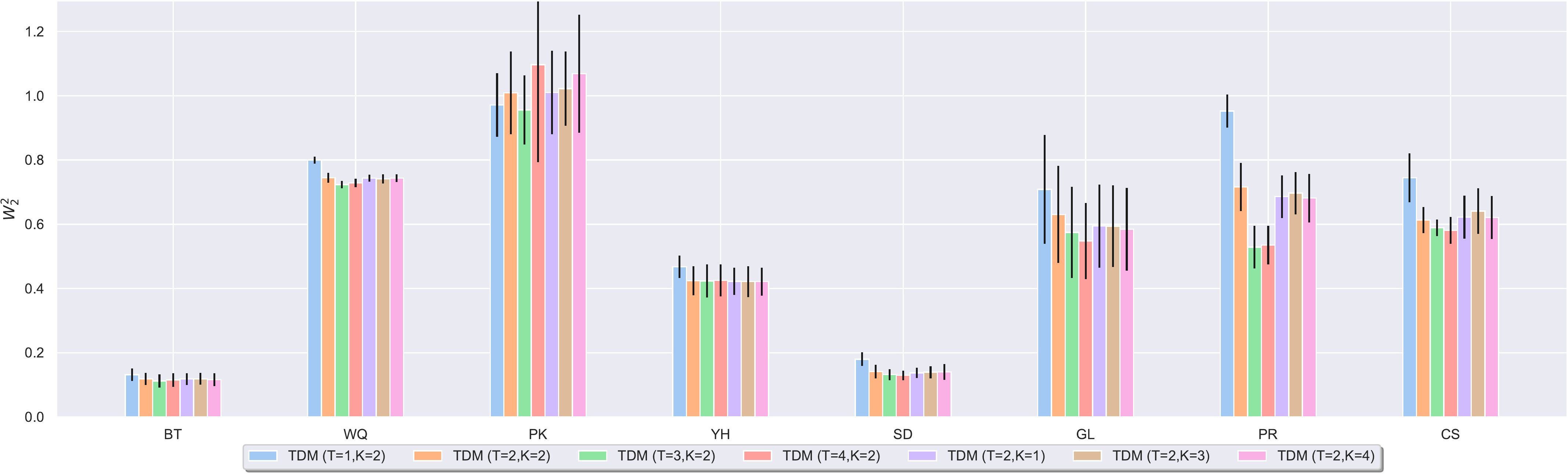}
         \end{subfigure}
 \caption{From top to bottom: $W^2_2$ in the MCAR, MAR, MNARL, and MNARQ settings for TDM with different  $T$ and $K$.}
  \label{fig-ab-ot}
 \vspace{-0.5cm}
\end{figure*}

\begin{figure*}[t]
\captionsetup[subfigure]{justification=centering}
        \centering
         \begin{subfigure}[b]{0.88\linewidth}
                 \centering                 \includegraphics[width=0.99\textwidth]{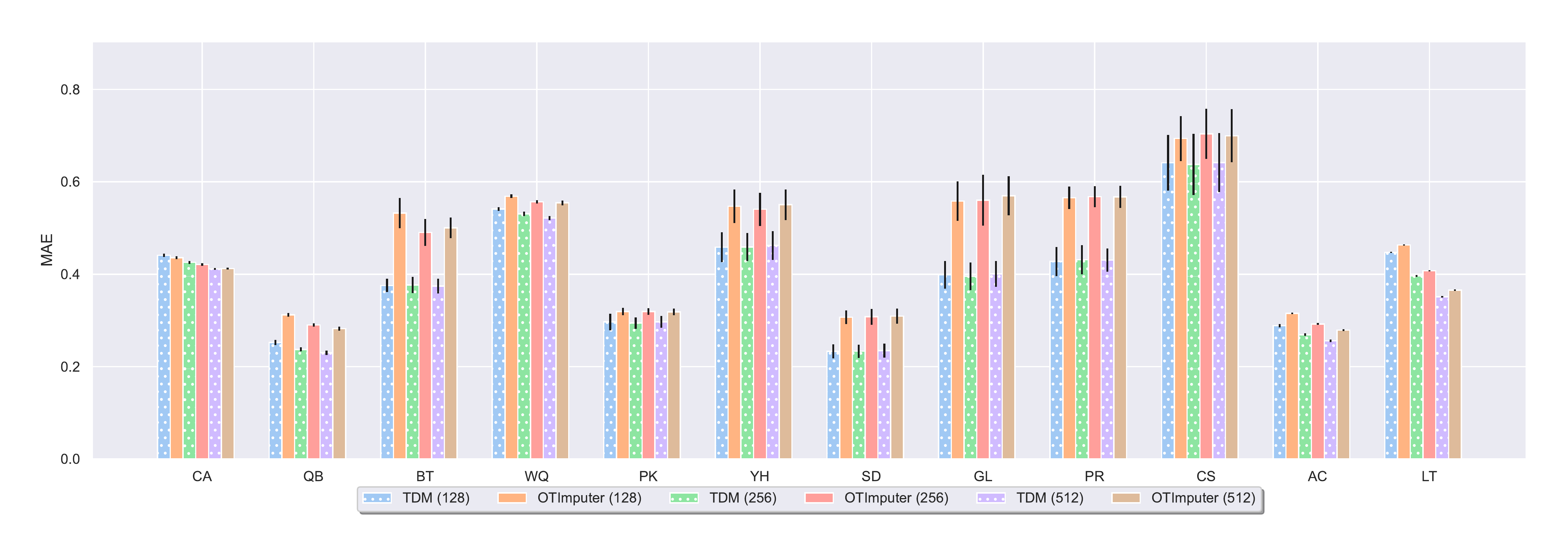}
         \end{subfigure}
         \\
         \begin{subfigure}[b]{0.88\linewidth}
                 \centering                 \includegraphics[width=0.99\textwidth]{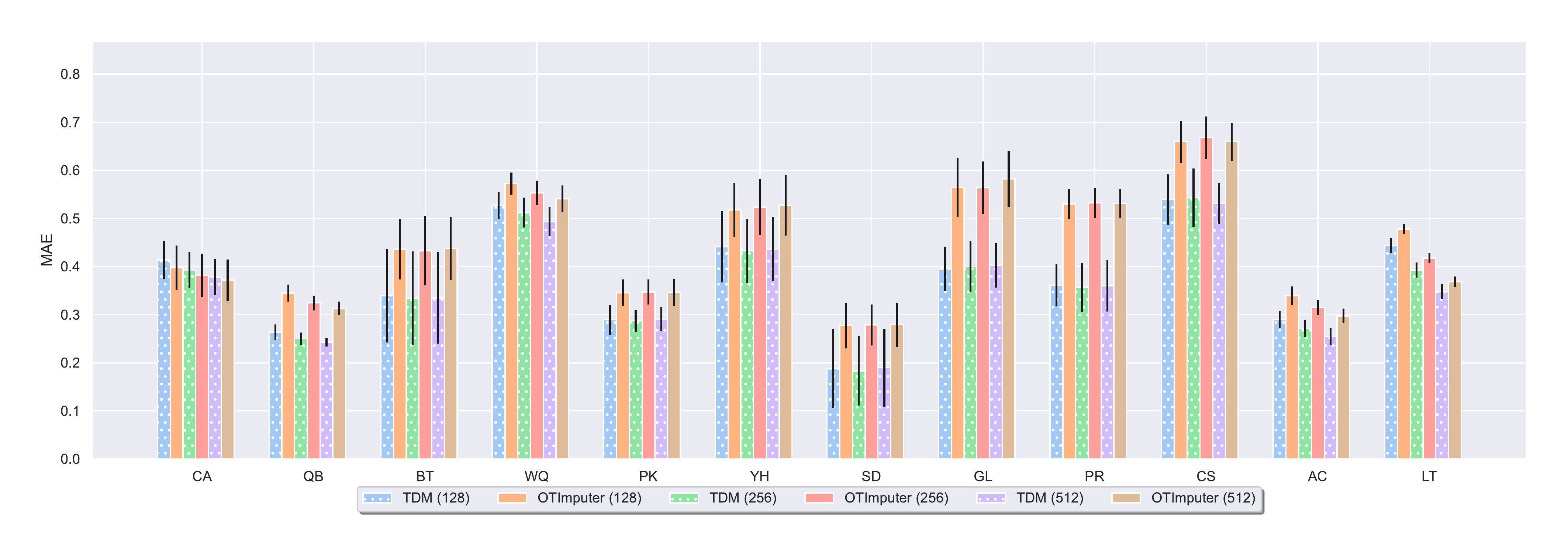}
         \end{subfigure}
         \\
         \begin{subfigure}[b]{0.88\linewidth}
                 \centering                 \includegraphics[width=0.99\textwidth]{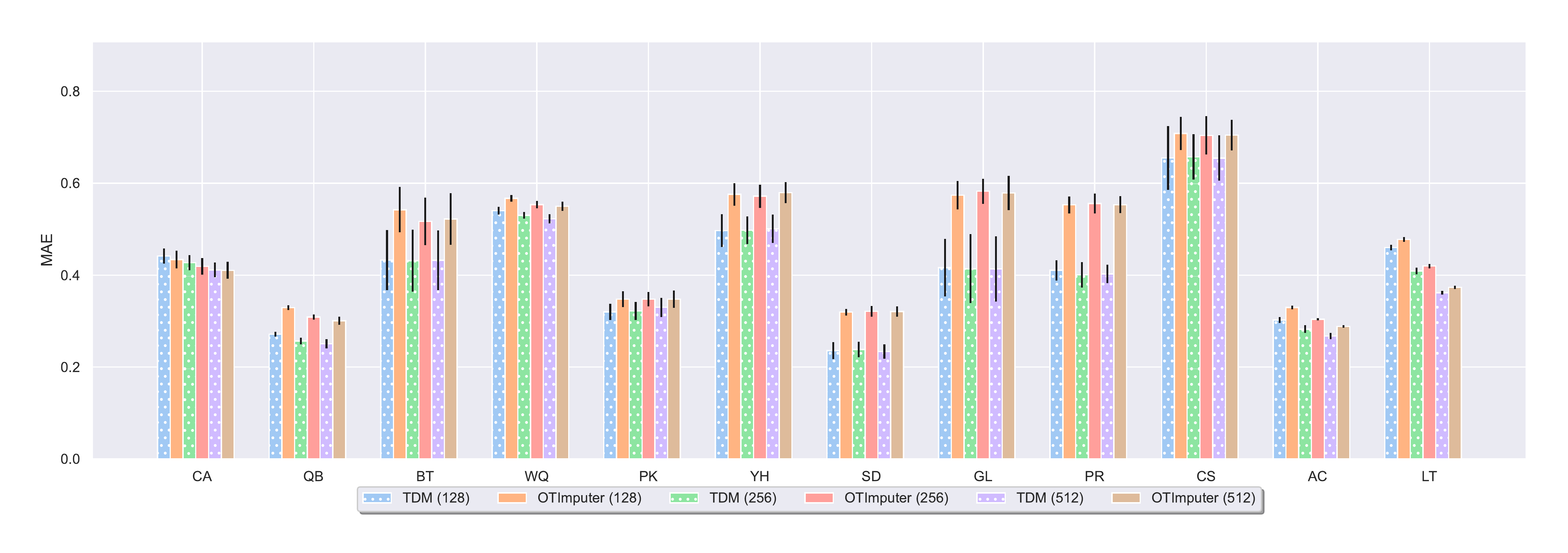}
         \end{subfigure}
         \\ 
         \begin{subfigure}[b]{0.88\linewidth}
                 \centering                 \includegraphics[width=0.99\textwidth]{rebuttal_figs/mae_mnar_l.pdf}
         \end{subfigure}
 \caption{From top to bottom: MAE of TDM and OTImputer with batch size varied in MCAR, MAR, MNARL, and MNARQ.}
  \label{fig-bs-mae}
 \vspace{-0.5cm}
\end{figure*}

\begin{figure*}[t]
\captionsetup[subfigure]{justification=centering}
        \centering
         \begin{subfigure}[b]{0.88\linewidth}
                 \centering                 \includegraphics[width=0.99\textwidth]{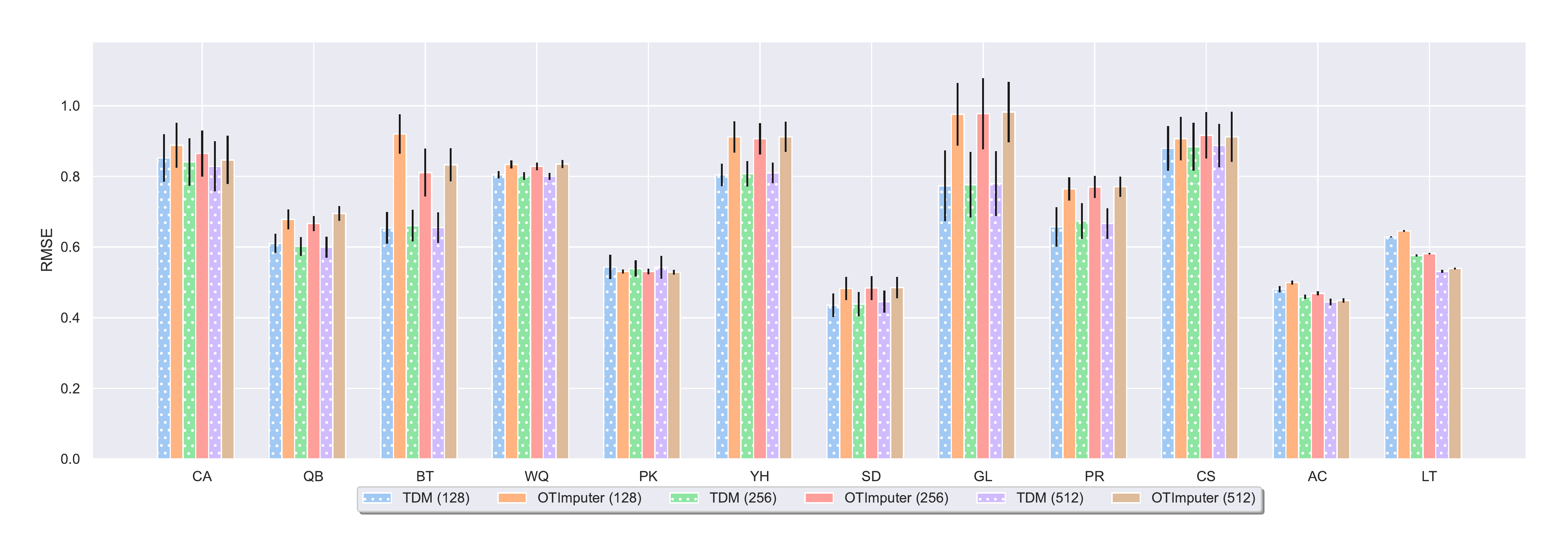}
         \end{subfigure}
         \\
         \begin{subfigure}[b]{0.88\linewidth}
                 \centering                 \includegraphics[width=0.99\textwidth]{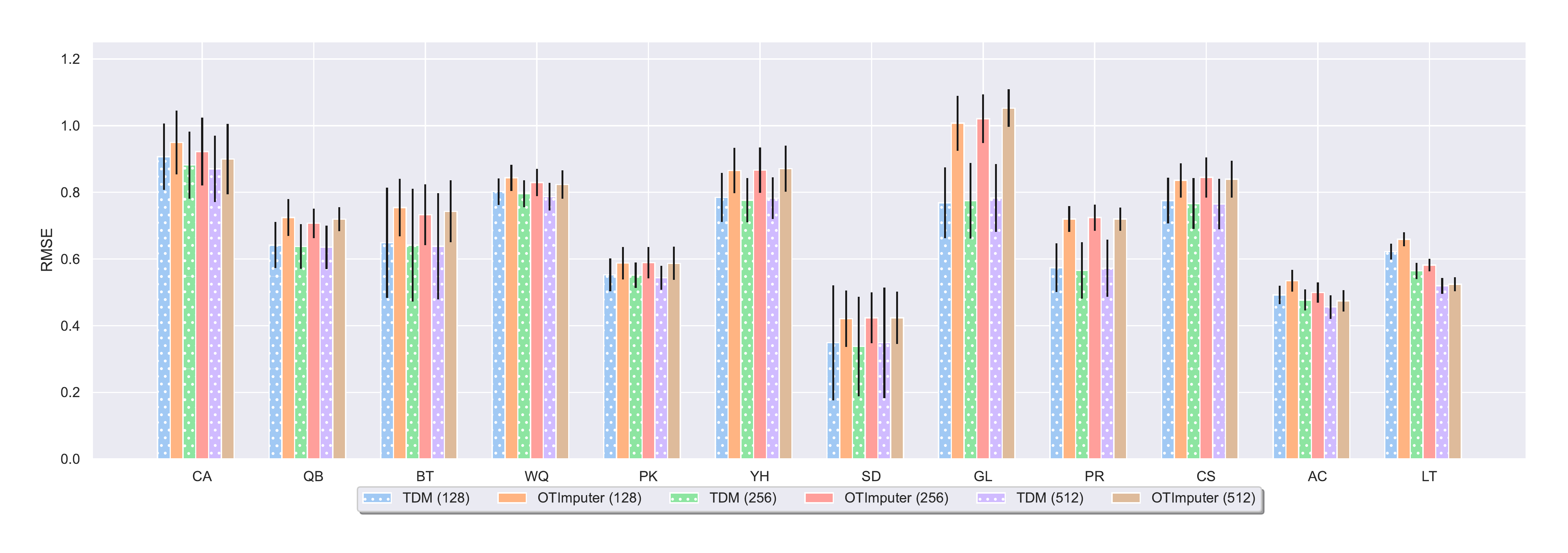}
         \end{subfigure}
         \\
         \begin{subfigure}[b]{0.88\linewidth}
                 \centering                 \includegraphics[width=0.99\textwidth]{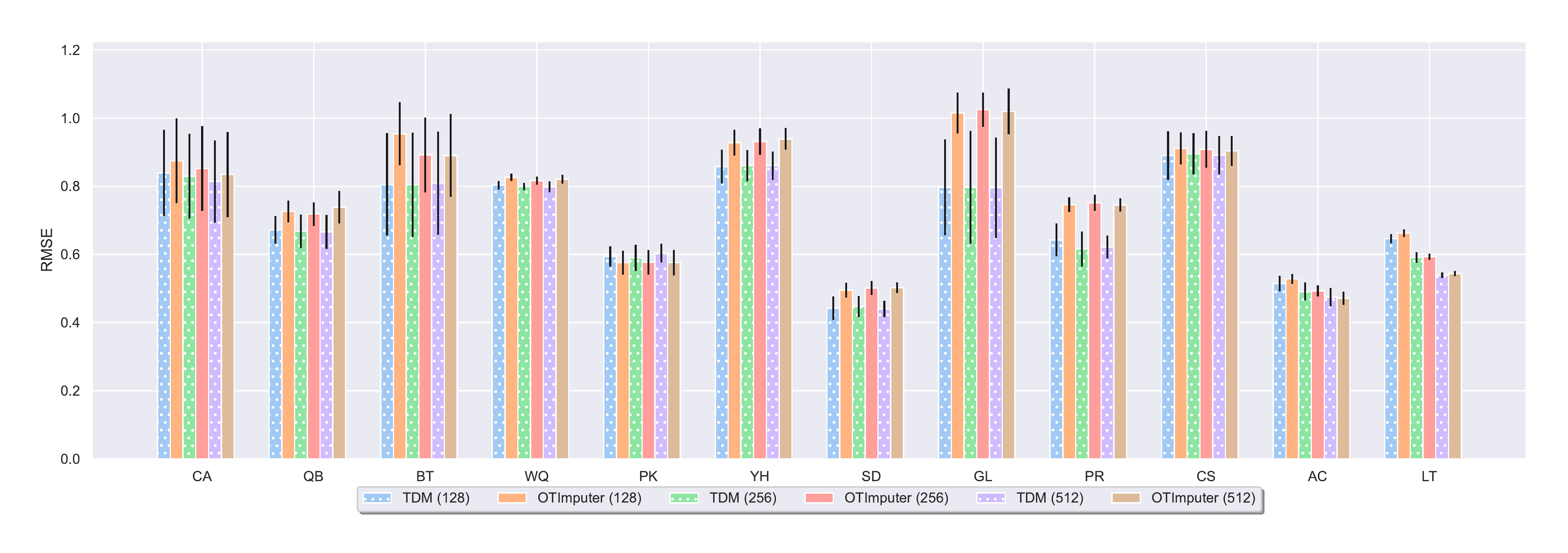}
         \end{subfigure}
         \\ 
         \begin{subfigure}[b]{0.88\linewidth}
                 \centering                 \includegraphics[width=0.99\textwidth]{rebuttal_figs/rmse_mnar_l.pdf}
         \end{subfigure}
 \caption{From top to bottom: RMSE of TDM and OTImputer with batch size varied in MCAR , MAR, MNARL, and MNARQ.}
  \label{fig-bs-rmse}
 \vspace{-0.5cm}
\end{figure*}

\begin{figure*}[t]
\captionsetup[subfigure]{justification=centering}
        \centering
         \begin{subfigure}[b]{0.88\linewidth}
                 \centering                 \includegraphics[width=0.99\textwidth]{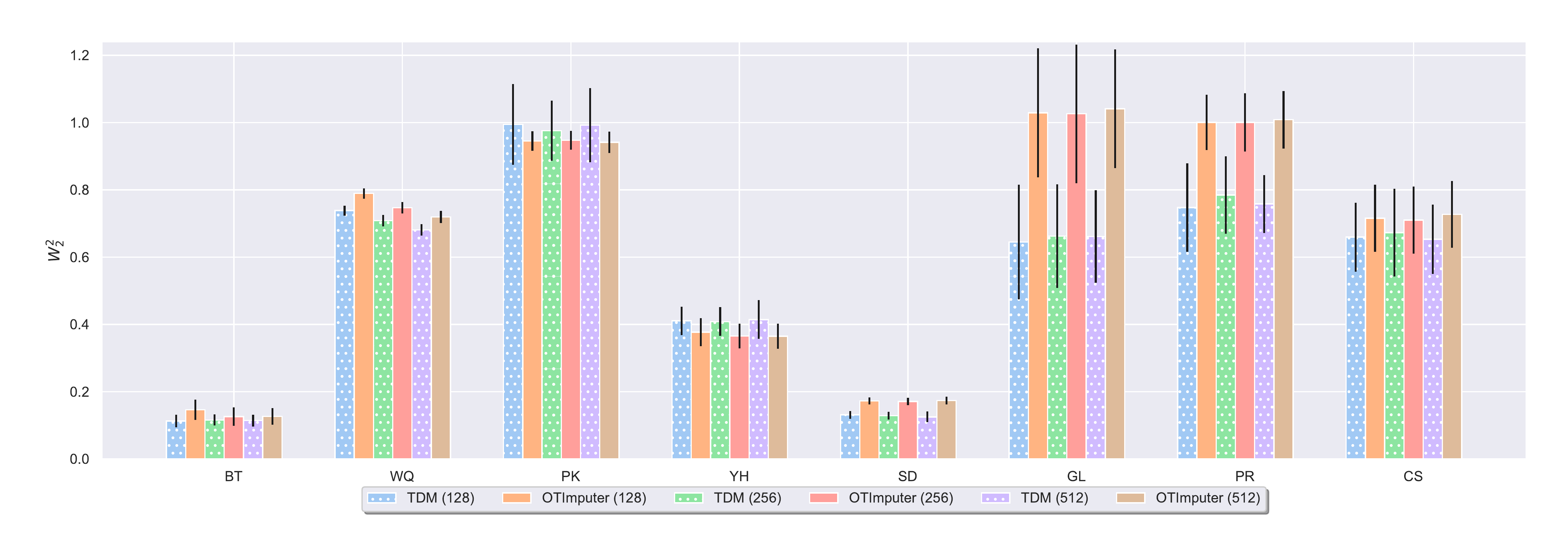}
         \end{subfigure}
         \\
         \begin{subfigure}[b]{0.88\linewidth}
                 \centering                 \includegraphics[width=0.99\textwidth]{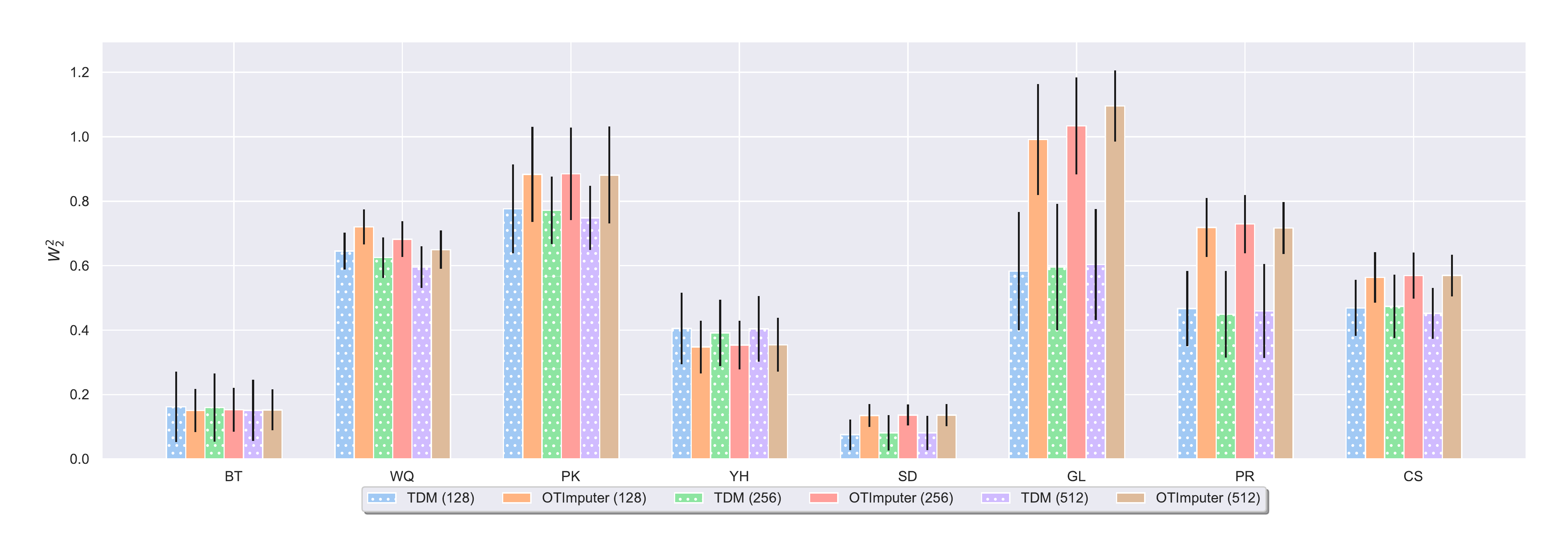}
         \end{subfigure}
         \\
         \begin{subfigure}[b]{0.88\linewidth}
                 \centering                 \includegraphics[width=0.99\textwidth]{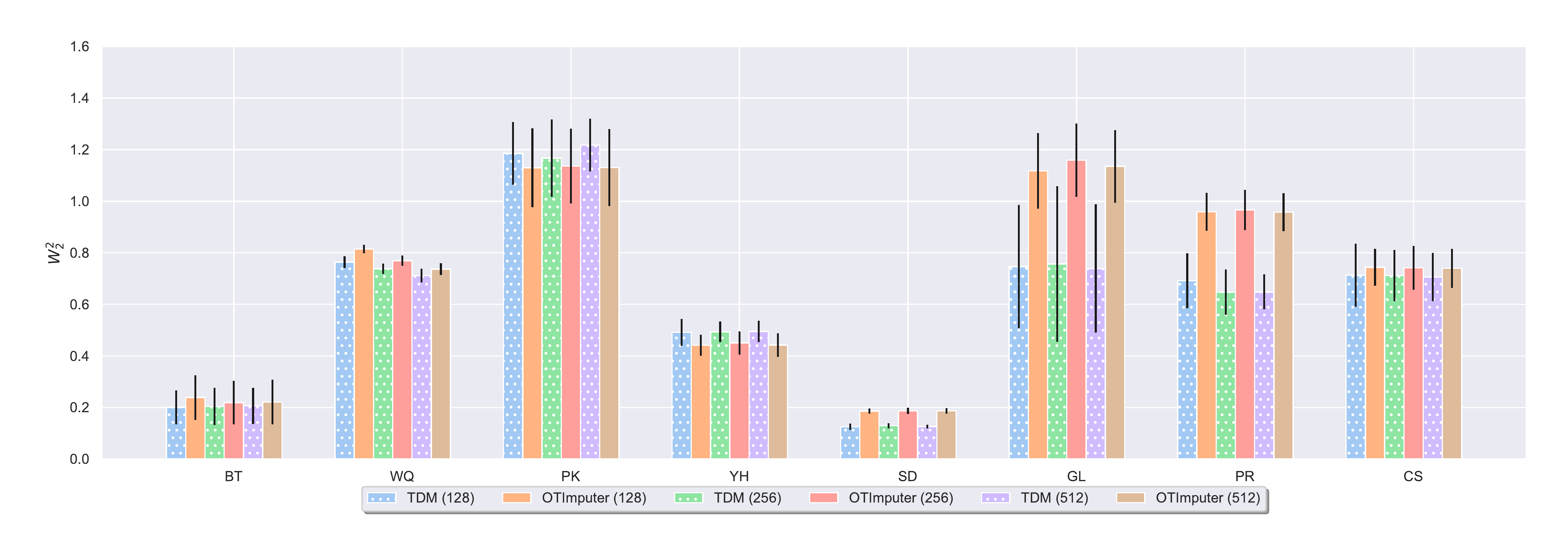}
         \end{subfigure}
         \\ 
         \begin{subfigure}[b]{0.88\linewidth}
                 \centering                 \includegraphics[width=0.99\textwidth]{rebuttal_figs/ot_mnar_l.pdf}
         \end{subfigure}
 \caption{From top to bottom: $W^2_2$ of TDM and OTImputer with batch size varied in MCAR , MAR, MNARL, and MNARQ.}
  \label{fig-bs-ot}
 \vspace{-0.5cm}
\end{figure*}

\begin{figure}[t]
\captionsetup[subfigure]{justification=centering}
        \centering
         \begin{subfigure}[b]{0.245\linewidth}
                 \centering 
                 \includegraphics[width=0.99\textwidth]{figs/iter/glass_mcar_mae-crop.pdf}
         \end{subfigure}
                 \begin{subfigure}[b]{0.245\linewidth}
                 \centering                 \includegraphics[width=0.99\textwidth]{figs/iter/seeds_mcar_mae-crop.pdf}
         \end{subfigure}
          \begin{subfigure}[b]{0.245\linewidth}
                 \centering                 \includegraphics[width=0.99\textwidth]{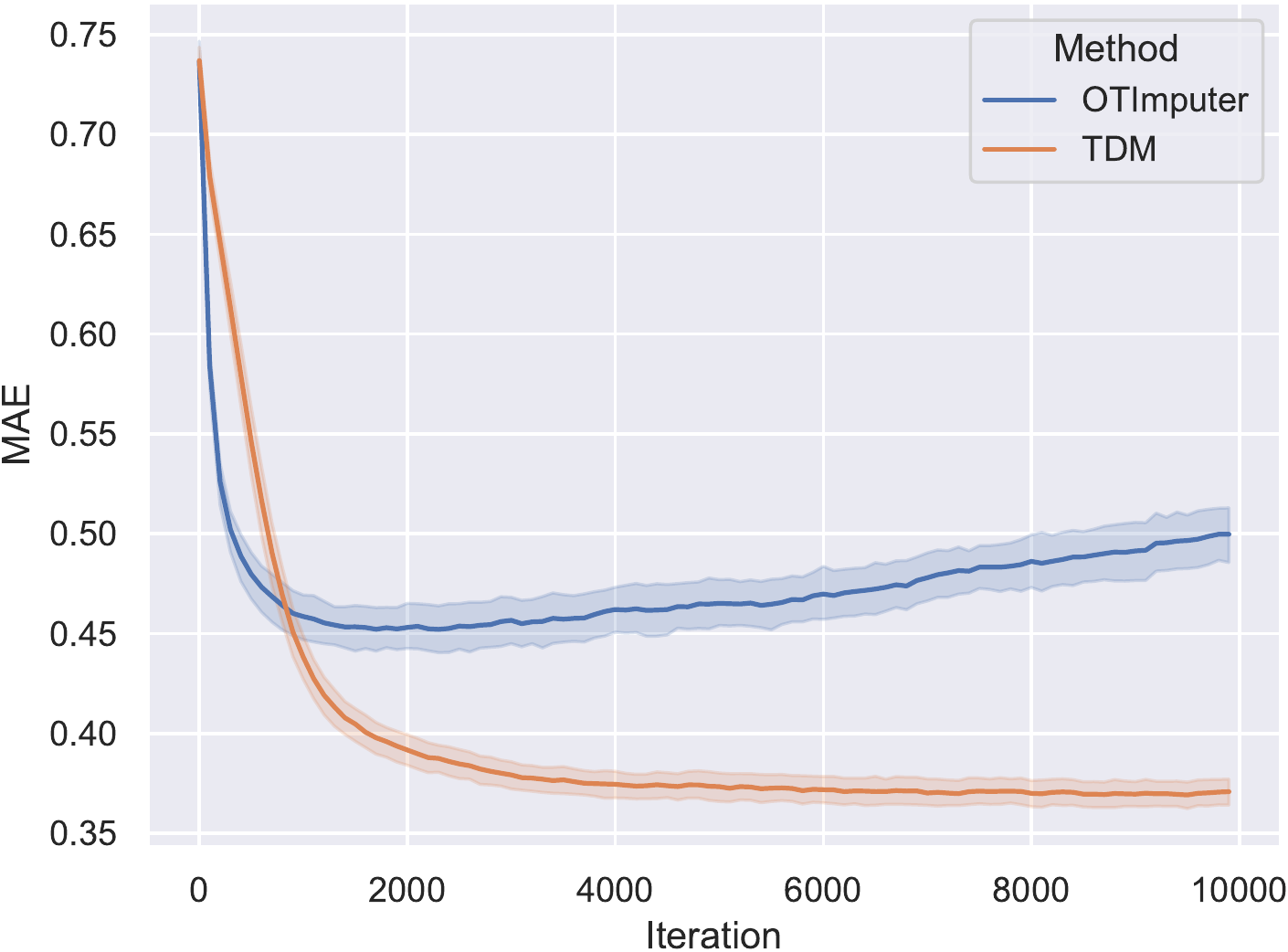}
         \end{subfigure}
                  \begin{subfigure}[b]{0.245\linewidth}
                 \centering                 \includegraphics[width=0.99\textwidth]{figs/iter/anuran_calls_mcar_mae-crop.pdf}
         \end{subfigure}\\
          \begin{subfigure}[b]{0.245\linewidth}
                 \centering                 \includegraphics[width=0.99\textwidth]{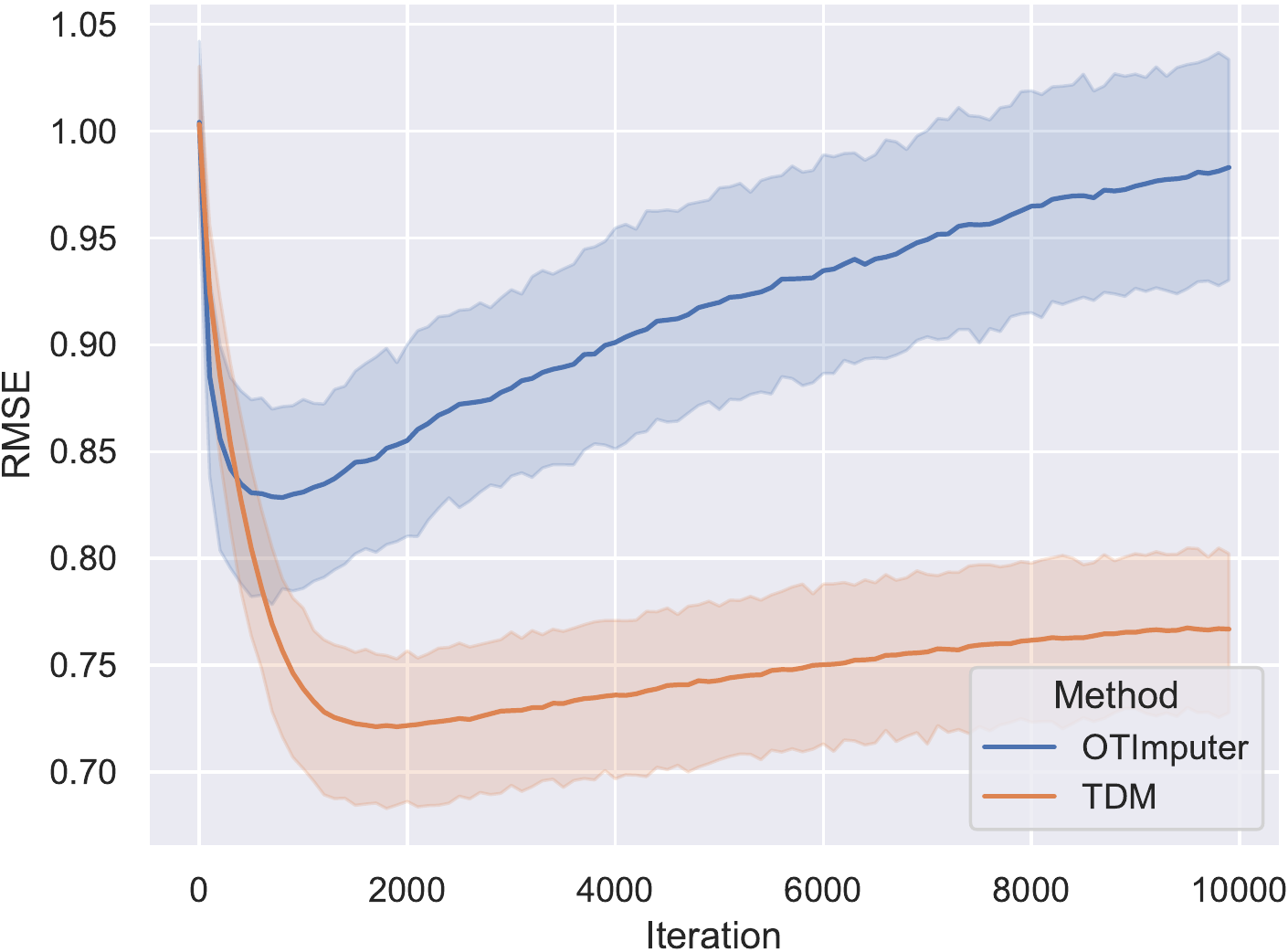}
         \end{subfigure}
                 \begin{subfigure}[b]{0.245\linewidth}
                 \centering                 \includegraphics[width=0.99\textwidth]{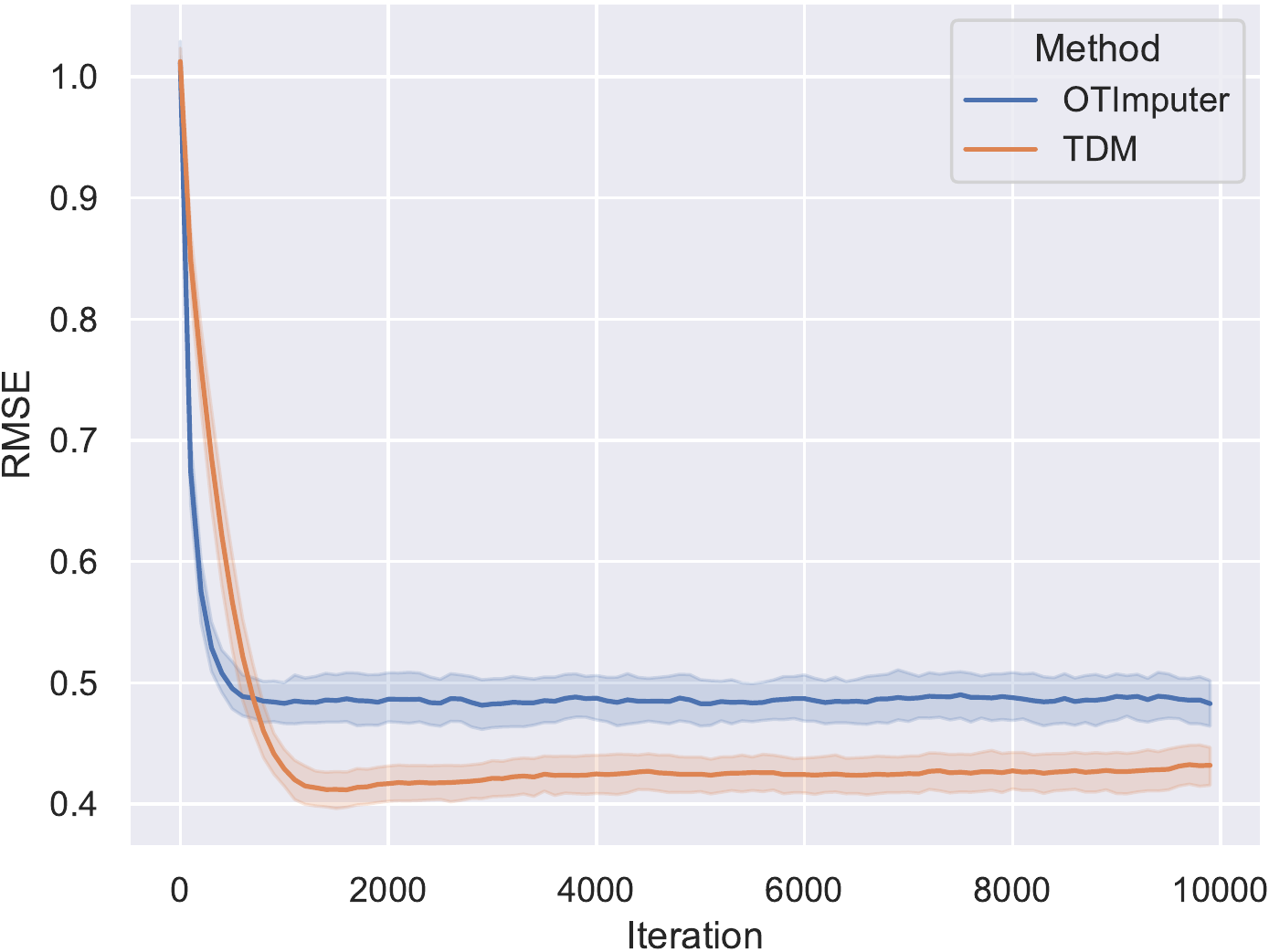}
         \end{subfigure}
          \begin{subfigure}[b]{0.245\linewidth}
                 \centering                 \includegraphics[width=0.99\textwidth]{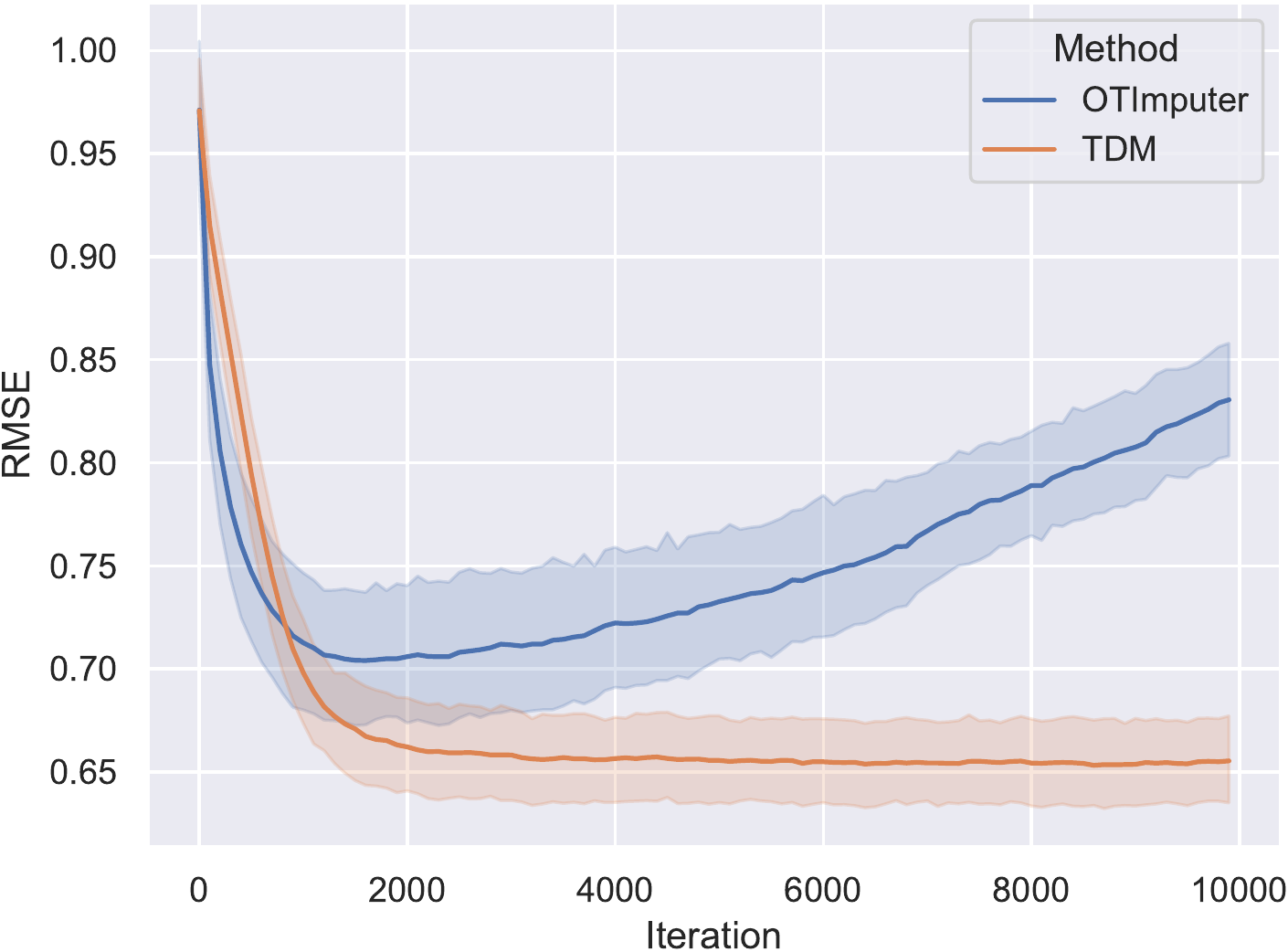}
         \end{subfigure}
                  \begin{subfigure}[b]{0.245\linewidth}
                 \centering                 \includegraphics[width=0.99\textwidth]{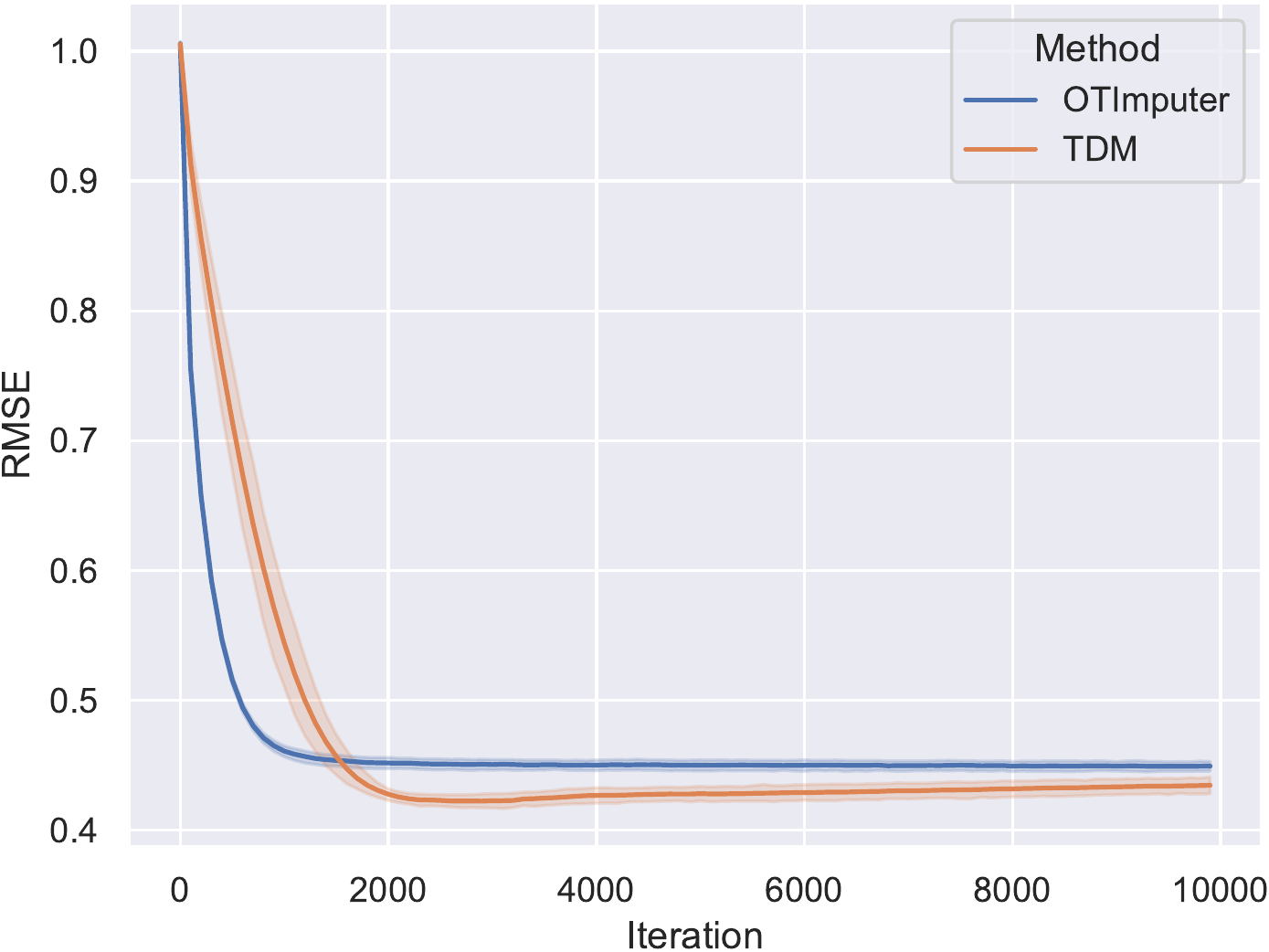}
         \end{subfigure}
 \caption{MAE and RMSE over training iterations of TDM and OTImputer on four datasets (from left to right: glass, seeds, blood\_transfusion, anuran\_calls) in MCAR.}
  \label{fig-iter-mcar}
 \vspace{-0.5cm}
\end{figure}

\begin{figure}[t]
\captionsetup[subfigure]{justification=centering}
        \centering
         \begin{subfigure}[b]{0.245\linewidth}
                 \centering                 \includegraphics[width=0.99\textwidth]{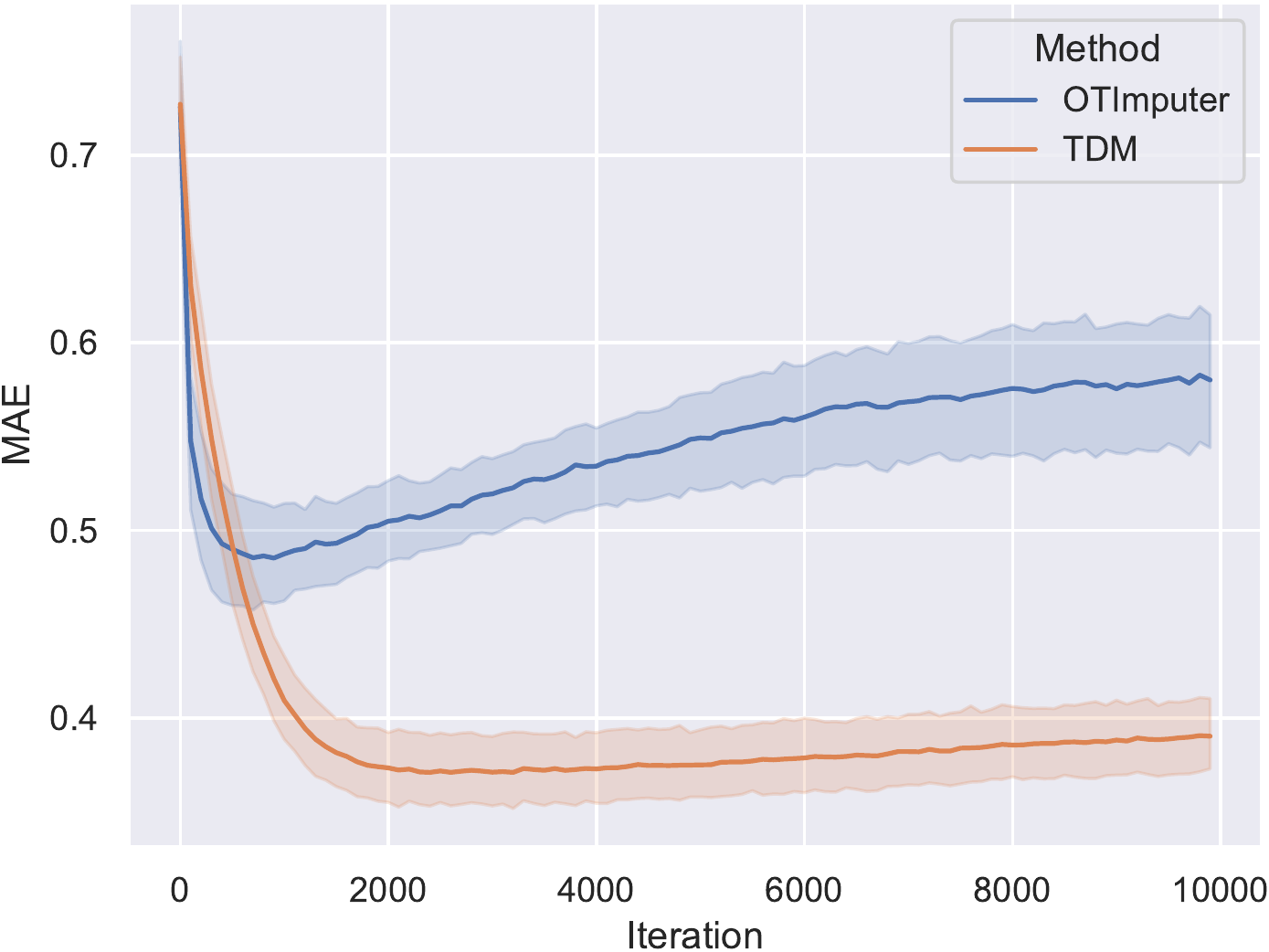}
         \end{subfigure}
                 \begin{subfigure}[b]{0.245\linewidth}
                 \centering                 \includegraphics[width=0.99\textwidth]{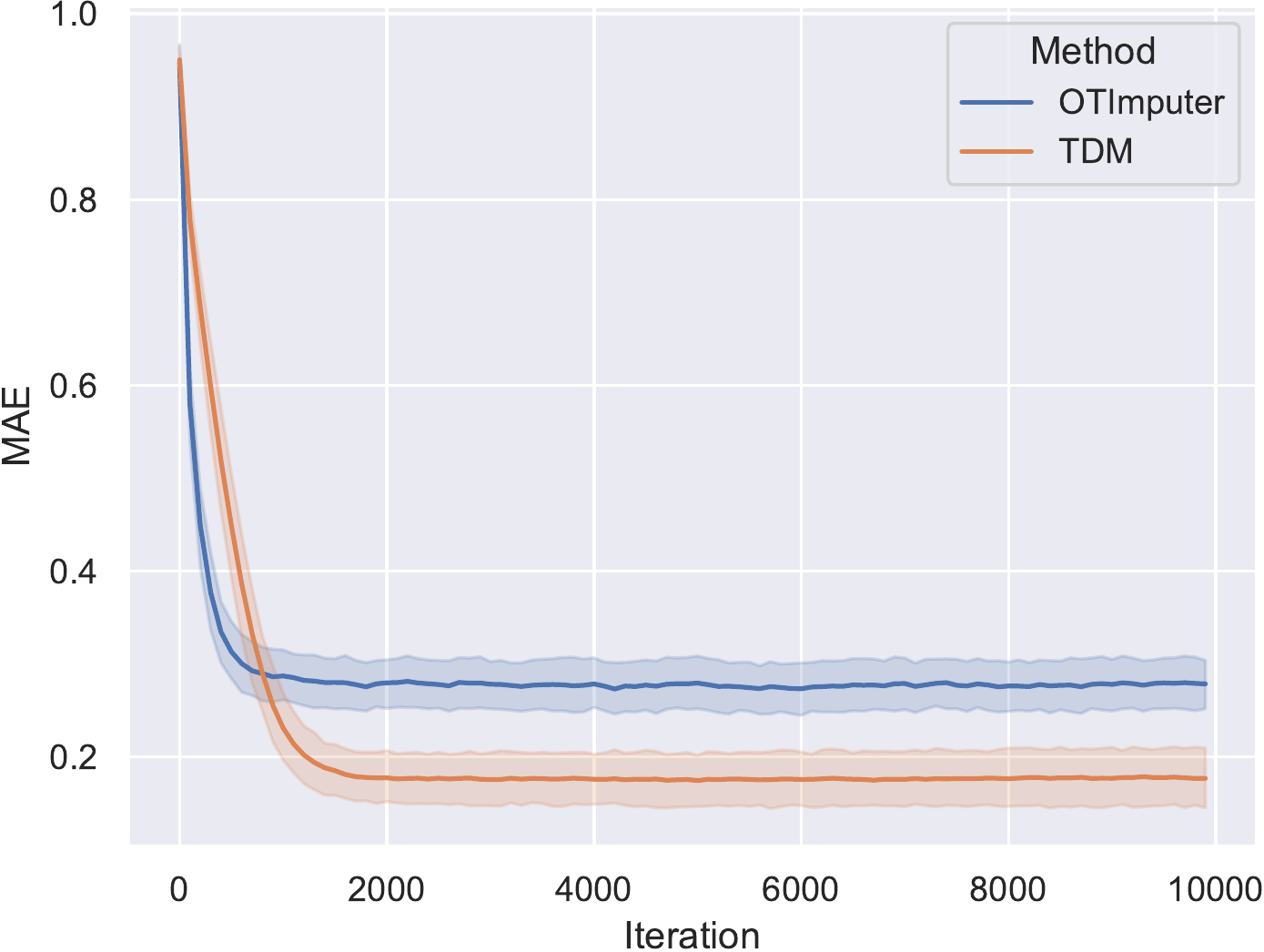}
         \end{subfigure}
          \begin{subfigure}[b]{0.245\linewidth}
                 \centering                 \includegraphics[width=0.99\textwidth]{figs/iter/blood_transfusion_mar_mae-crop.pdf}
         \end{subfigure}
                  \begin{subfigure}[b]{0.245\linewidth}
                 \centering                 \includegraphics[width=0.99\textwidth]{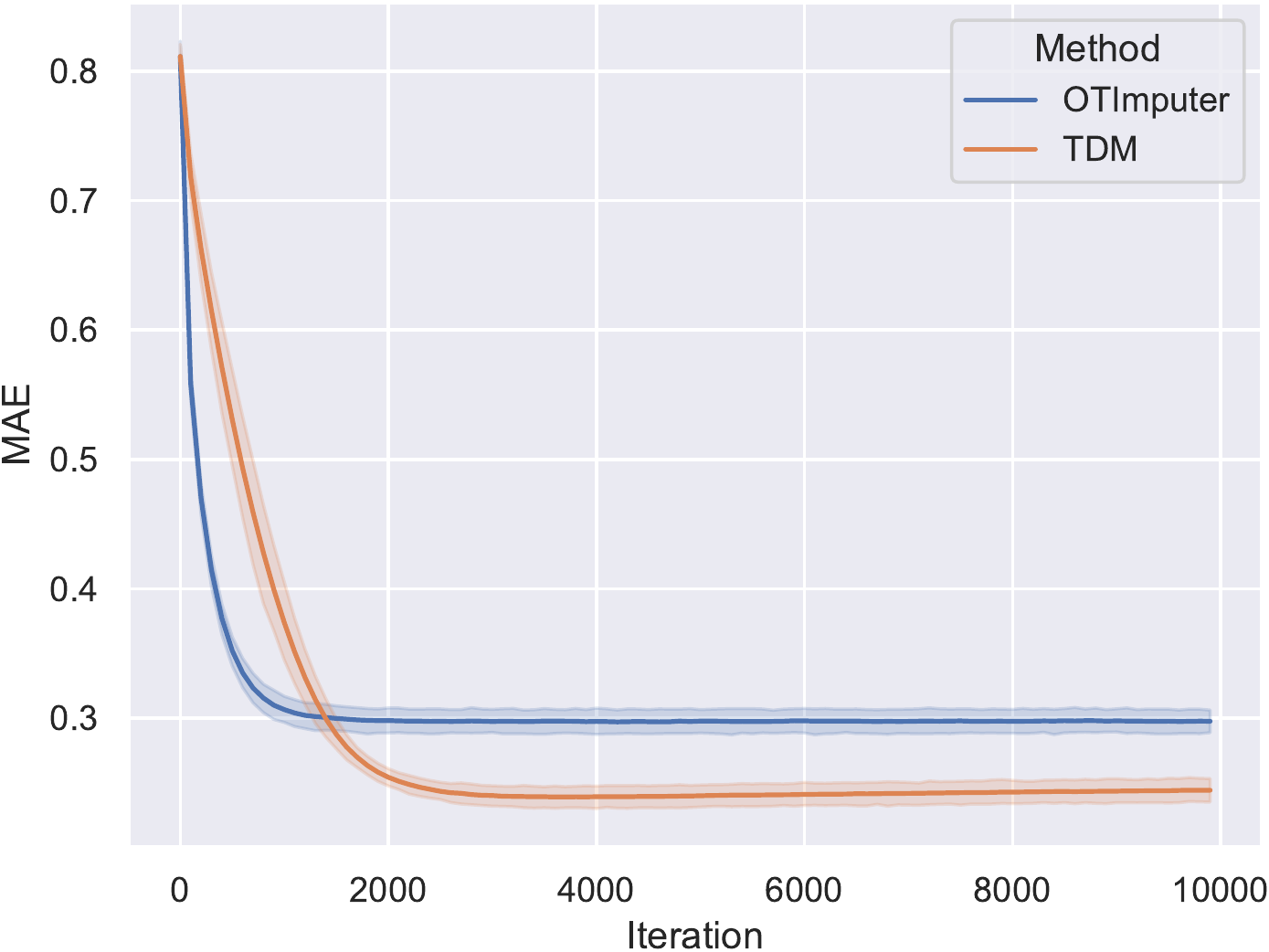}
         \end{subfigure}\\
          \begin{subfigure}[b]{0.245\linewidth}
                 \centering                 \includegraphics[width=0.99\textwidth]{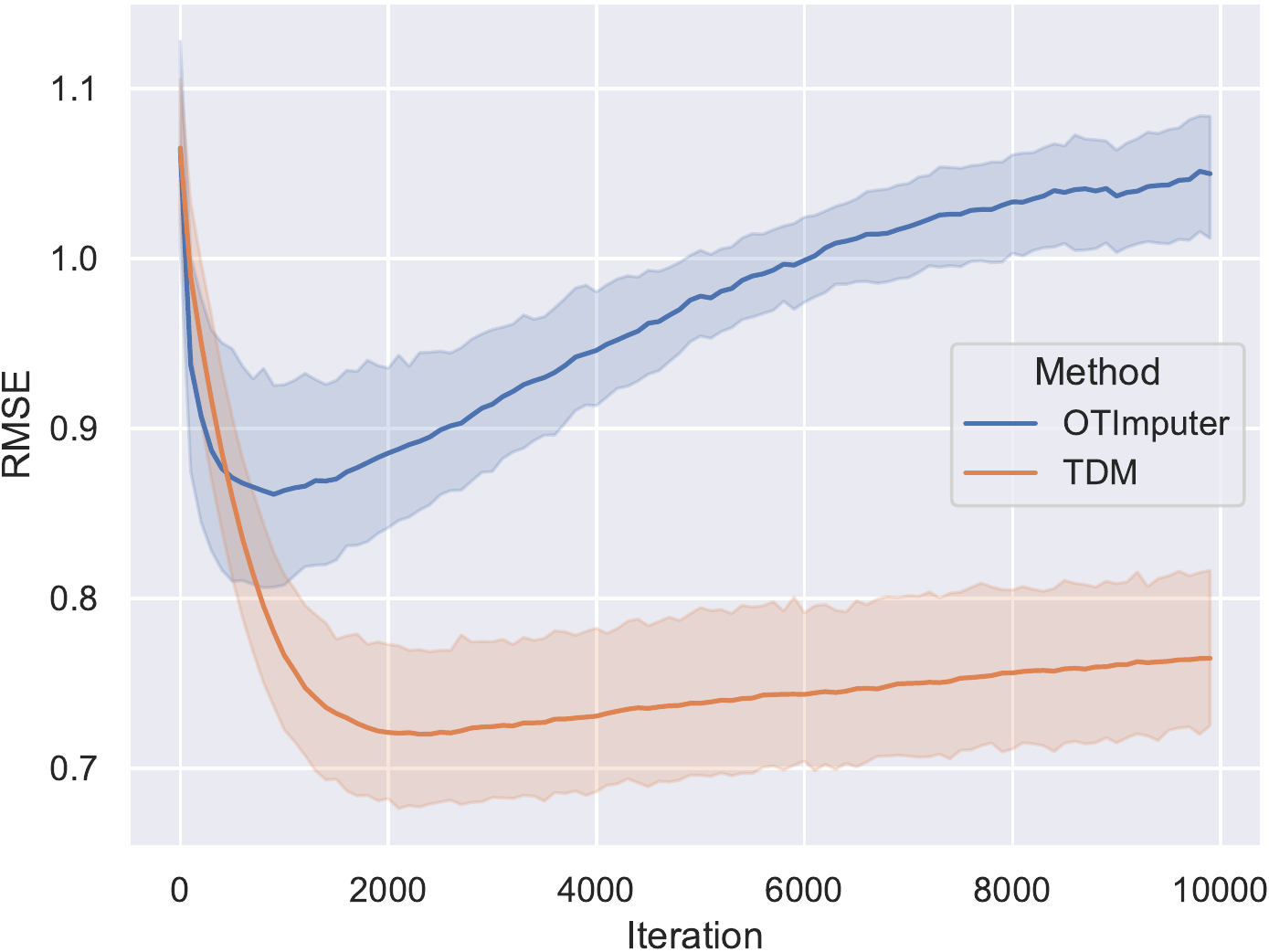}
         \end{subfigure}
                 \begin{subfigure}[b]{0.245\linewidth}
                 \centering                 \includegraphics[width=0.99\textwidth]{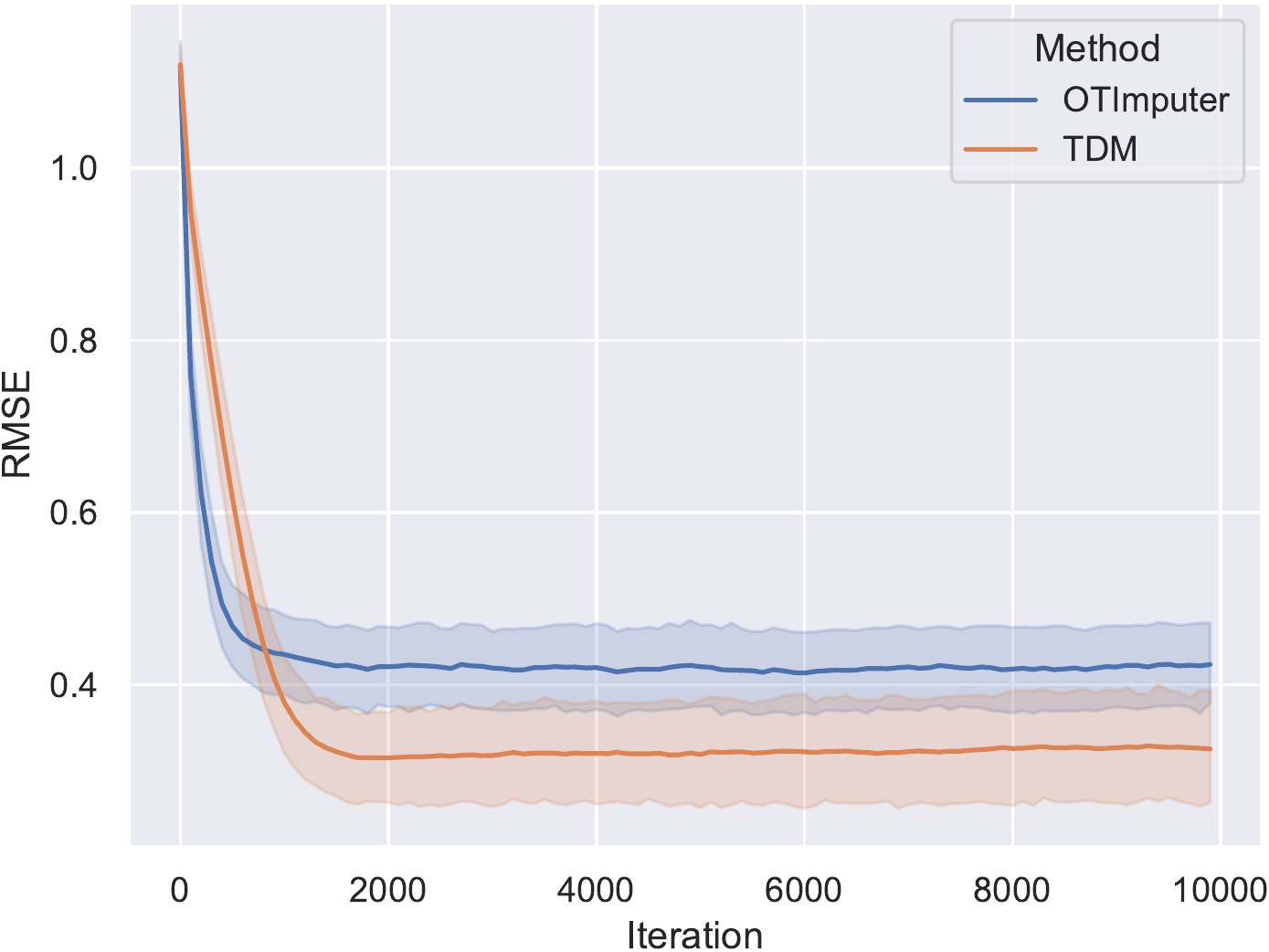}
         \end{subfigure}
          \begin{subfigure}[b]{0.245\linewidth}
                 \centering                 \includegraphics[width=0.99\textwidth]{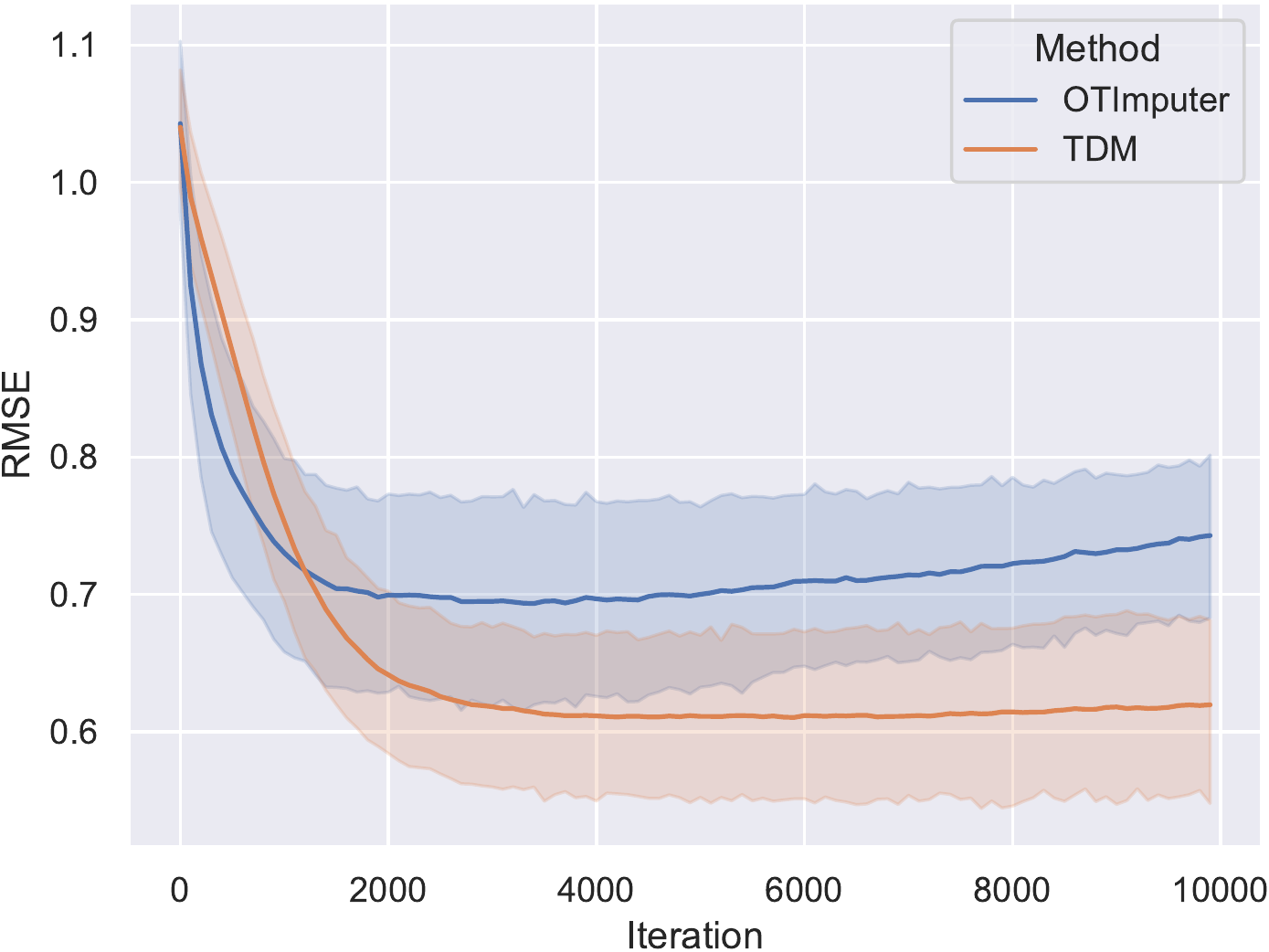}
         \end{subfigure}
                  \begin{subfigure}[b]{0.245\linewidth}
                 \centering                 \includegraphics[width=0.99\textwidth]{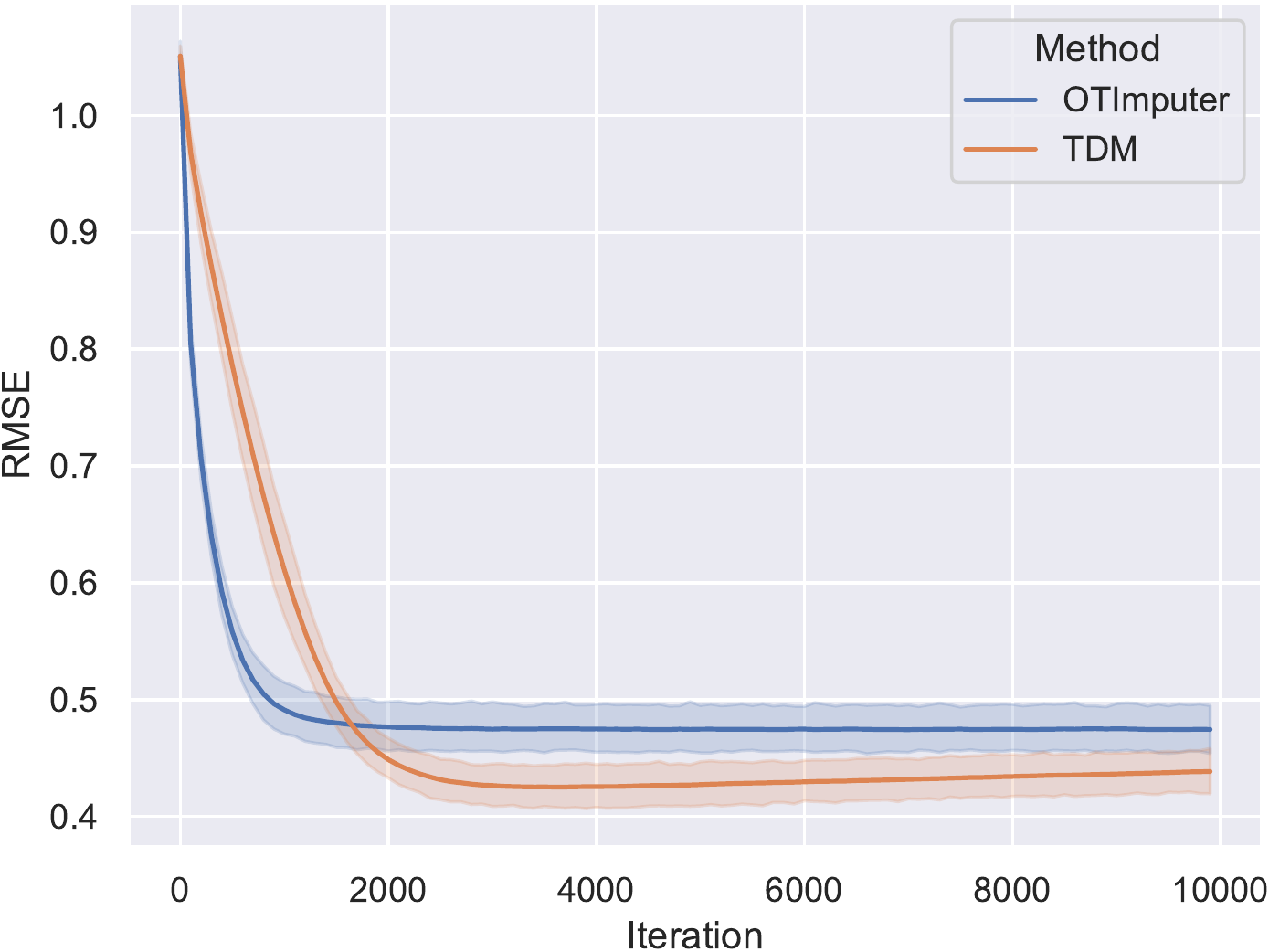}
         \end{subfigure}
 \caption{MAE and RMSE over training iterations of TDM and OTImputer on four datasets (from left to right: glass, seeds, blood\_transfusion, anuran\_calls) in MAR.}
  \label{fig-iter-mar}
 \vspace{-0.5cm}
\end{figure}

\begin{figure}[t]
\captionsetup[subfigure]{justification=centering}
        \centering
         \begin{subfigure}[b]{0.245\linewidth}
                 \centering                 \includegraphics[width=0.99\textwidth]{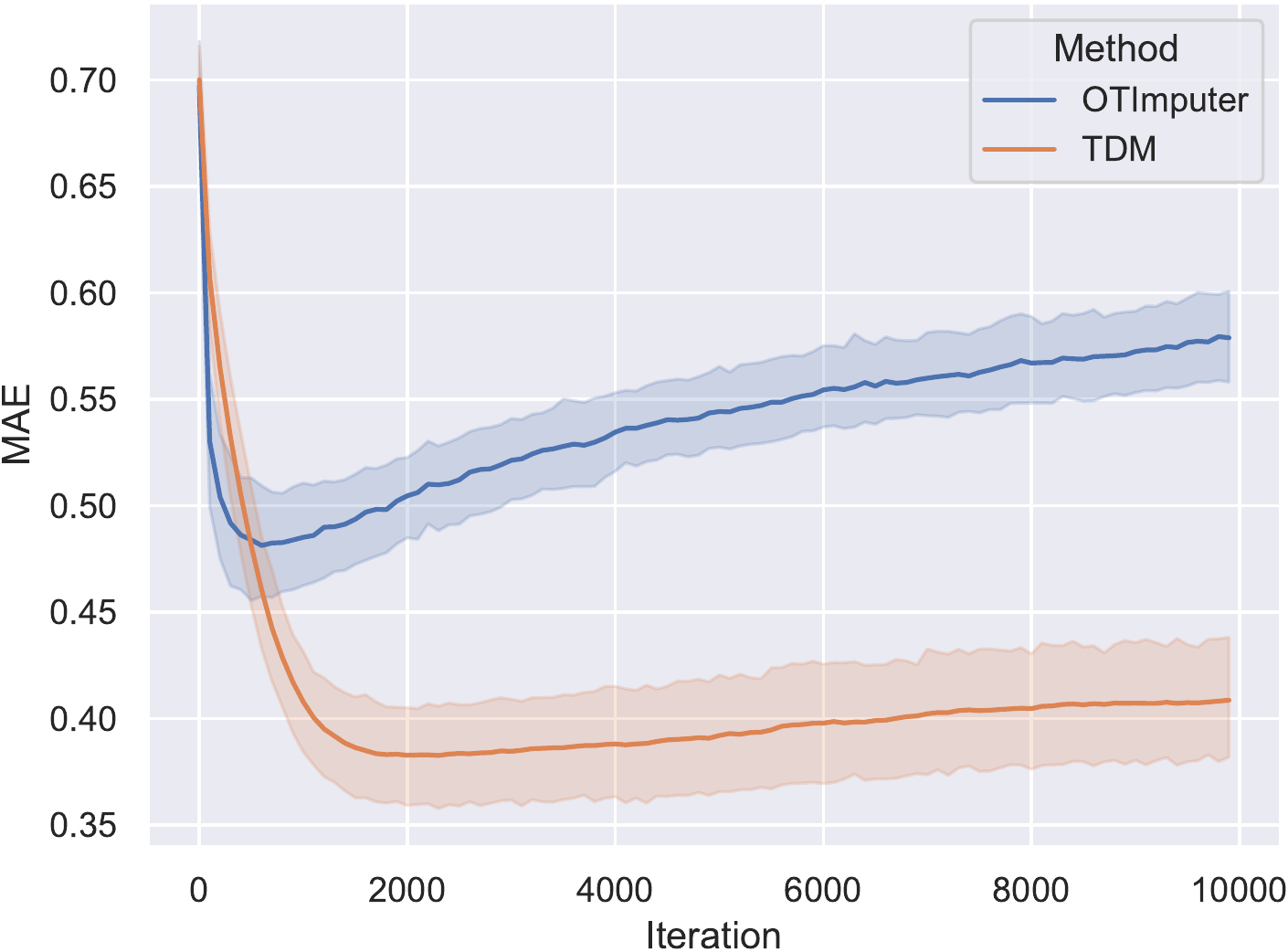}
         \end{subfigure}
                 \begin{subfigure}[b]{0.245\linewidth}
                 \centering                 \includegraphics[width=0.99\textwidth]{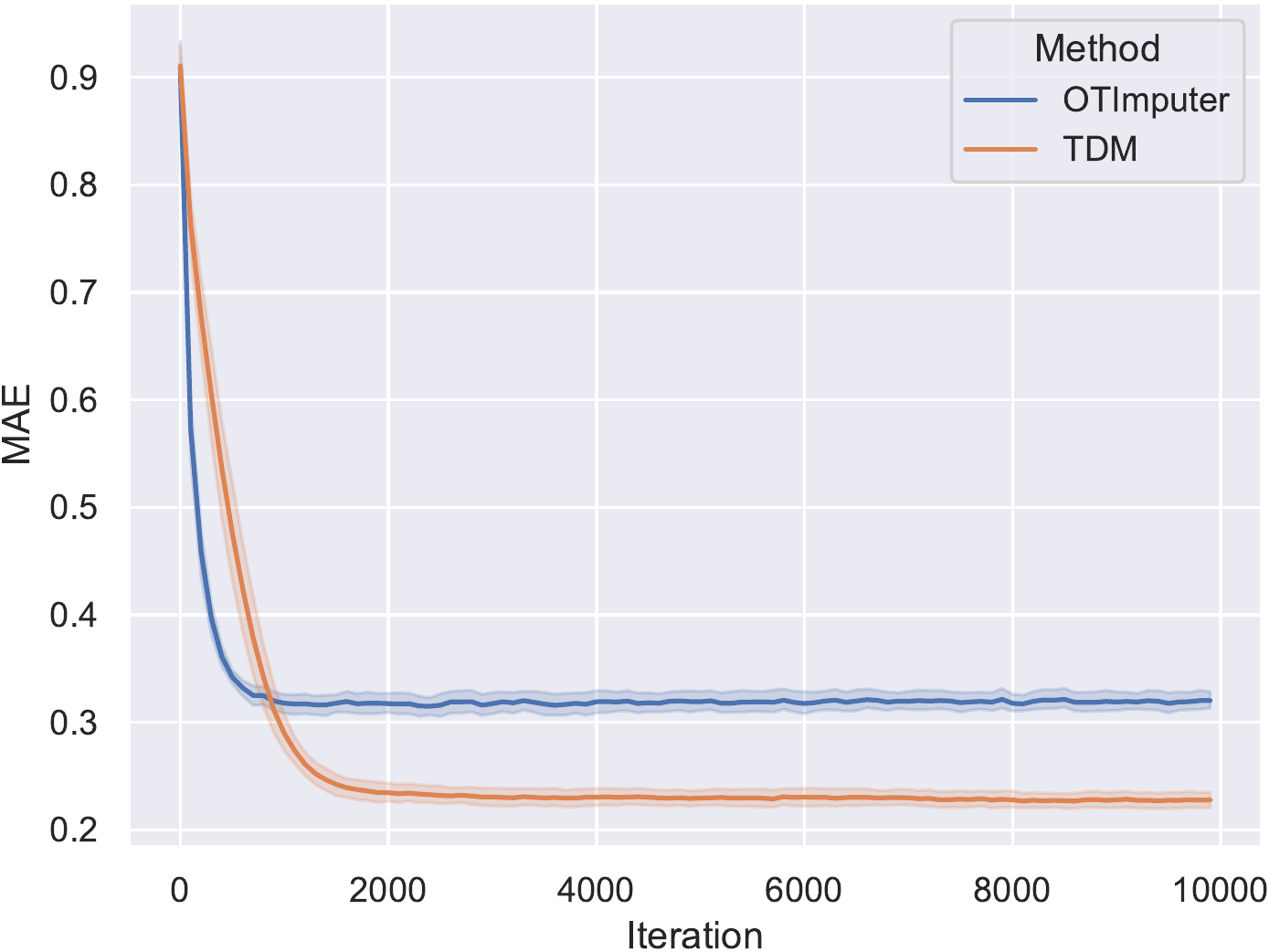}
         \end{subfigure}
          \begin{subfigure}[b]{0.245\linewidth}
                 \centering                 \includegraphics[width=0.99\textwidth]{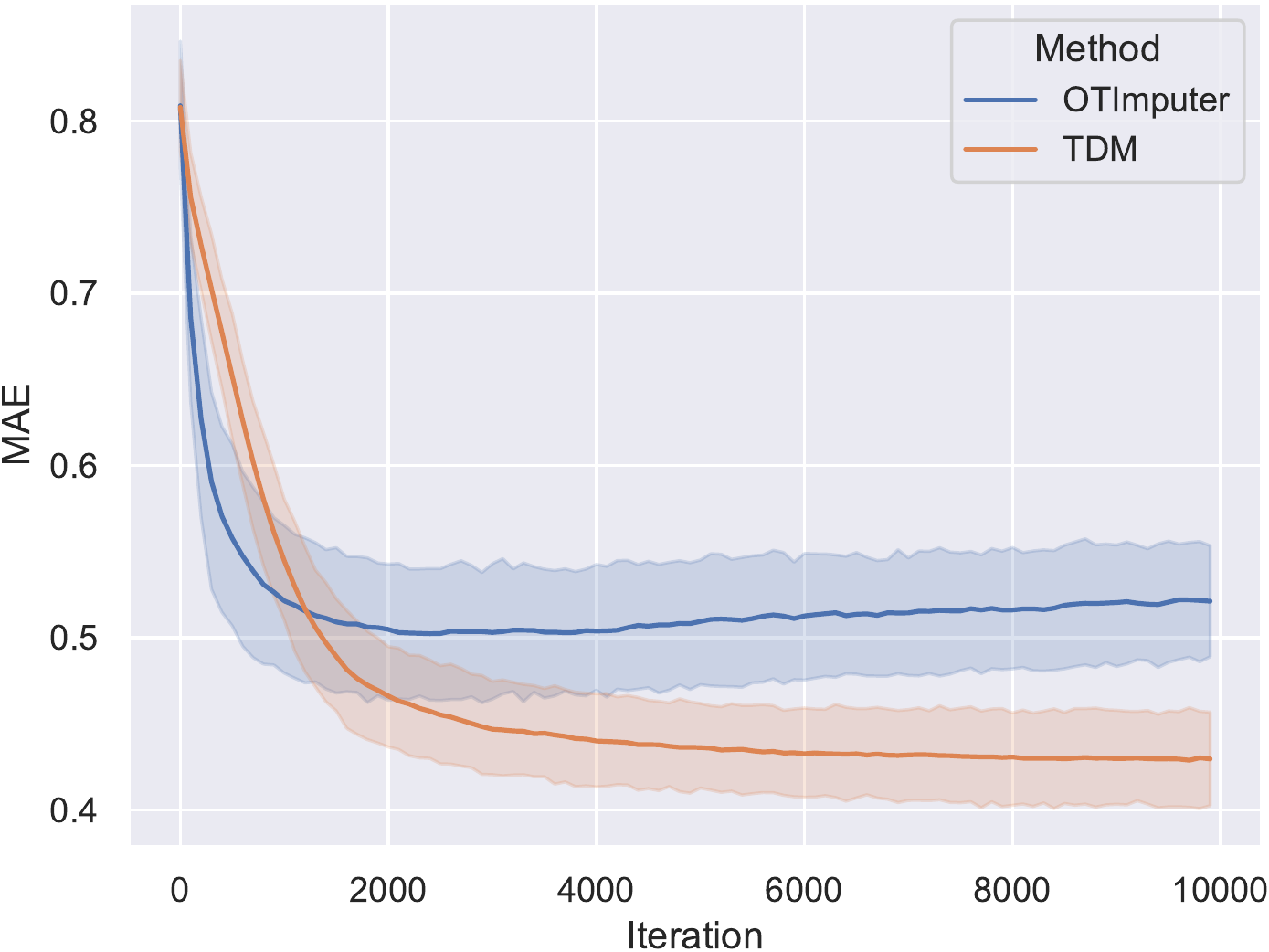}
         \end{subfigure}
                  \begin{subfigure}[b]{0.245\linewidth}
                 \centering                 \includegraphics[width=0.99\textwidth]{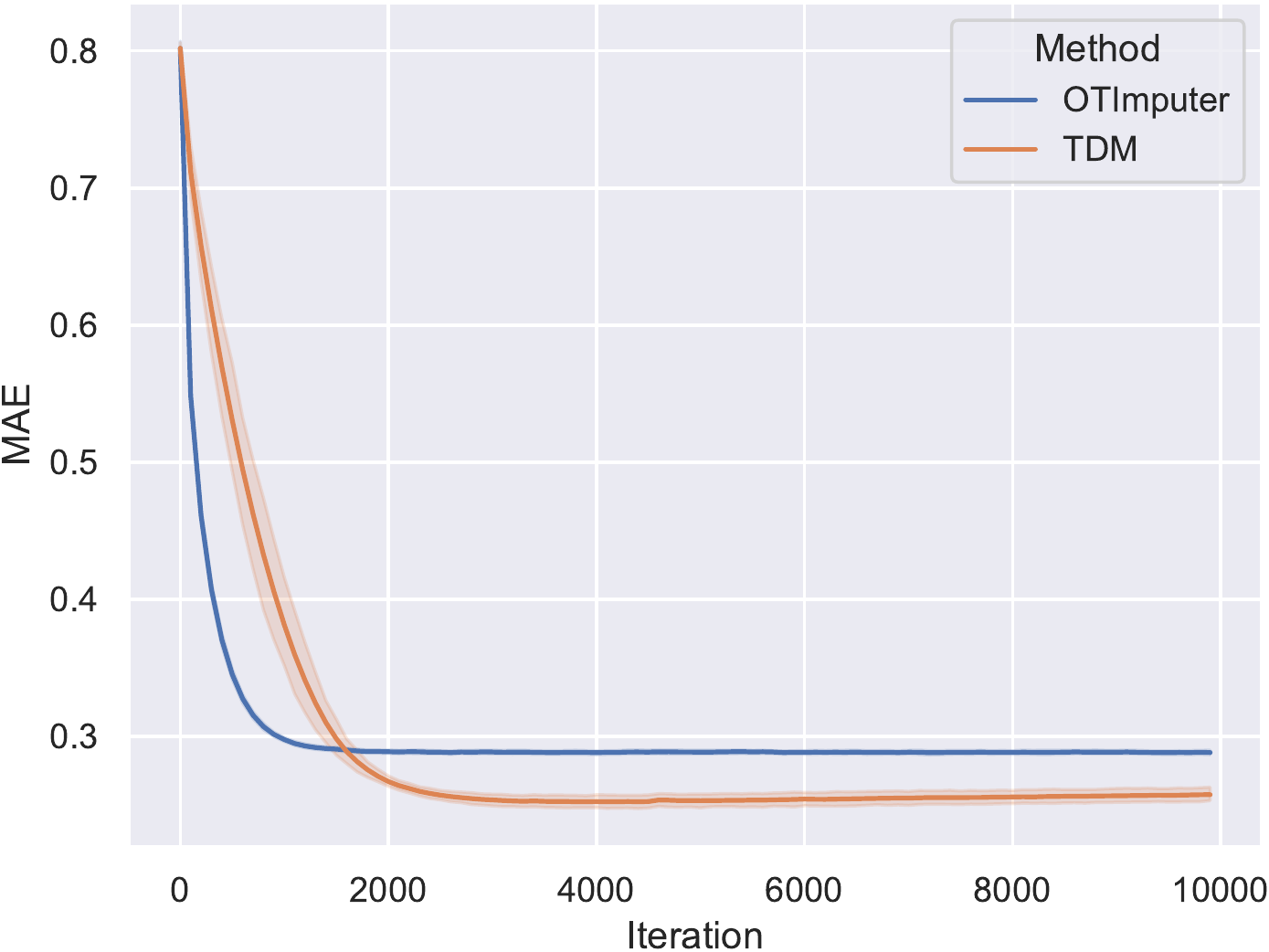}
         \end{subfigure}\\
          \begin{subfigure}[b]{0.245\linewidth}
                 \centering                 \includegraphics[width=0.99\textwidth]{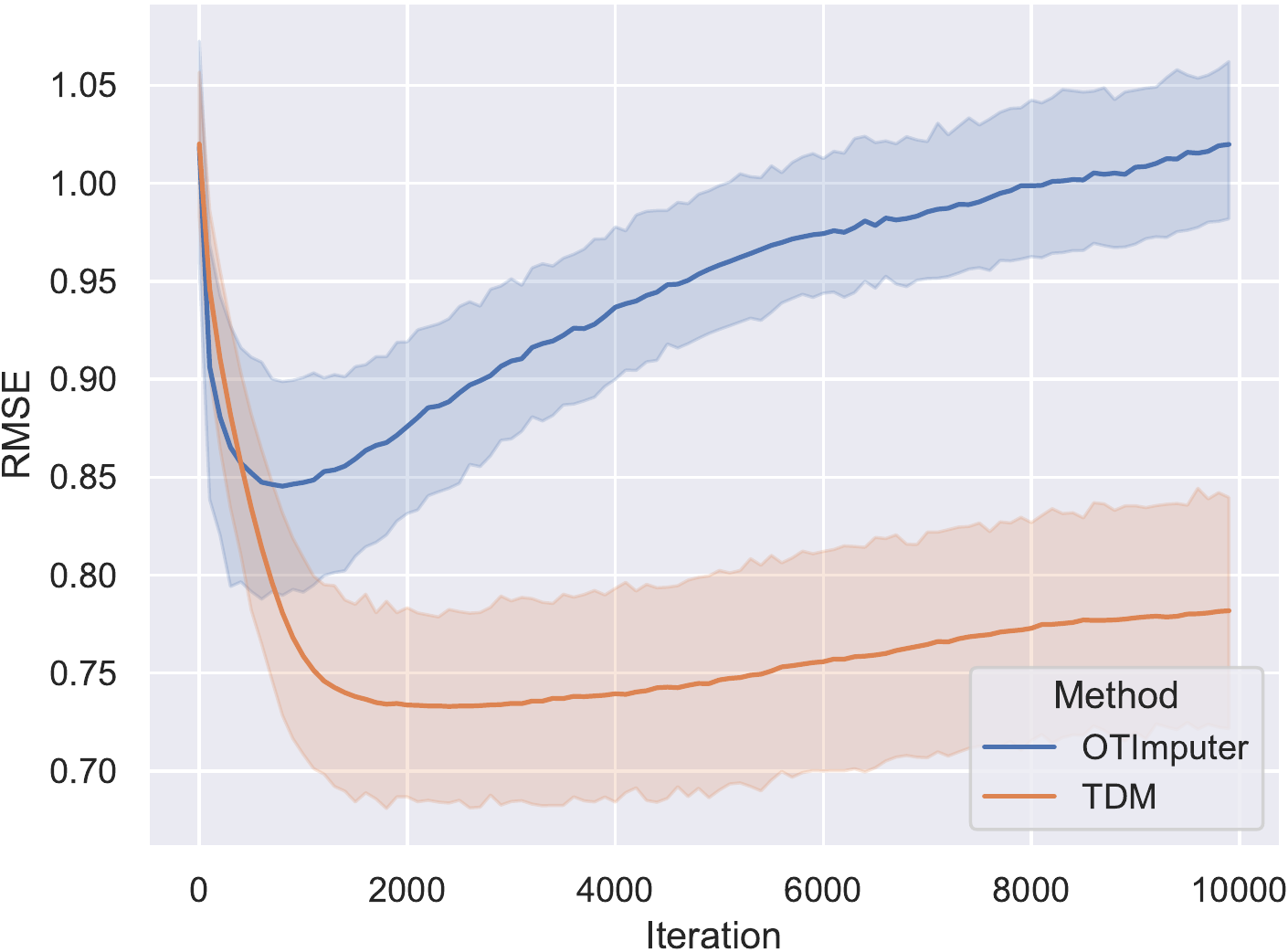}
         \end{subfigure}
                 \begin{subfigure}[b]{0.245\linewidth}
                 \centering                 \includegraphics[width=0.99\textwidth]{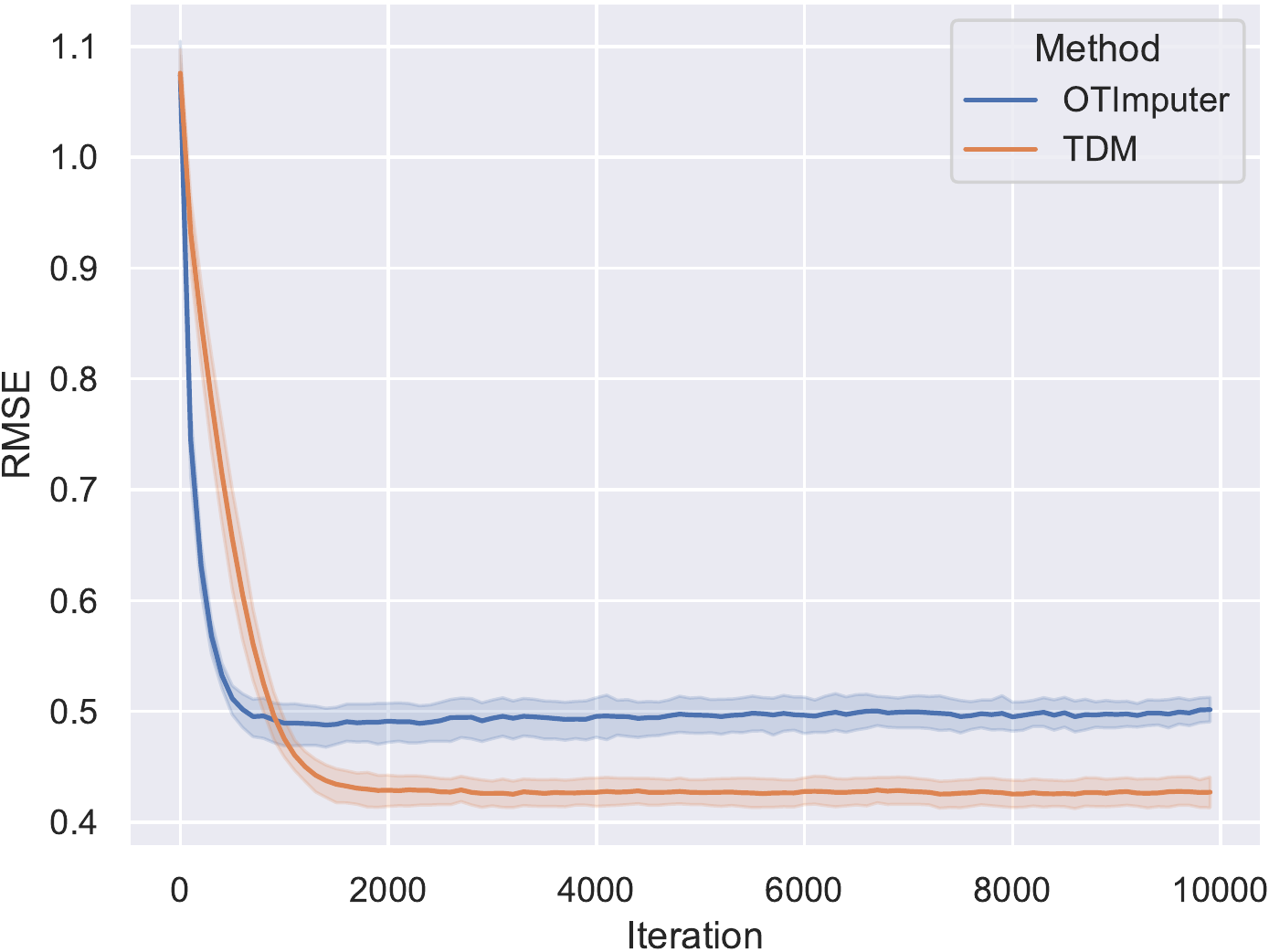}
         \end{subfigure}
          \begin{subfigure}[b]{0.245\linewidth}
                 \centering                 \includegraphics[width=0.99\textwidth]{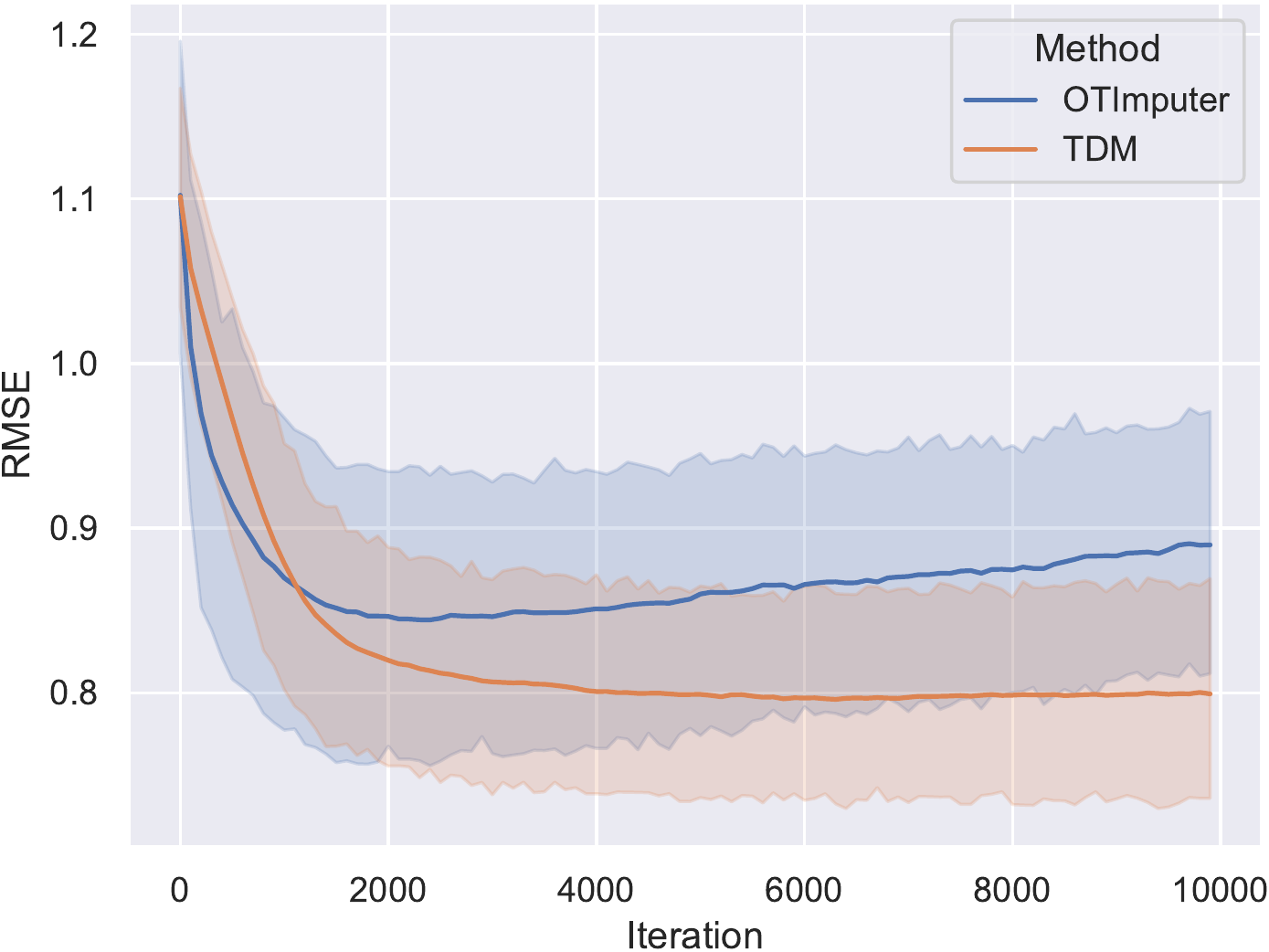}
         \end{subfigure}
                  \begin{subfigure}[b]{0.245\linewidth}
                 \centering                 \includegraphics[width=0.99\textwidth]{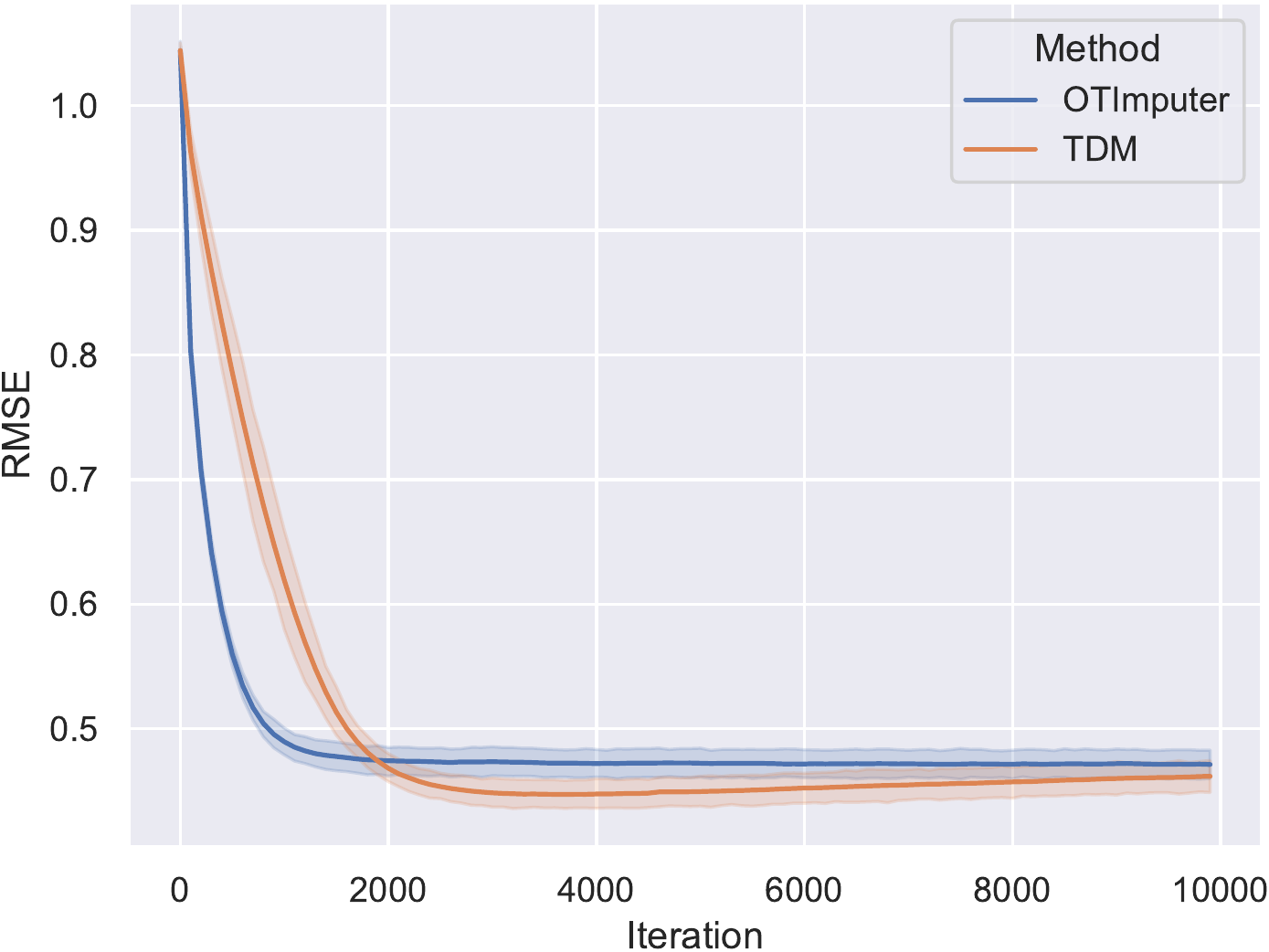}
         \end{subfigure}
 \caption{MAE and RMSE over training iterations of TDM and OTImputer on four datasets  (from left to right: glass, seeds, blood\_transfusion, anuran\_calls) in MNARL.}
  \label{fig-iter-mnar-l}
 \vspace{-0.5cm}
\end{figure}

\begin{figure}[t]
\captionsetup[subfigure]{justification=centering}
        \centering
         \begin{subfigure}[b]{0.245\linewidth}
                 \centering                 \includegraphics[width=0.99\textwidth]{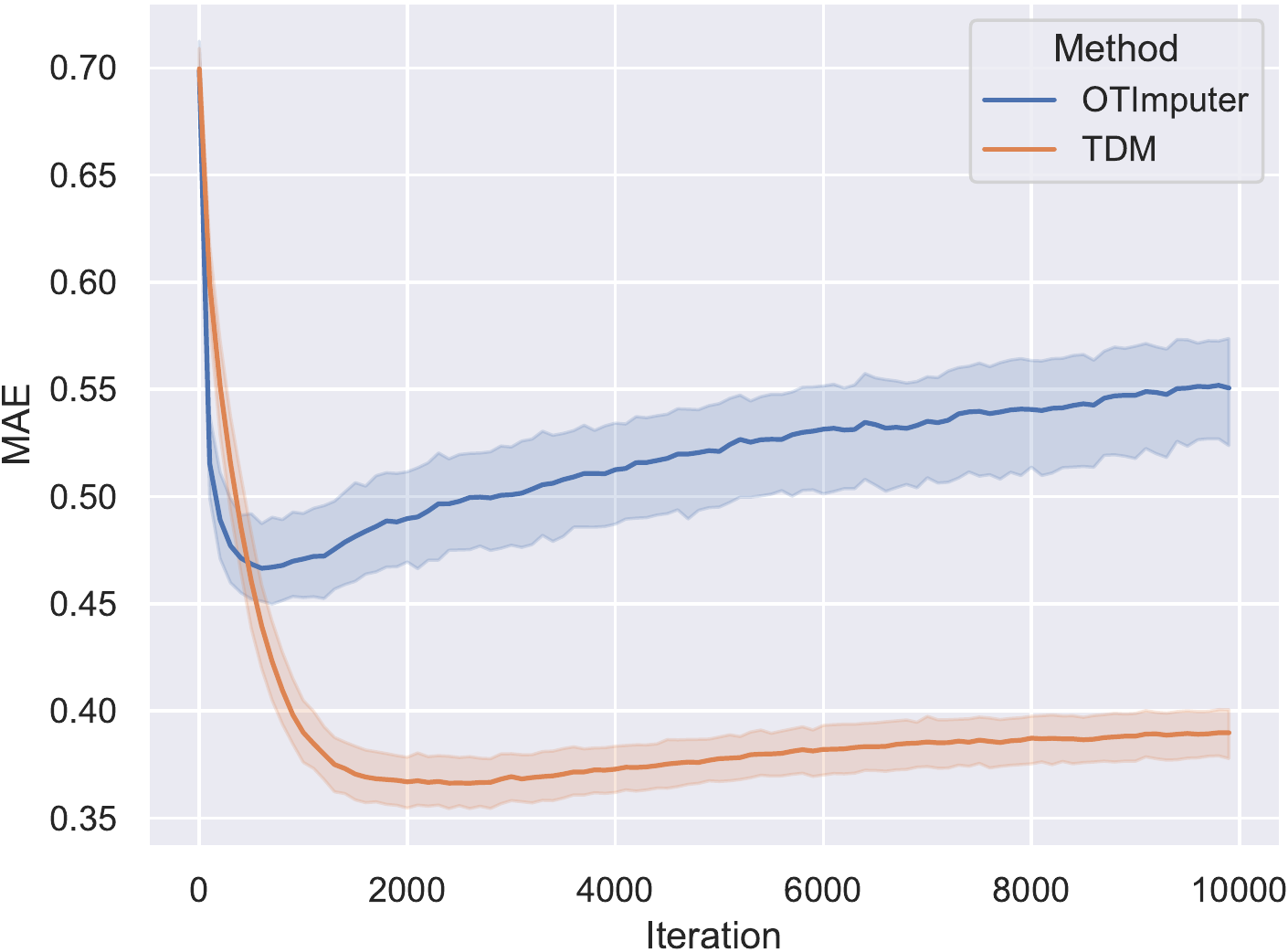}
         \end{subfigure}
                 \begin{subfigure}[b]{0.245\linewidth}
                 \centering                 \includegraphics[width=0.99\textwidth]{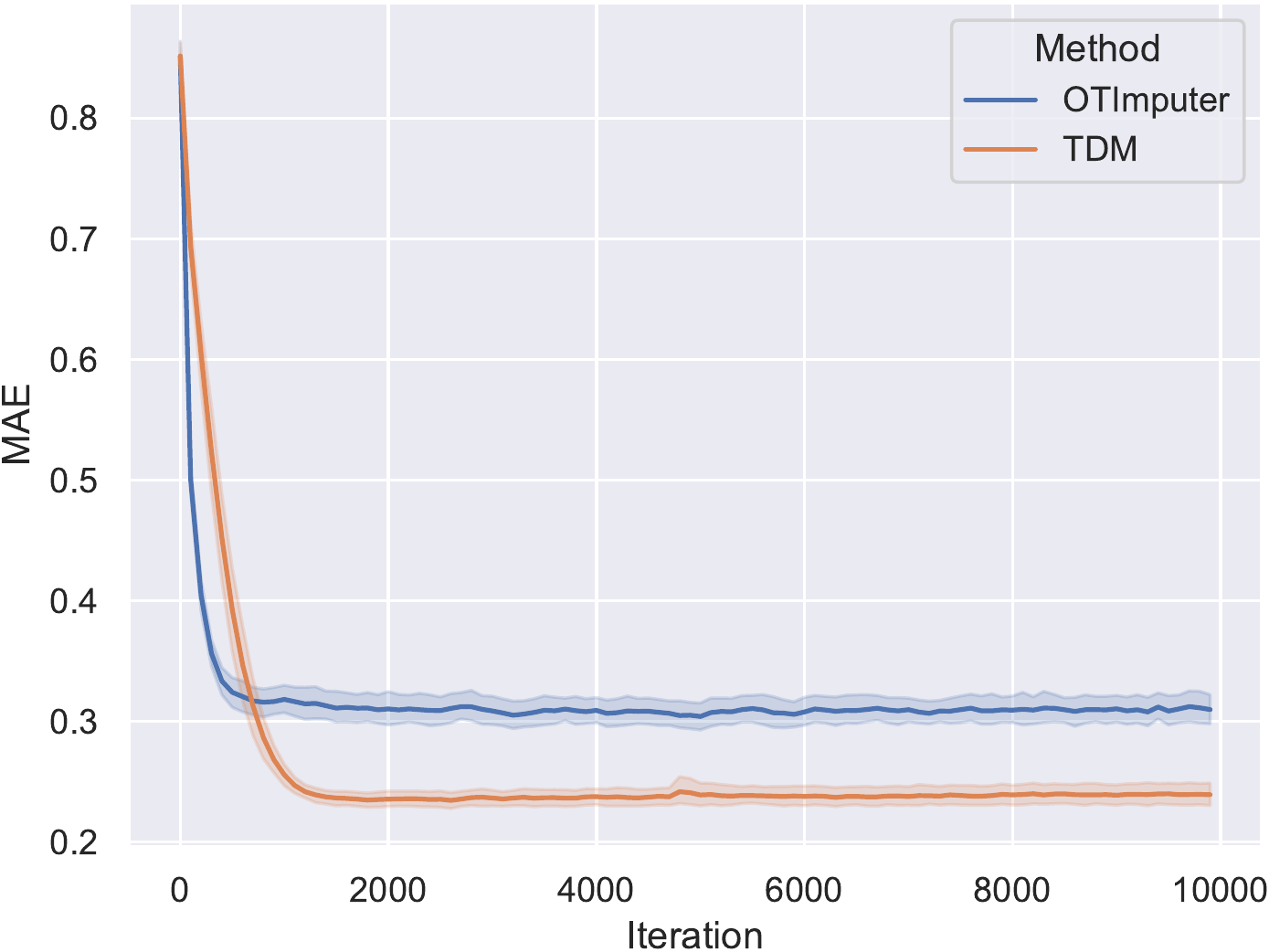}
         \end{subfigure}
          \begin{subfigure}[b]{0.245\linewidth}
                 \centering                 \includegraphics[width=0.99\textwidth]{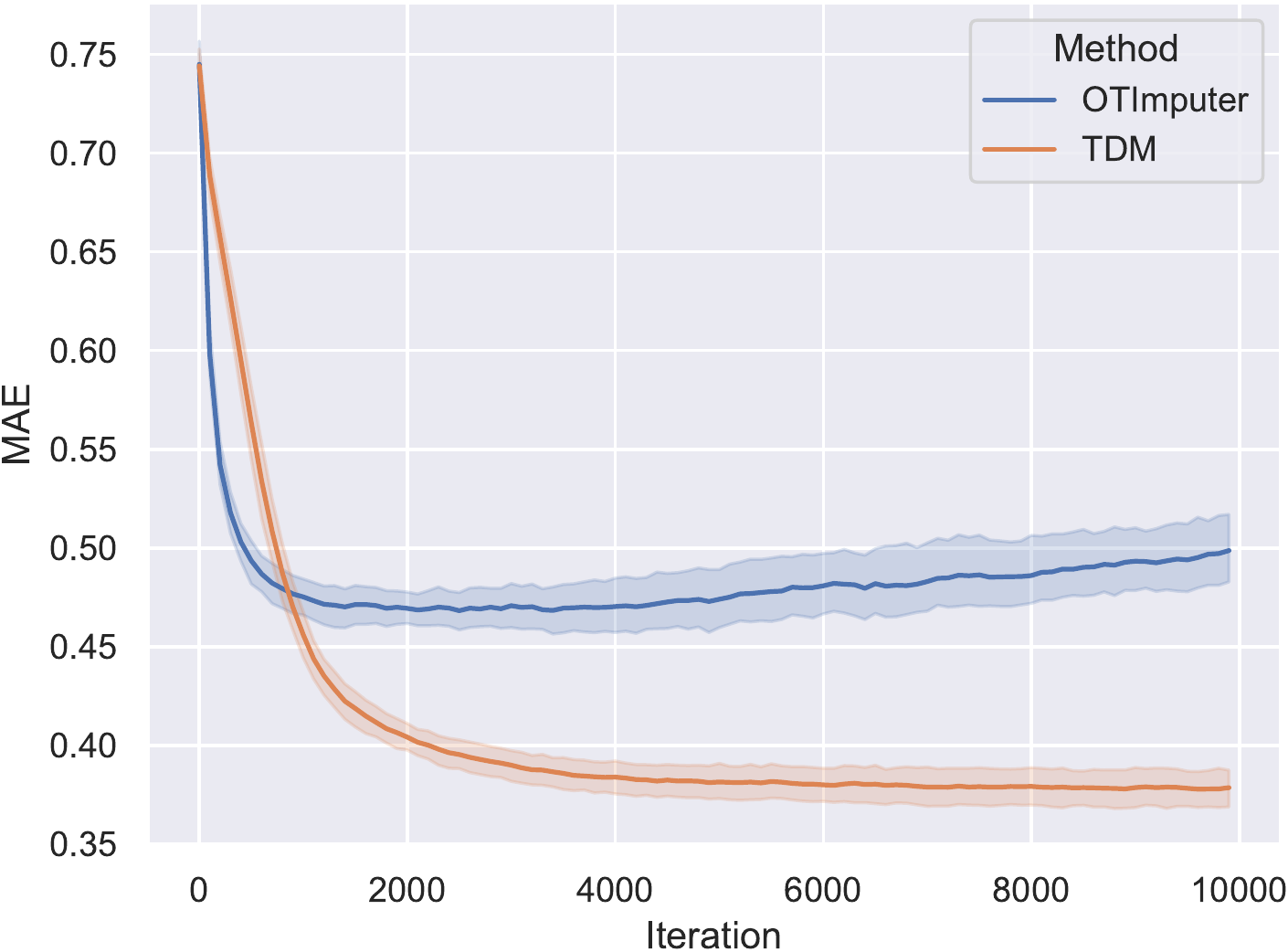}
         \end{subfigure}
                  \begin{subfigure}[b]{0.245\linewidth}
                 \centering                 \includegraphics[width=0.99\textwidth]{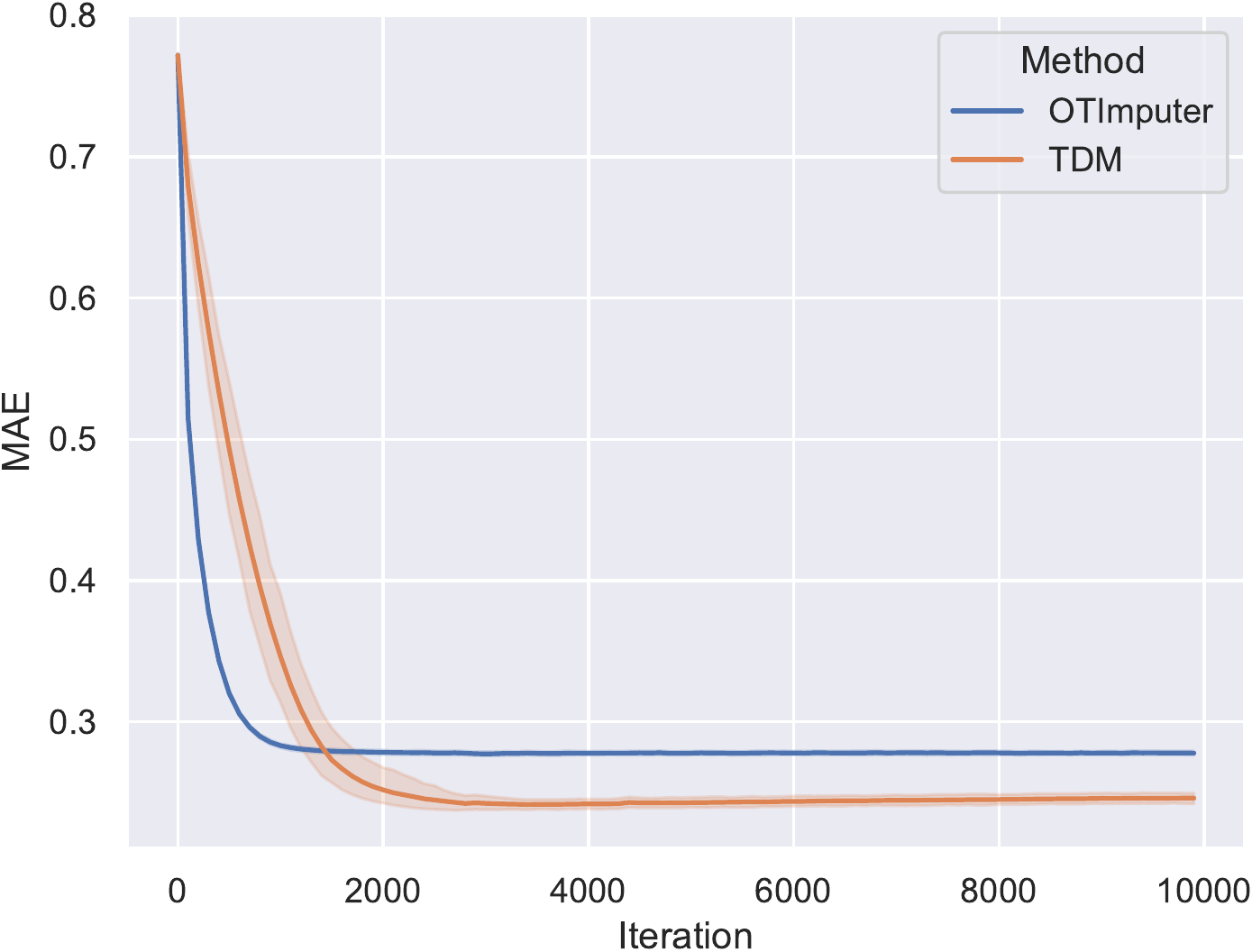}
         \end{subfigure}\\
          \begin{subfigure}[b]{0.245\linewidth}
                 \centering                 \includegraphics[width=0.99\textwidth]{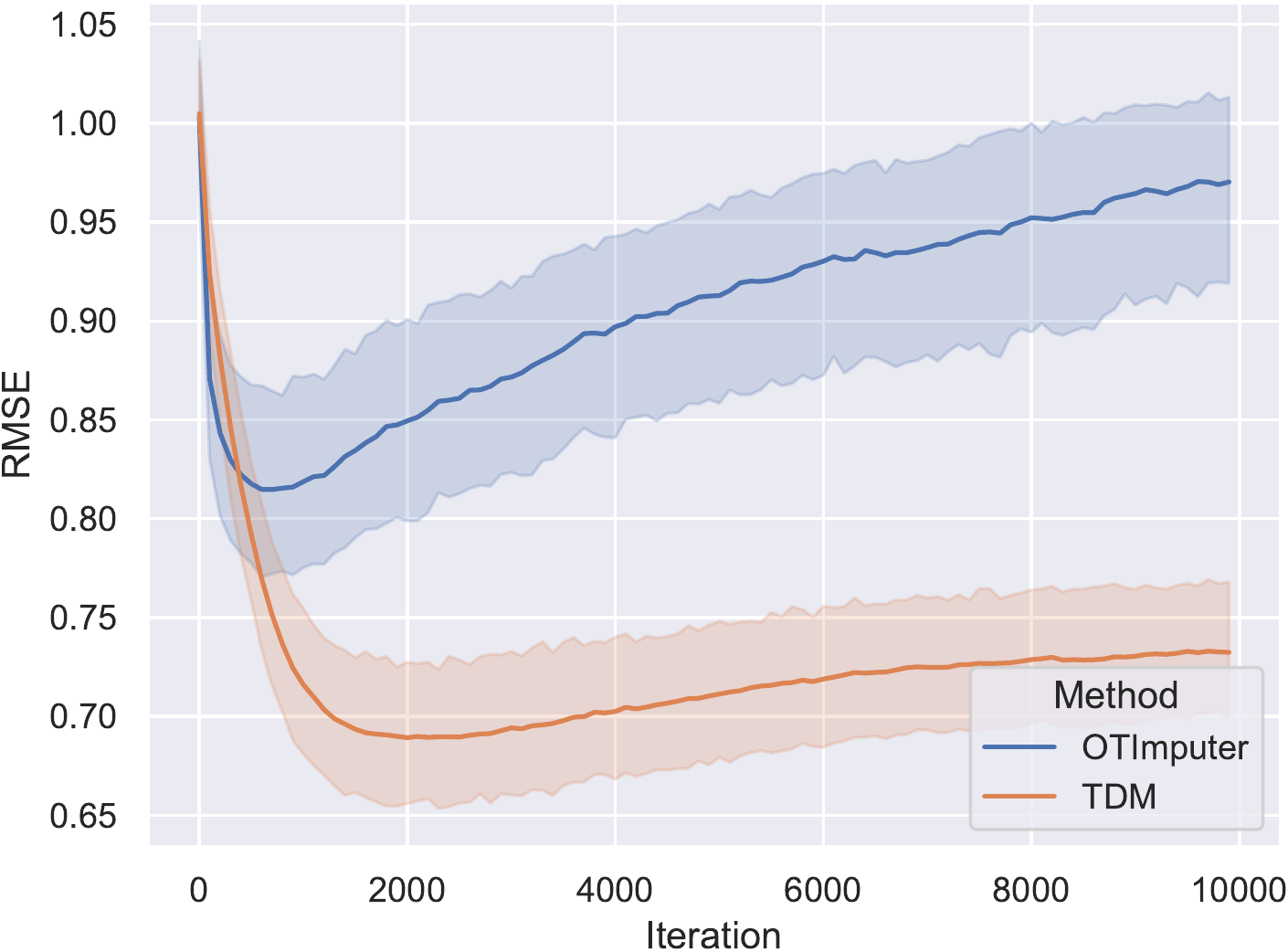}
         \end{subfigure}
                 \begin{subfigure}[b]{0.245\linewidth}
                 \centering                 \includegraphics[width=0.99\textwidth]{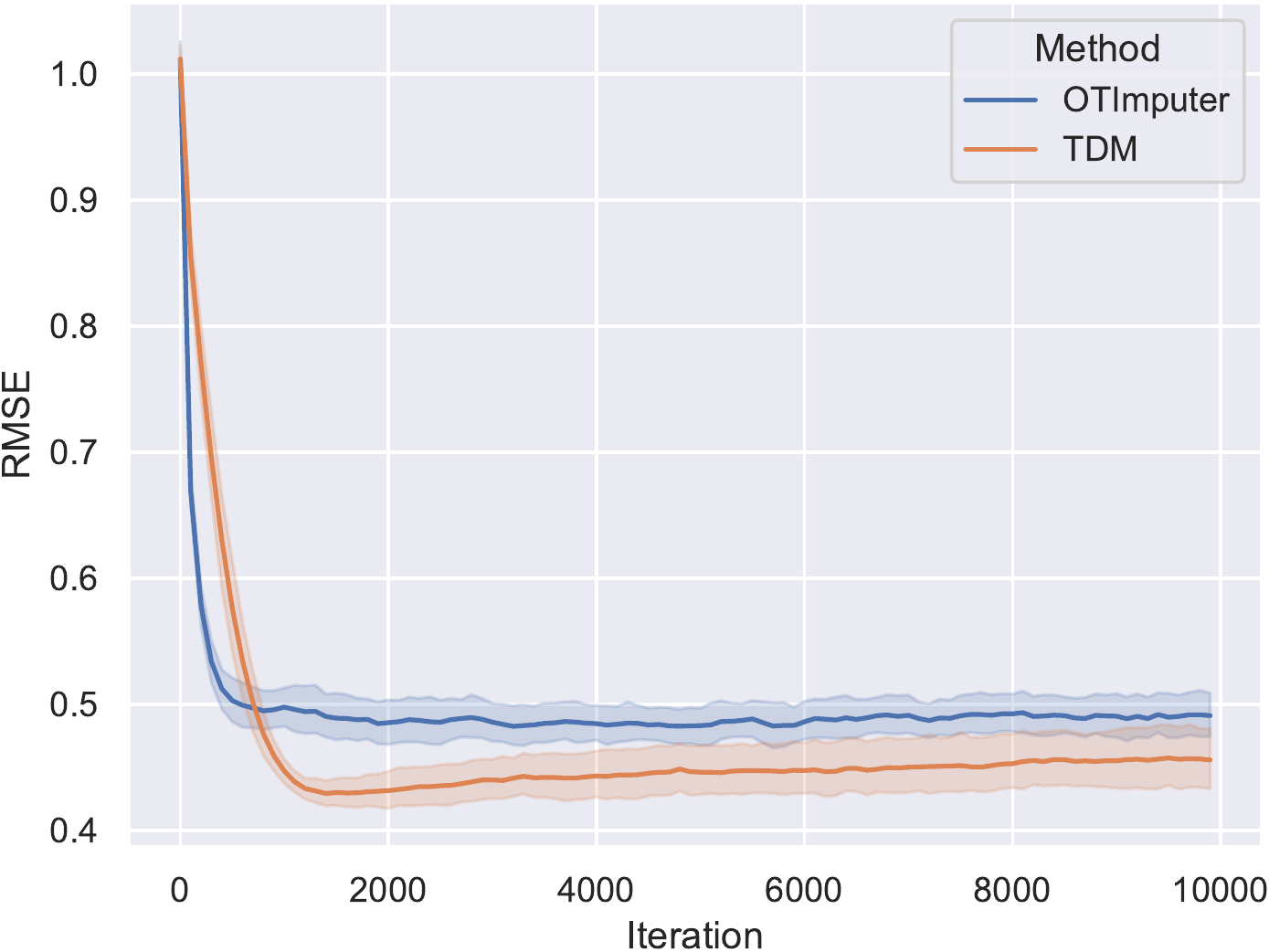}
         \end{subfigure}
          \begin{subfigure}[b]{0.245\linewidth}
                 \centering                 \includegraphics[width=0.99\textwidth]{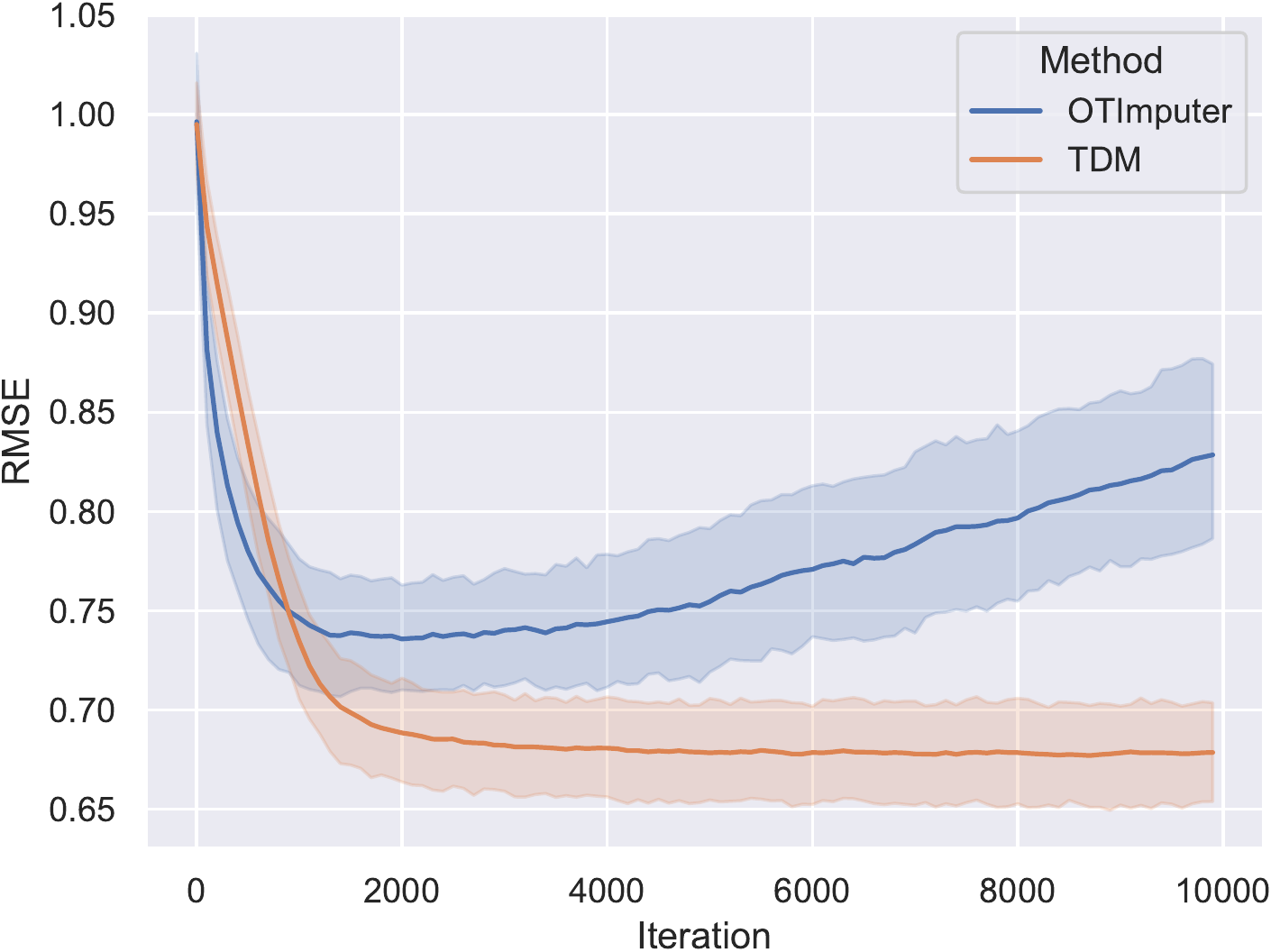}
         \end{subfigure}
                  \begin{subfigure}[b]{0.245\linewidth}
                 \centering                 \includegraphics[width=0.99\textwidth]{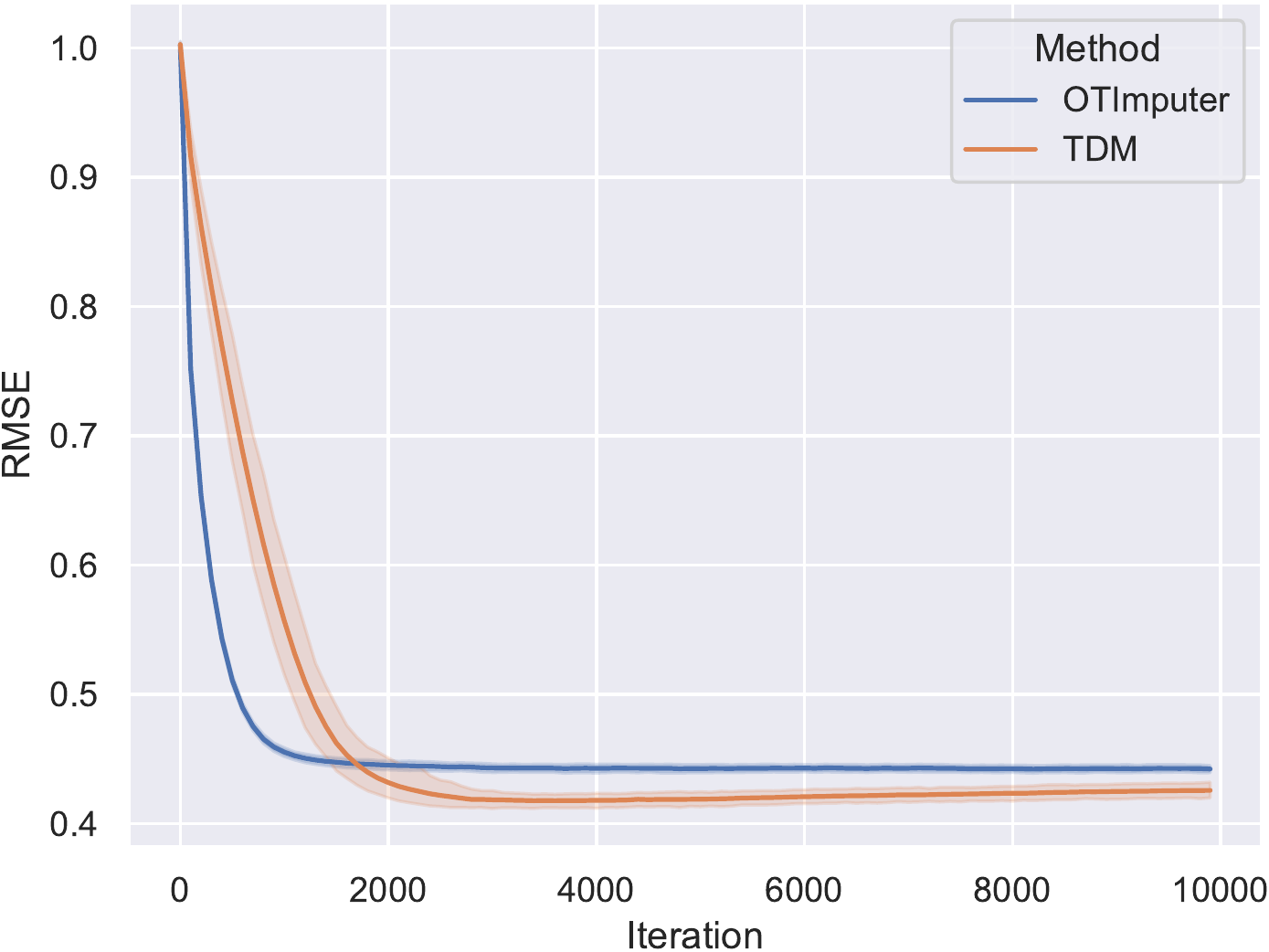}
         \end{subfigure}
 \caption{MAE and RMSE over training iterations of TDM and OTImputer on four datasets  (from left to right: glass, seeds, blood\_transfusion, anuran\_calls) in MNARQ.}
  \label{fig-iter-mnar-q}
 \vspace{-0.5cm}
\end{figure}

\end{document}